\documentclass{article}

\PassOptionsToPackage{numbers, compress}{natbib}
\usepackage[final]{neurips_2020}

\usepackage[utf8]{inputenc} % allow utf-8 input
\usepackage[T1]{fontenc}    % use 8-bit T1 fonts
\usepackage{url}            % simple URL typesetting
\usepackage{booktabs}       % professional-quality tables
\usepackage{amsfonts}       % blackboard math symbols
\usepackage{nicefrac}       % compact symbols for 1/2, etc.
\usepackage{microtype}      % microtypography

\usepackage{graphicx}
\usepackage{subcaption}
\usepackage{enumitem}

\usepackage{csquotes}
\usepackage{todonotes}
\usepackage{xspace}
\usepackage{amsmath}
\usepackage{amssymb}
\usepackage{float}
\usepackage{wrapfig}
\usepackage{lipsum}
\usepackage{xcolor,colortbl}
\usepackage{fdsymbol}
\usepackage{multirow}
\usepackage{rotating}
\usepackage{algorithm,algpseudocode}

\usepackage{hyperref}

\newcommand{\sclub}{$\textcolor{blue1}{\clubsuit}$}
\newcommand{\sspade}{$\textcolor{green1}{\spadesuit}$}
\newcommand{\sdiamond}{$\textcolor{orange1}{\vardiamondsuit}$}

\newcommand{\sstar}{$\textcolor{pink1}{\textbf{$\star$}}$}
\newcommand{\sinfinite}{$\textcolor{magenta}{\textbf{$\infty$}}$}

\newcommand{\sigm}{{\mathrm{sigmoid}}}

\newcommand{\rnn}{{\mathrm{RNN}}}
\newcommand{\ours}{GATA\xspace}
\newcommand{\oursgtp}{GATA-GTP\xspace}
\newcommand{\oursgtf}{GATA-GTF\xspace}
\newcommand{\transdqn}{Tr-DQN\xspace}
\newcommand{\transdrqn}{Tr-DRQN\xspace}
\newcommand{\transdrqnp}{Tr-DRQN+\xspace}
\newcommand{\ftwp}{\emph{FTWP}\xspace}

\newcommand{\code}[1]{\texttt{#1}}
\newcommand{\cmd}[1]{\textcolor{darkblue}{\textbf{\small{\code{#1}}}}}

\usepackage{pifont}
\newcommand{\cmark}{\ding{51}}%
\newcommand{\xmark}{\ding{55}}%

\definecolor{color1}{HTML}{da6752}
\definecolor{color2}{HTML}{5573a6}
\definecolor{green1}{HTML}{0b5400}
\definecolor{orange1}{HTML}{f3905c}
\definecolor{orange}{HTML}{ff7700}
\definecolor{cyan}{HTML}{008b8b}
\definecolor{blue}{HTML}{0000ff}
\definecolor{lb}{HTML}{deecff}
\definecolor{ly}{HTML}{fffeba}
\definecolor{purple1}{HTML}{9258cc}
\definecolor{blue1}{HTML}{027db5}
\definecolor{cyan1}{HTML}{00a0b5}
\definecolor{pink1}{HTML}{ff7a7a}
\definecolor{red1}{HTML}{8a0000}
\definecolor{blue2}{HTML}{041480}
\definecolor{yellowishgreen}{HTML}{2e8a00}
\definecolor{magenta}{HTML}{9b00a1}
\definecolor{darkblue}{HTML}{240394}

\definecolor{lg}{HTML}{edfff0}
\definecolor{lr}{HTML}{ffebeb}
\definecolor{lgy}{HTML}{f7f7f7}

\definecolor{lred}{HTML}{ffb8b8}
\definecolor{lgreen}{HTML}{b8ffc3}
\definecolor{lblue}{HTML}{b8e6ff}
\definecolor{lpurple}{HTML}{cab8ff}
\definecolor{lorange}{HTML}{ffdab8}
\definecolor{lcyan}{HTML}{abfff9}
\definecolor{lmagenta}{HTML}{ffbfed}

\newcommand{\pc}{\cellcolor{lr}}
\newcommand{\nc}{\cellcolor{lg}}
\newcommand{\oc}{\cellcolor{lgy}}

\title{Learning Dynamic Belief Graphs to Generalize\\ on Text-Based Games}

\author{
Ashutosh Adhikari$^\dag$\thanks{\:\:\:\:Equal contribution.} \:\:\:\: Xingdi Yuan$^\heartsuit$\footnotemark[1]  \:\:\:\: Marc-Alexandre C\^ot\'{e}$^\heartsuit$\footnotemark[1]\\
\textbf{Mikul\'{a}\v{s} Zelinka$^\ddag$ \:\:\:\: Marc-Antoine Rondeau$^\heartsuit$}\\
\textbf{Romain Laroche$^\heartsuit$ \:\: Pascal Poupart$^{\dag\P}$ \:\: Jian Tang$^{\spadesuit\clubsuit}$}\\
\textbf{Adam Trischler$^{\heartsuit}$ \:\: William L. Hamilton$^{\diamondsuit\clubsuit}$}\\
$^\dag$University of Waterloo \:\:\:\: $^\heartsuit$Microsoft Research, Montr\'{e}al \:\:\:\: $^\ddag$Charles University\\
$^\clubsuit$Mila \:\:\:\: $^\diamondsuit$McGill University \:\:\:\:$^\spadesuit$ HEC Montr\'{e}al \:\:\:\: $^\P$Vector Institute\\
eric.yuan@microsoft.com\\
}

\begin{document}

\maketitle

\begin{abstract}
%%%%%%%%%%%%%%%%%%%%%%%%%%%%%%%%%%%%%%%%%%%%%
% DO NOT MODIFY THIS ABSTRACT
%%%%%%%%%%%%%%%%%%%%%%%%%%%%%%%%%%%%%%%%%%%%%
Playing text-based games requires skills in processing natural language and sequential decision making. 
Achieving human-level performance on text-based games remains an open challenge, and prior research has largely relied on hand-crafted structured representations and heuristics. % to enable good performance.
In this work, we investigate how an agent can plan and generalize in text-based games using graph-structured representations learned end-to-end from raw text.
We propose a novel graph-aided transformer agent (GATA) that infers and updates latent {\em belief graphs} during planning to enable effective action selection by capturing the underlying game dynamics.
\ours is trained using a combination of reinforcement and self-supervised learning.
Our work demonstrates that the learned graph-based representations help agents converge to better policies than their text-only counterparts and facilitate effective generalization across game configurations.
Experiments on 500+ unique games from the TextWorld suite show that our best agent outperforms text-based baselines by an average of 24.2\%.
\end{abstract}
%%%%%%%%%%%%%%%%%%%%%%%%%%%%%%%%%%%%%%%%%%%%%
% DO NOT MODIFY THIS ABSTRACT
%%%%%%%%%%%%%%%%%%%%%%%%%%%%%%%%%%%%%%%%%%%%%
\section{Introduction}
\label{section:intro}
% Story Line
% 1. language is the key to interact with human world --->
% 2, text game is a good proxy of language+interaction --->
% 3. they are hard, existing literature either overfit or hand-craft --->
% 4. thus we learn to build belief graph by the agent itself
% 5. we propose \ours for generalization
%Language is at the core of many human interactions, especially when interactions involve the transfer of knowledge. Consider recording information in written text, extracting information from text by reading, communicating with other humans, and so on.
%We expect intelligent agents to have the ability to interact with the human world
%either in reality (e.g., customer service systems) or in simulated environments (e.g., video games).
%Therefore, it makes sense to teach agents to interact through language.

Text-based games are complex, interactive simulations in which the game state is described with text and players act using simple text commands (e.g., \cmd{light torch with match}). 
They serve as a proxy for studying how agents can exploit language to comprehend and interact with the environment.
Text-based games are a useful challenge in the pursuit of intelligent agents that communicate with humans (e.g., in customer service systems).

Solving text-based games requires a combination of reinforcement learning (RL) and natural language processing (NLP) techniques.
However, inherent challenges like partial observability, long-term dependencies, sparse rewards, and combinatorial action spaces make these games very difficult.\footnote{We challenge readers to solve this representative game: \url{https://aka.ms/textworld-tryit}.}
For instance, \citet{hausknecht19nail} show that a state-of-the-art model achieves a mere $2.56\%$ of the total possible score on a curated set of text-based games for human players~\citep{Atkinson2018}.
On the other hand, while text-based games exhibit many of the same difficulties as linguistic tasks like open-ended dialogue, they are more structured and constrained.

To design successful agents for text-based games, previous works have relied largely on heuristics that exploit games' inherent structure.
For example, several works have proposed rule-based components that prune the action space or shape the rewards according to \emph{a priori} knowledge of the game dynamics~\citep{yuan2018counting, Lima2019, adolphs19ledeepchef, yin19learn}.
More recent approaches take advantage of the graph-like structure of text-based games by building knowledge graph (KG) representations of the game state:
\citet{ammanabrolu19graph, ammanabrolu2020graph}, for example, use hand-crafted heuristics to populate a KG that feeds into a deep neural agent to inform its policy. 
Despite progress along this line, we expect more general, effective representations for text-based games to arise in agents that learn and scale more automatically, which replace heuristics with learning~\citep{suttonbitterlession}.

This work investigates how we can learn graph-structured state representations for text-based games in an entirely data-driven manner.
We propose the graph aided transformer agent (\ours)\footnote{Code and dataset used: \url{https://github.com/xingdi-eric-yuan/GATA-public}} that, in lieu of heuristics, \emph{learns} to construct and update graph-structured beliefs\footnote{Text-based games are partially observable environments.} and use them to further optimize rewards.
% We propose the Graph Aided Transformer Agent (\ours).~\footnote{Code and dataset used: \url{https://github.com/<BLIND>/<BLIND>.git}}
% \ours is a deep neural agent that, in lieu of heuristics, \emph{learns} to construct and update a graph-structured beliefs\footnote{Text-based games are partially observable environments.} and use them to optimize rewards.
We introduce two self-supervised learning strategies---based on text reconstruction and mutual information maximization---which enable our agent to learn latent graph representations without direct supervision or hand-crafted heuristics.

We benchmark \ours on 500+ unique games generated by TextWorld~\citep{cote18textworld}, evaluating performance in a setting that requires generalization across different game configurations. 
We show that \ours outperforms strong baselines, including text-based models with recurrent policies. %that leverage a recurrent hidden state.
In addition, we compare \ours to agents with access to ground-truth graph representations of the game state. We show that \ours achieves competitive performance against these baselines even though it receives only partial text observations of the state.
Our findings suggest, promisingly, that graph-structured representations provide a useful inductive bias for learning and generalizing in text-based games, and act as a memory 
enabling agents to optimize rewards in a partially observed setting.

\begin{figure*}[t!]
    \centering
    \includegraphics[width=1.0\textwidth]{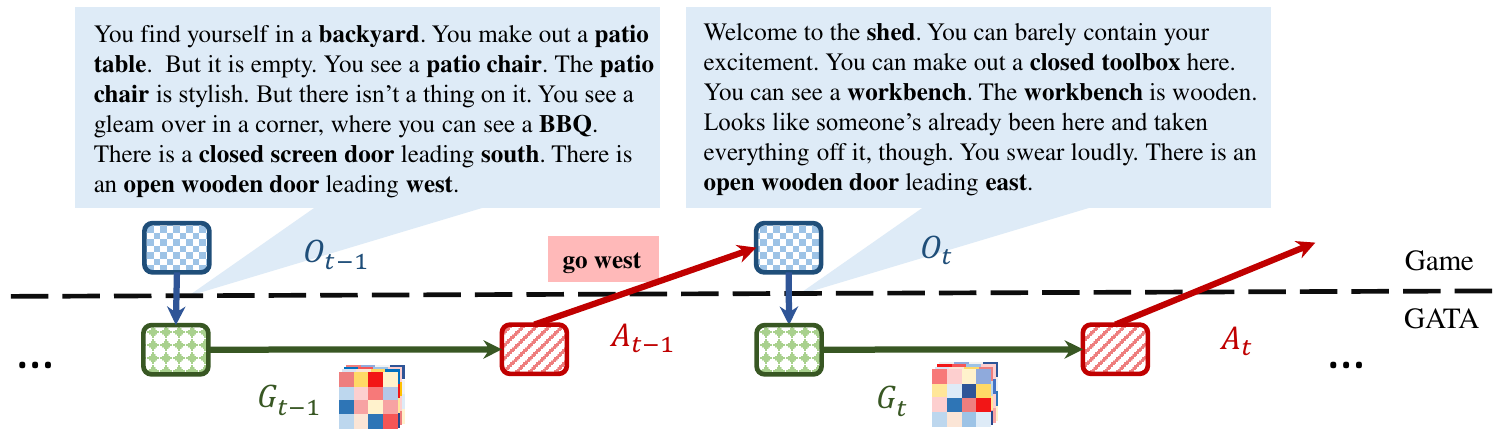}
    \caption{\ours playing a text-based game by updating its belief graph. In response to action \textcolor{red1}{$A_{t-1}$}, the environment returns text observation \textcolor{blue2}{$O_t$}. Based on  \textcolor{blue2}{$O_t$} and \textcolor{green1}{$\mathcal{G}_{t-1}$}, the agent updates \textcolor{green1}{$\mathcal{G}_{t}$} and selects a new action \textcolor{red1}{$A_{t}$}. In the figure, \textcolor{blue2}{blue box with squares} is the game engine, \textcolor{green1}{green box with diamonds} is the graph updater, \textcolor{red1}{red box with slashes} is the action selector.}
    \label{fig:kg}
\end{figure*}

\section{Background}
\label{section:background}

\textbf{Text-based Games:}
% The following sentence is a bit redundant given the introduction.
Text-based games can be formally described as partially observable Markov decision processes (POMDPs)~\citep{cote18textworld}.
%They are interactive environments in which a player must reach a specified objective leveraging the text-only descriptions by issuing text actions (e.g., \cmd{insert yellow pepper into fridge}, \cmd{go east}). Figure \ref{fig:kg} illustrates an excerpt from a game in progress where the player decides to go west after being given the textual description of her current location (i.e. backyard).
They are environments in which the player receives text-only observations $O_t$ (these describe the observable state, typically only partially) and interacts by issuing short text phrases as actions $A_t$ (e.g., in Figure~\ref{fig:kg}, \cmd{go west} moves the player to a new location).
Often, the end goal is not clear from the start; %, and the player must figure it out while being guided by some sparse reward.
the agent must infer the objective by earning sparse rewards for completing subgoals.
%These games have pre-defined objectives, usually unbeknownst to the player, for which a player may get points upon completing them. can award point.% an agent is required to navigate, collect, craft, even combat by interacting with the partially observed environment to achieve the objectives.
%, represented as $(S, T, A, \Omega, O, R, \gamma)$, where $S$ and $A$ are the sets of states and text actions, $T$ is the conditional transition probability function, $\Omega$ is the set of observations, $O$ represents the conditional observation probabilities, $R$ indicates the reward function, and $\gamma$ is the discount factor. 
% Text-based games come in a variety of types and difficulty levels. 
Text-based games have a variety of difficulty levels determined mainly by the environment's complexity (i.e., how many locations in the game, and how many objects are interactive), the game length (i.e., optimally, how many actions are required to win), and the verbosity (i.e., how much text information is irrelevant to solving the game).

% Text-based games fall broadly into two settings: \emph{parser-based}, where the player must generate text actions word by word or character by character, and \emph{choice-based}, where the player chooses action phrase $A_t$ from a list of action candidates $C_t$ at every game step $t$. Further, they have a variety of difficulty levels determined mainly by the environment's complexity (i.e., how many locations in the game, and how many objects are interactive), the game length (i.e., optimally, how many actions are required to win), and the verbosity (i.e., how much text information is irrelevant to solving the game).

\textbf{Problem Setting:}
We use TextWorld~\citep{cote18textworld} to generate unique \emph{choice-based} games of varying difficulty. All games share the same overarching theme: an agent must gather and process cooking ingredients, placed randomly across multiple locations, according to a recipe it discovers during the game. The agent earns a point for collecting each ingredient and for processing it correctly. The game is won upon completing the recipe.
Processing any ingredient incorrectly terminates the game (e.g., \cmd{slice carrot} 
when the recipe asked for a \emph{diced carrot}). 
% Further, a point is awarded for combining the ingredient using a special action \cmd{prepare meal}, and another for eating the meal, at which point the game is won. 
To process ingredients, an agent must find and use appropriate tools (e.g., a knife to \cmd{slice}, \cmd{dice}, or \cmd{chop}; a stove to \cmd{fry}, an oven to \cmd{roast}). 
% the mis-operation of any ingredient will lead to a game loss.
% Similarly but more interestingly, to cook an ingredient, an agent needs to stand by a stove or oven and issue \cmd{cook} action, depending on the heat source, the product will be fried or roasted.

We divide generated games, all of which have unique recipes and map configurations, into sets for training, validation, and test.
Adopting the supervised learning paradigm for evaluating generalization, we tune hyperparameters on the validation set and report performance on a test set of previously unseen games.
Testing agents on unseen games (within a difficulty level) is uncommon in prior RL work, where it is standard to train and test on a single game instance.
Our approach enables us to measure the robustness of learned policies as they generalize (or fail to) across a ``distribution'' of related but distinct games.
%while previous work end up developing configuration-specific rather than game-specific policies. 
Throughout the paper, we use the term {\em generalization} to imply the ability of a single policy to play a distribution of related games (within a particular difficulty level).

% We have a total of 99 nodes and 10 relations across all games, however each game has a randomly built map, a set of randomly placed objects and a randomly generated recipe, there is always only a subset of the 99 nodes appeared in each game.

% Previously, \citet{trischler19ftwp} presented the \emph{First TextWorld Problems (FTWP)} dataset, which consists of TextWorld games that follow a cooking theme across a wide range of difficulty levels. 
% Although that dataset is analogous to what we use in this work, it has only 10 games per difficulty level.
% This is insufficient for reliable experiments on generalization, so
% we generate new game sets for our work.
% We use \ftwp games for self-supervised learning tasks, however (see Section~\ref{subsection:graph_update}).
%Note that there is no overlap between the \ftwp games and the games we use in Section~\ref{section:exp}.

\textbf{Graphs and Text-based Games:}
We expect graph-based representations to be effective for text-based games because the state in these games adheres to a graph-like structure.
The essential content in most observations of the environment corresponds either to entity attributes (e.g., the state of the \cmd{carrot} is \cmd{sliced}) or to relational information about entities in the environment (e.g., the \cmd{kitchen} is \cmd{north\_of} the \cmd{bedroom}).
This information is naturally represented as a dynamic graph $\mathcal{G}_t=(\mathcal{V}_t, \mathcal{E}_t)$, where the vertices $\mathcal{V}_t$ represent entities (including the player, objects, and locations) and their current conditions (e.g., closed, fried, sliced), while the edges $\mathcal{E}_t$ represent relations between entities (e.g., \cmd{north\_of}, \cmd{in}, \cmd{is}) that hold at a particular time-step $t$.
By design, in fact, the full state of any game generated by TextWorld can be represented explicitly as a graph of this type \cite{zelinka2019building}.
%This underlying graphical game state inspired the \ours model.
The aim of our model, \ours, is to estimate the game state by learning to build graph-structured beliefs from raw text observations.
In our experiments, we benchmark \ours against models with direct access to the ground-truth game state rather than \ours's noisy estimate thereof inferred from text.

\section{Graph Aided Transformer Agent (\ours)}
\label{section:gata}

In this section, we introduce \ours, a novel transformer-based neural agent that can infer a graph-structured belief state and use that state to guide action selection in text-based games. 
As shown in Figure~\ref{fig:agent_diagram}, the agent consists of two main modules: a graph updater and an action selector.
\footnote{The graph updater and action selector share some structures but not their parameters (unless specified).} 
%\footnote{The graph updater and action selector share architectures in some sub-modules. However, by default they do not share parameters of these sub-modules.} 
At game step $t$, the graph updater extracts relevant information from text observation $O_t$ and updates its belief graph $\mathcal{G}_t$ accordingly.
The action selector issues action $A_t$ conditioned on $O_t$ and the belief graph $\mathcal{G}_t$.
Figure~\ref{fig:kg} illustrates the interaction between \ours and a text-based game. 

\subsection{Belief Graph}

We denote by $\mathcal{G}$ a belief graph representing the agent's belief about the true game state according to what it has observed so far.
We instantiate $\mathcal{G} \in [-1, 1]^{\mathcal{R} \times \mathcal{N} \times \mathcal{N}}$ as a real-valued adjacency tensor, where $\mathcal{R}$ and $\mathcal{N}$ indicate the number of relation types and entities. %, respectively.
Each entry ${\{r,i,j\}}$ in $\mathcal{G}$ indicates the strength of an inferred relationship $r$ from entity $i$ to entity $j$.
% Note that there is no {\em a priori} semantics assigned to the different relationships; the model is free to use the $\mathcal{R}$ different relation dimensions as it sees fit.
% However, we do ground the $\mathcal{N}$ entities to the actual game itself: we use one dimension to represent each different entity in the environment.\footnote{TextWorld has a maximum number of 99 entities, so we keep the number of entities fixed at $\mathcal{N}=99$.} 
We select $\mathcal{R}=10$ and $\mathcal{N}=99$ to match the maximum number of relations and entities in our TextWorld-generated games.
%We ground the $\mathcal{N}$ entities and $\mathcal{R}$ relations in the environment: since the TextWorld-generated games we use have a maximum number of 99 entities and 10 relations, we fix the number of entities and relations at $\mathcal{N}=99$ and $\mathcal{R}=10$.
In other words, we assume that \ours\ has access to the {\em vocabularies} of possible relations and entities but it must learn the structure among these objects, and their semantics, from scratch.

\subsection{Graph Updater}
\label{subsection:graph_update}

\begin{figure}[t!]
    \centering
    \includegraphics[width=0.95\textwidth]{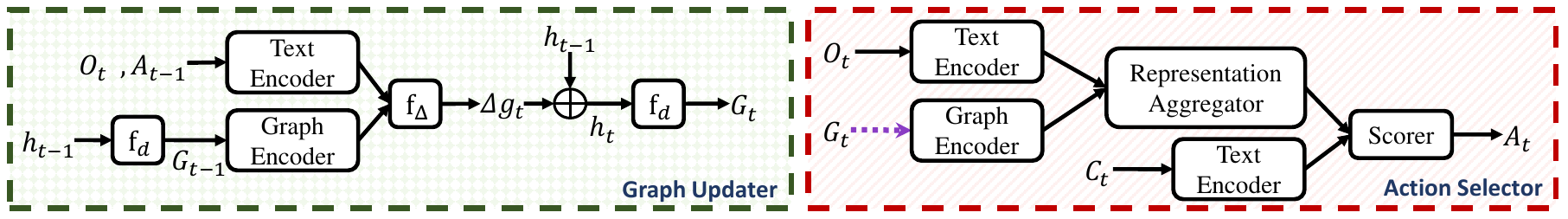}
    \caption{\ours in detail. The coloring scheme is same as in Figure~\ref{fig:kg}. The \textcolor{green1}{graph updater} first generates $\Delta g_t$ using $\mathcal{G}_{t-1}$ and $O_t$. Afterwards the \textcolor{red1}{action selector} uses $O_t$ and the updated graph $\mathcal{G}_t$ to select $A_t$ from the list of action candidates $C_t$. Purple dotted line indicates a detached connection (i.e., no back-propagation through such connection).}
    \label{fig:agent_diagram}
\end{figure}

The graph updater constructs and updates the dynamic belief graph $\mathcal{G}$ from text observations $O_t$.
% Rather than generating the entire belief graph at every game step $t$, we generate a graph update $\Delta g_t \in \mathbb{R}^\mathcal{F}$ that represents the change of the agent's belief after receiving a new observation $O_t$: 
Rather than generating the entire belief graph at each step $t$, we generate a graph update, $\Delta g_t$, that represents the change of the agent's belief after receiving a new observation. This is motivated by the fact that observations $O_t$ typically communicate only incremental information about the state's change from time step $t-1$ to $t$. The relation between $\Delta g_t$ and $\mathcal{G}$ is given by
\begin{equation}
\label{eqn:update}
\mathcal{G}_t = \mathcal{G}_{t-1} \oplus \Delta g_t,
\end{equation}
where $\oplus$ is a graph operation function that produces the new belief graph $\mathcal{G}_t$ given $\mathcal{G}_{t-1}$ and $\Delta g_t$.
We formulate the graph operation function $\oplus $ using a recurrent neural network (e.g., a GRU~\citep{cho2014gru}) as:
\begin{equation}
\begin{aligned}
\label{eqn:continuous_graph_updater}
\Delta g_t &= \mathrm{f_\Delta}(h_{\mathcal{G}_{t-1}}, h_{O_t}, h_{A_{t-1}}); \\
h_t &= \rnn(\Delta g_t, h_{t-1});\\
\mathcal{G}_{t} &= \mathrm{f_{d}}(h_t). \\
\end{aligned}
\end{equation}
The function $\mathrm{f_\Delta}$ aggregates the information in $\mathcal{G}_{t-1}$, $A_{t-1}$, and $O_t$ to generate the graph update $\Delta g_t$.
$h_{\mathcal{G}_{t-1}}$ denotes the representation of $\mathcal{G}_{t-1}$ from the graph encoder.
$h_{O_t}$ and $h_{A_{t-1}}$ are outputs of the text encoder (refer to Figure~\ref{fig:agent_diagram}, left part).
The vector $h_t$ is a recurrent hidden state from which we decode the adjacency tensor $\mathcal{G}_{t}$; $h_t$ acts as a memory that carries information across game steps---a crucial function for solving POMDPs \cite{hausknecht15drqn}.
The function $\mathrm{f_d}$ is a multi-layer perceptron (MLP) that decodes the recurrent state $h_t$ into a real-valued adjacency tensor (i.e., the belief graph $\mathcal{G}_{t}$).
We elaborate on each of the sub-modules in Appendix~\ref{appendix:model}.

\textbf{Training the Graph Updater:}
We pre-train the graph updater using two self-supervised training regimes to learn structured game dynamics. After pre-training, the graph updater is fixed during \ours's interaction with games; at this time it provides belief graphs $\mathcal{G}$ to the action selector.
We train the action selector subsequently via RL.
Both pre-training tasks share the same goal: to ensure that $\mathcal{G}_t$ 
% inferred from the agent's history belief and the new observations $O_t$,
% produce the text observation $O_t$ using the agent's belief graph $\mathcal{G}_t$.
% The motivation being, in order to produce $O_t$, $\mathcal{G}_t$ should
encodes sufficient information about the environment state at game step $t$.
% We pre-train the graph updater in a self-supervised fashion.
For training data, we gather a collection of transitions by following walkthroughs in \ftwp games.\footnote{This is an independent and unique set of TextWorld games \citep{trischler19ftwp}. Details are provided in Appendix~\ref{appendix:ftwp}.}
%of the \ftwp dataset \cite{trischler19ftwp}.
To ensure variety in the training data, we also randomly sample trajectories off the optimal path.
Next we describe our pre-training approaches for the graph updater.
% The two tasks share the same goal: to ensure that $\mathcal{G}_t$, inferred from the agent's history belief and the new observations $O_t$,
% % produce the text observation $O_t$ using the agent's belief graph $\mathcal{G}_t$.
% % The motivation being, in order to produce $O_t$, $\mathcal{G}_t$ should
% encodes sufficient information about the environmental state at game step $t$.
%The two tasks however differ in their downstream objectives:

\textbf{$\bullet$\quad Observation Generation (OG):}
Our first approach to pre-train the graph updater involves training a decoder model to reconstruct text observations from the belief graph.
%The intuition is that we will be able to perfectly reconstruct the game observations only if the belief graph contains all the necessary information in the game state.
Conditioned on the belief graph, $\mathcal{G}_t$, and the action performed at the previous game step, $A_{t-1}$, the observation generation task aims to reconstruct $O_t=\{O_t^1,\dots,O_t^{L_{O_t}}\}$ token by token, where $L_{O_t}$ is the length of $O_t$. 
We formulate this task as a sequence-to-sequence (Seq2Seq) problem and use a transformer-based model \citep{vaswani17transformer} to generate the output sequence. 
% We use standard teacher forcing to train the Seq2Seq model.
%Specifically, conditioned on $\mathcal{G}_t$ and $A_{t-1}$, the transformer decoder is required to predict the $i^{\text{th}}$ token of $O_t$ given the previous $i-1$ ground-truth tokens.
Specifically, conditioned on $\mathcal{G}_t$ and $A_{t-1}$, the transformer decoder predicts the next token $O_t^i$ given $\{O_t^1,\dots,O_t^{i-1}\}$.
We train the Seq2Seq model using teacher-forcing to optimize the negative log-likelihood loss: % The loss function to optimize is the negative log-likelihood:% (omitting the subscript $t$ in $O_t$ for simplicity):
\begin{equation}
    %\mathcal{L}_{\text{OG}} = -\sum^{L_{O_t}}_{i=1} \log p_{\text{OG}}(O_i | O_1,...O_{i-1}, \mathcal{G}_t, A_{t-1}).
    \mathcal{L}_{\text{OG}} = -\sum^{L_{O_t}}_{i=1} \log p_{\text{OG}}(O_t^i | O_t^1,...,O_t^{i-1}, \mathcal{G}_t, A_{t-1}),
\end{equation}%$L_{O_t}$ is the number of tokens in $O_t$,
where $p_{\text{OG}}$ is the conditional distribution parametrized by the observation generation model.

\textbf{$\bullet$\quad Contrastive Observation Classification (COC):}
Inspired by the literature on contrastive representation learning \citep{vandenoord2018cpc,hjelm2019learning,velickovic2018deep,bachman2019amdim}, we reformulate OG mentioned above as a contrastive prediction task. 
We use contrastive learning to maximize mutual information between the predicted $\mathcal{G}_t$ and the text observations $O_t$. 
Specifically, we train the model to differentiate between representations corresponding to true observations $O_t$ and ``corrupted'' observations $\widetilde{O}_t$, conditioned on $\mathcal{G}_t$ and $A_{t-1}$. 
To obtain corrupted observations, we sample randomly from the set of all collected observations across our pre-training data.
We use a noise-contrastive objective 
%to maximize the mutual information $\mathcal{I}(\mathcal{G}_t , O_t )$.
and minimize the binary cross-entropy (BCE) loss given by
\begin{equation}
    \mathcal{L}_{\text{COC}} =  \frac{1}{K} \sum_{t=1}^{K} \left( \mathbb{E}_{O} \left[ \text{log} \mathcal{D}\left( h_{O_t}, h_{\mathcal{G}_t} \right) \right]  +   \mathbb{E}_{\widetilde{O}} \left[ \text{log} \left(1 -  \mathcal{D}\left( h_{\widetilde{O}_t}, h_{\mathcal{G}_t} \right) \right) \right]\right).
\end{equation}
Here, $K$ is the length of a trajectory as we sample a positive and negative pair at each step and $\mathcal{D}$ is a {\em discriminator} that differentiates between positive and negative samples. 
%and paired with $\mathcal{G}$ obtained from $O$ to form negative samples.
The motivation behind contrastive unsupervised training is that one does not require to train complex decoders. 
Specifically, compared to OG, the COC's objective relaxes the need for learning syntactical or grammatical features and allows GATA to focus on learning the semantics of the $O_t$.

We provide further implementation level details on both these self-supervised objectives in Appendix~\ref{appendix:pretrain_continuous}.

\subsection{Action Selector}
\label{subsection:action_selection}

The graph updater discussed in the previous section defines a key component of \ours that enables the model to maintain a structured belief graph based on text observations.
The second key component of \ours\ is the {\em action selector}, which uses the belief graph $\mathcal{G}_t$ and the text observation $O_t$ at each time-step to select an action. 
As shown in Figure~\ref{fig:agent_diagram}, the action selector consists of four main components: 
the {\em text encoder} and {\em graph encoder} convert text inputs and graph inputs, respectively, into hidden representations; a {\em representation aggregator} fuses the two representations using an attention mechanism; and a {\em scorer} ranks all candidate actions based on the aggregated representations.
%In some experimental settings, the text observation $O_t$ can be absent. In these cases, the representation aggregator is disabled and the output of the graph encoder is fed directly into the scorer. Next, we briefly describe the components used in both the action selector and the graph updater:
%Due to limited space, we describe the high level properties of each component in this subsection and give detailed explanation in Appendix~\ref{appendix:model}. 

\textbf{$\bullet$\quad Graph Encoder:~} 
\ours's belief graphs, which estimate the true game state, are multi-relational by design. Therefore, we use relational graph convolutional networks (R-GCNs)~\citep{schlichtkrull2018rgcn} to encode the belief graphs from the updater into vector representations.
We also adapt the R-GCN model to use embeddings of the available relation labels, so that we can capture semantic correspondences among relations (e.g., \cmd{east\_of} and \cmd{west\_of} are reciprocal relations).
% Since we assume access to a vocabulary of relation types, we also adapt the R-GCN to leverage embeddings of relation labels. The goal is to capture semantic correspondences among relations in these embeddings (e.g., \cmd{east\_of} and \cmd{west\_of} are reciprocal relations).
We do so by learning a vector representation for each relation in the vocabulary that we condition on the word embeddings of the relation's name. We concatenate the resulting vector with the standard node embeddings during R-GCN's message passing phase. 
%(See Appendix~\ref{appendix:pretrain_scores} for an ablation study).
%We use the concatenation of the resulting vector with the node embeddings as the input to each R-GCN layer (an ablation study for relation embeddings is provided in Appendix~\ref{appendix:pretrain_scores}).
% R-GCNs use relation-specific transformations for message passing between the neighbouring nodes at all the layers. 
%A standard R-GCN does not consider semantic information contained in the relation labels.
%In our task, however, relation labels contain useful information (e.g., \cmd{east\_of} and \cmd{west\_of} are reciprocal relations).
%Therefore, we learn a vector representation for each relation conditioned on the text label's word embeddings.
%We use the concatenation of the resulting vector with the node embeddings as the input to each R-GCN layer (an ablation study for relation embeddings is provided in Appendix~\ref{appendix:pretrain_scores}).
Our R-GCN implementation uses basis regularization \citep{schlichtkrull2018rgcn} and highway connections \citep{srivastava15highway} between layers for faster convergence. 
Details are given in Appendix~\ref{appd:graph_encoder}. 

\textbf{$\bullet$\quad Text Encoder:~} 
%Transformer models~\citep{vaswani17transformer} achieve strong performance on a host of NLP tasks~\citep{devlin2018bert, yu18qanet}.
We adopt a transformer encoder \citep{vaswani17transformer} to convert text inputs from $O_t$ and $A_{t-1}$ into contextual vector representations. 
Details are provided in Appendix~\ref{appd:text_encoder}.
%Specifically, the transformer converts the word embeddings of a list of $N$ tokens into a list of $N$ real-valued vectors, where each of the vectors is conditioned on all other tokens.

\textbf{$\bullet$\quad Representation Aggregator:~} 
To combine the text and graph representations, \ours uses a bi-directional attention-based aggregator~\citep{yu18qanet,seo2016bidaf}.
Attention from text to graph enables the agent to focus more on nodes that are currently observable, which are generally more relevant;
% Attention from text to graph enables the agent to focus more on graph nodes that appear in the current text observation, which are generally more relevant at the given time step;
attention from nodes to text enables the agent to focus more on tokens that appear in the graph, which are therefore connected with the player in certain relations.
% attention from nodes to text enables the agent to focus on text representations whose underlying tokens appear in the graph, which are therefore connected with the player in certain relations.
Details are provided in Appendix~\ref{appd:rep_aggregator}.

\textbf{$\bullet$\quad Scorer:~} 
The scorer consists of a self-attention layer cascaded with an MLP layer.
First, the self-attention layer reinforces the dependency of every token-token pair and node-node pair in the aggregated representations.
The resulting vectors are concatenated with the representations of action candidates $C_t$ (from the text encoder), after which the MLP generates a single scalar for every action candidate as a score.
% The MLP then uses the concatenated vectors to generate a single scalar for every action candidate.
% used to select the best action.
Details are provided in Appendix~\ref{appd:scorer}.

\textbf{Training the Action Selector:}
We use Q-learning~\citep{watkins1992qlearning} to optimize the action selector on reward signals from the training games. 
Specifically, we use Double DQN~\citep{Hasselt2015DeepRL} combined with multi-step learning~\citep{sutton1988} and prioritized experience replay~\citep{schaul2016replay}. 
To enable \ours to scale and generalize to multiple games, we adapt standard deep
Q-Learning by sampling a new game from the set of training games to collect an episode.
Consequently, the replay buffer contains transitions from episodes of different games.
% Q-Learning by sampling a new game $x$ from the set of training games $\mathcal{X}$ to collect an episode, meaning that the replay buffer $B$ contains transitions from episodes on different games. 
We provide further details on this training procedure in Appendix~\ref{appendix:implementation_details_action_selector}. 

\subsection{Variants Using Ground-Truth Graphs}
\label{subsection:gata_with_gt}

In \ours, the belief graph is learned entirely from text observations. 
However, the TextWorld API also provides access to the underlying graph states for games, in the format of discrete KGs.
Thus, for comparison, we also consider two models that learn from or encode ground-truth graphs directly.

\textbf{\oursgtp: Pre-training a \emph{discrete} graph updater using ground-truth graphs.}
We first consider a model that uses ground-truth graphs to pre-train the graph updater, in lieu of self-supervised methods.
\oursgtp uses ground-truth graphs from \ftwp during pre-training, but infers belief graphs from the raw text 
% without ground-truth input 
during RL training of the action selector to compare fairly against \ours.
Here, the belief graph $\mathcal{G}_t$ is a discrete multi-relational graph.
%where both the entity and relation vocabularies are grounded to the ground truth extracted from the TextWorld game state. 
To pre-train a discrete graph updater, we adapt the command generation approach proposed by \citet{zelinka2019building}. 
%We optimize a Seq2Seq model to generate a sequence of commands (e.g., \cmd{add node1 node2 relation}) based on the text observations to update the discrete graph.
We provide details of this approach in Appendix~\ref{appendix:gatagtp}.

\textbf{\oursgtf: Training the action selector using ground-truth graphs.}
To get a sense of the upper bound on performance we might obtain using a belief graph, we also train an agent that uses the full ground-truth graph $\mathcal{G}^\text{full}$ during action selection.
This agent requires no graph updater module; we simply feed the ground-truth graphs into the action selector (via the graph encoder).
The use of ground-truth graphs allows \oursgtf to escape the error cascades that may result from inferred belief graphs. 
Note also that the ground-truth graphs contain full state information, relaxing 
% the difficulty introduced by
partial observability of the games.
% Consequently, we expect a more effective reward optimization process for \oursgtf, compared to all other agents --- this will also verify our hypothesis that structured representations facilitate effective and general policies.
Consequently, we expect more effective reward optimization for \oursgtf compared to other graph-based agents. \oursgtf's comparison with text-based agents is a sanity check for our hypothesis---that structured representations help learning general policies.
% Note that the ground-truth KGs contain information that the agent may not have encountered until now, thus relaxing the issue of partial observability. 
% As a result, we expect reward optimization for \oursgtf to be fairly easier than other graph-based agents. 
% However, effectiveness of \oursgtf compared with the text-only agents is crucial towards the hypothesis of this work--structured representations facilitate effective and general policies.
% During both the training and testing phases, the input graph to the action selector at each time step is the direct adjacency-tensor representation of the full game state. 
% Note that this also includes elements of the environment that the agent does not observe at the current time step, removing the issue of partial observability. 

\section{Experiments and Analysis}
\label{section:exp}

We conduct experiments on generated text-based games (Section~\ref{section:background}) to answer two key questions:\\
\textbf{Q1:} Does the belief-graph approach aid \ours in achieving high rewards on unseen games after training? In particular, does  \ours\ improve performance compared to SOTA text-based models? \\
% \textbf{Q1:} Does \ours\ improve over SOTA text-based models, including those with recurrent policy?\\
\textbf{Q2:} How does \ours compare to models that have access to ground-truth graph representations?

% Note that for all our experiments we train on a set of multiple training games and evaluate them on a test set of games. This is done to avoid policies from memorizing the game configuration and rather learn game dynamics as opposed to other works which train and validate on the same games.
%We conduct experiments on cooking themed text-based games (as described in Section~\ref{section:background}) that we generate with TextWorld~\citep{cote18textworld}.
%Our goal is to answer two key questions:
% \begin{enumerate}[label=\textbf{Q\arabic*}, itemsep=0pt, parsep=0pt, topsep=0pt, leftmargin=*]
%     \item Does \ours\ improve over SOTA text-based models, including those with recurrent policy?
%     %\item Does the graph-structured belief state in \ours\ provide improvements compared to state-of-the-art text-based models, including models that use recurrent hidden states?
%     \item How does \ours compare to models that have access to ground-truth graphs representations?
%     %\item How does the performance of \ours 
%     % which learns belief graphs directly from text observations,
%     %compare to models that have access to ground-truth graphs representations?
%     % \item Which components of \ours\ are most essential to achieve strong performance? 
% \end{enumerate}

\subsection{Experimental Setup and Baselines}
\label{subsection:baselines}

We divide the games into four subsets with one difficulty level per subset.
%Each subset contains 100 training games, 20 validation games, and 20 test games, which are sampled from a distribution determined by their difficulty level.
Each subset contains 100 training, 20 validation, and 20 test games, which are sampled from a distribution determined by their difficulty level.
% \ash{Evaluating agents on such a dataset of games (per difficulty level) is uncommon in reinforcement learning and specifically in the domain of text-based games--most previous works train and test on a single instance of games. However, doing so allows us to check robustness and generalization nature of the learnt policies. Throughout the course of the paper, we use the term ``generalization'' to imply the ability of a single policy to play a distribution of games defined by their difficulty.}
To elaborate on the diversity of games: for easier games, the recipe might only require a single ingredient and the world is limited to a single location, whereas harder games might require an agent to navigate a map of 6 locations to collect and appropriately process up to three ingredients.
% whereas harder games might have a recipe requiring up to three ingredients randomly distributed across a map of 6 locations. 
We also test \ours's transferability across difficulty levels by mixing the four difficulty levels to build level 5. 
% To make the number of training games consistent with the other difficulty levels, we sample 25 games from each of the four difficulty levels.
We sample 25 games from each of the four difficulty levels to build a training set.
% However, we use all validation and test games (80 games each) as the validation and test sets for level 5.
% See Table~\ref{tab:stats} for detailed statistical characteristics of the game sets.
We use all validation and test games from levels 1 to 4 for level 5 validation and test.
In all experiments, we select the top-performing agent on validation sets and report its test scores; all validation and test games are unseen in the training set.
Statistics of the games are shown in Table~\ref{tab:stats}.
%See Table~\ref{tab:stats} for detailed statistical characteristics of the game sets.

%\begin{wraptable}{r}{0.5\textwidth}
\begin{table}[h!]
    % 1.  tw-cooking-recipe1+take1+open+train
    % 2.  tw-cooking-recipe1+take1+cook+open+train
    % 3.  tw-cooking-recipe1+take1+cut+open+train
    % 4.  tw-cooking-recipe1+take1+go6+open+train
    % 5.  tw-cooking-recipe1+take1+go9+open+train
    % 6.  tw-cooking-recipe1+take1+go12+open+train
    % 7.  tw-cooking-recipe1+take1+cook+cut+open+train
    % 8.  tw-cooking-recipe3+take3+go6+open+train
    % 9.  tw-cooking-recipe3+take3+go6+cook+cut+open+train
    % 10. tw-cooking-recipe3+take3+cook+cut+open+go12+train
    %\small
    \scriptsize
    \caption{Games statistics (averaged across all games within a difficulty level).} 
    % \#I: number of ingredients per recipe. \#L: number of locations. S: max score. \#C: average number of action candidates per step. \#O: average number of objects per game.
    \label{tab:stats}
    \centering
    \vspace{0.8em}
    % \begin{tabular}{c|c|c|c|c|c|c|c}
    \begin{tabular}{cccccccc}
        \toprule
        Level  & Recipe Size & \#Locations & Max Score   & Need Cut & Need Cook  & \#Action Candidates & \#Objects \\
        \midrule
        1   & 1 & 1     &  4  & \cmark & \xmark  & 11.5 & 17.1 \\  % difficulty level 3
        2   & 1 & 1     &  5  & \cmark & \cmark  & 11.8 & 17.5 \\  % difficulty level 7
        3   & 1 & 9     &  3  & \xmark & \xmark  & 7.2 & 34.1 \\  % difficulty level 5
        4   & 3 & 6     &  11 & \cmark & \cmark  & 28.4 & 33.4 \\  % difficulty level 9
        \midrule
        5   & \multicolumn{7}{c}{Mixture of levels \{1,2,3,4\}} \\ 
        \bottomrule
    \end{tabular}
\end{table}
%\end{wraptable}

% Note we focus on training and testing agents on multiple games, where the term \textit{multiple games} refers to a set of games with the same difficulty level; we do not consider cross-difficulty level learning or evaluation.

% \subsection{Baselines}
% \label{subsection:baselines}
%Our baselines include agents which process $O_t$ directly for action selection; agents which have access to ground truth $G_t$; and agents which extract structured information from $O_t$ to act optimally.
As baselines, we use our implementation of LSTM-DQN~\citep{narasimhan15lstmdqn} and LSTM-DRQN~\citep{yuan2018counting}, both of which use only $O_t$ as input. 
Note that LSTM-DRQN uses an RNN to enable an implicit memory (i.e., belief); it also uses an episodic counting bonus to encourage exploration~\citep{yuan2018counting}.
This draws an interesting comparison with \ours, wherein the belief is extracted and updated dynamically, in the form of a graph.
% For fair comparison, we replace the LSTM-based text encoder of both LSTM-DQN and LSTM-DRQN with a transformer-based text encoder as in \ours.
For fair comparison, we replace the LSTM-based text encoders with a transformer-based text encoder as in \ours.
We denote those agents as \transdqn and \transdrqn respectively.
We denote a \transdrqn equipped with the episodic counting bonus as \transdrqnp. These three text-based baselines are representative of the current top-performing neural agents on text-based games.

Additionally, we test the variants of \ours that have access to ground-truth graphs (as described in Section~\ref{subsection:gata_with_gt}).
Comparing with \ours, the \oursgtp agent also maintains its belief graphs throughout the game; however, its graph updater is pre-trained on \ftwp using ground-truth graphs---a stronger supervision signal.
\oursgtf, on the other hand, does not have a graph updater. It directly uses ground-truth graphs as input during game playing. 

\begin{table}[t!]
    \scriptsize
    \centering
    \caption{Agents' normalized \textbf{test} scores and averaged relative improvement ($\%\uparrow$) over \transdqn across difficulty levels. An agent $\text{m}$'s relative improvement over \transdqn is defined as $(\text{R}_{\text{m}} - \text{R}_{\text{Tr-DQN}}) / \text{R}_{\text{Tr-DQN}}$ where R is the score. All numbers are percentages. \sdiamond represents ground-truth full graph; \sclub represents discrete $\mathcal{G}_t$ generated by \oursgtp; \sspade represents $O_t$. \sstar and \sinfinite are continuous $G_t$ generated by \ours, when the graph updater is pre-trained with OG and COC tasks, respectively.}
    \label{tab:test_score}
    \vspace{0.8em}
    % \begin{tabular}{c||c|c|c|c|c|c||c|c|c|c|c|c||c}
    \begin{tabular}{c|cccccc|cccccc|c}
        \toprule
         & \multicolumn{6}{c|}{20 Training Games} & \multicolumn{6}{c|}{100 Training Games} & Avg. \\
        \midrule
        Difficulty Level & 1   & 2  & 3   &  4 & 5 & $\%\uparrow$ & 1   & 2  & 3   &  4 & 5 & $\%\uparrow$  & $\%\uparrow$\\
        \midrule   
        Agent & \multicolumn{13}{c}{Text-based Baselines} \\
        \midrule 
        \transdqn         & 66.2 & 26.0 & 16.7 & 18.2 & \textbf{27.9} & ----- & 62.5 & 32.0 & 38.3 & 17.7 & 34.6 & ---- & ----\\
        \midrule   
        \transdrqn        & 62.5 & 32.0 & 28.3 & 12.7 & 26.5 & +10.3 & 58.8 & 31.0 & 36.7 & 21.4 & 27.4 & -2.6 & +3.9 \\
        \midrule   
        \transdrqnp       & 65.0 & 30.0 & 35.0 & 11.8 & 18.3 & +10.7 & 58.8 & 33.0 & 33.3 & 19.5 & 30.6 & -3.4 & +3.6  \\
        \midrule    
        Input & \multicolumn{13}{c}{\ours} \\
        \midrule 
        \sstar            & 70.0 & 20.0 & 20.0 & 18.6 & 26.3 & -0.2  & 62.5 & 32.0 & 46.7 & \textbf{27.7} & 35.4 & \textbf{+16.1} & +8.0  \\
        \midrule 
        \sstar \sspade    & 66.2 & \textbf{48.0} & 26.7 & 15.5 & 26.3 & +24.8 & \textbf{66.2} & \textbf{36.0} & \textbf{58.3} & 14.1 & \textbf{45.0} & \textbf{+16.1} & +20.4  \\
        \midrule 
        \sinfinite        & \textbf{73.8} & 42.0 & 26.7 & \textbf{20.9} & 24.5 & +27.1 & 62.5 & 30.0 & 51.7 & 23.6 & 36.0 & +13.2 & +20.2  \\
        \midrule 
        \sinfinite \sspade& 68.8 & 33.0 & \textbf{41.7} & 17.7 & 27.0 & \textbf{+34.9} & 62.5 & 33.0 & 46.7 & 25.9 & 33.4 & +13.6 & \textbf{+24.2}  \\
        \midrule
         & \multicolumn{13}{c}{\oursgtp} \\
        \midrule 
        \sclub            & 56.2 & 26.0 & 40.0 & 17.3 & 17.7 & +16.6 & 37.5 & 31.0 & 45.0 & 13.6 & 18.7 & -18.9 & -1.2  \\
        \midrule 
        \sclub \sspade    & 65.0 & 32.0 & 41.7 & 12.3 & 23.5 & +24.6 & 62.5 & 32.0 & 51.7 & 21.8 & 23.5 & +5.2 & +14.9  \\
        \midrule 
         & \multicolumn{13}{c}{\oursgtf} \\
        \midrule   
        \sdiamond         & 48.7 & 61.0 & 46.7 & 23.6 & 28.9 & +64.2 & 95.0 & 95.0 & 70.0 & 37.3 & 52.8 & +99.0 & +81.6 \\
        \bottomrule
    \end{tabular}
\end{table}

\subsection*{Q1: Performance of \ours compared to text-based baselines}
\label{subsection:gata_vs_baseline}

In Table~\ref{tab:test_score}, we show the normalized test scores achieved by agents trained on either 20 or 100 games for each difficulty level.
%We observe that GATA, equipped with the belief graphs, significantly outperforms all text-based baselines. The Tr-DRQN and Tr-DRQN+ baselines outperform Tr-DRQN with 3.9\% and 3.6\% relative improvement ($\%\uparrow$), highlighting the effectiveness of recurrent components.
Equipped with belief graphs, \ours significantly outperforms all text-based baselines.
The graph updater pre-trained on both of the self-supervised tasks (Section~\ref{subsection:graph_update}) leads to better performance than the baselines (\sstar~and \sinfinite).
We observe further improvements in \ours's policies when the text observations (\sspade) are also available. 
% We believe this is because the text observations along with the attention mechanism in the representations aggregator helps \ours to focus onto currently observable objects.
% This attention may further help \ours to counteract error accumulation in the belief graphs.
We believe the text observations guide \ours's action scorer to focus on currently observable objects through the bi-attention mechanism. 
The attention may further help \ours to counteract accumulated errors from the belief graphs.
In addition, we observe that \transdrqn and \transdrqnp outperform \transdqn, with 3.9\% and 3.6\% relative improvement ($\%\uparrow$). 
This suggests the implicit memory of the recurrent components improves performance.
% However, when the text observations (\sspade) are also available, \ours is improved significantly and achieves its best performance.
% Regarding this clear gap, we hypothesize that direct text observations, combined with \ours's attention mechanisms in the aggregator, guide the agent's focus onto objects that are currently observable, this may help \ours to counteract error accumulation in the belief graphs.
% We also observe that, compared to the two DRQN variants, \ours has an obvious advantage when the training set is larger.
We also observe \ours substantially outperforms \transdqn when trained on 100 games, whereas the DRQN agents struggle to optimize rewards on the larger training sets.
% Both DRQN agents have negative relative improvements when trained on 100 games, whereas all \ours variants achieve significantly higher scores over \transdqn in this setting.

\subsection*{Q2: Performance of \ours compared to models with access to the ground-truth graph}
\label{subsection:gata_vs_gatagt}
Table~\ref{tab:test_score} also reports test performance for \oursgtp (\sclub)  and \oursgtf (\sdiamond). 
Consistent with \ours, we find \oursgtp also performs better when given text observations (\sspade) as additional input to the action scorer.
Although \oursgtp outperforms \transdqn by 14.9\% when text observations are available, its overall performance is still substantially poorer than \ours. 
Although the graph updater in \oursgtp is trained with ground-truth graphs, we believe the discrete belief graphs and the discrete operations for updating them (Appendix~\ref{appendix:discrete_graph_updater}) make this approach vulnerable to an accumulation of errors over game steps, as well as errors introduced by the discrete nature of the predictions (e.g., round-off error).
%Furthermore, errors across game steps accumulate more easily within discrete channels (in \oursgtp's case, the belief graph).
%Despite having a graph updater trained with ground-truth graphs, we believe the discrete belief graphs generated by \oursgtp, as well as the discrete operations for updating discrete graphs (i.e., \cmd{add} and \cmd{delete} commands as described in Appendix~\ref{appendix:discrete_graph_updater}), make this approach vulnerable to error accumulation and optimization difficulties.
In contrast, we suspect that the continuous belief graph and the learned graph operation function (Eqn.~\ref{eqn:continuous_graph_updater}) are easier to train and recover more gracefully from errors.

Meanwhile, \oursgtf, which uses ground-truth graphs $\mathcal{G}^\text{full}$ during training and testing, obtains significantly higher scores than does \ours and all other baselines.
% GATA-GTF, with access to ground-truth graphs $\mathcal{G}^\text{full}$ during reward optimization, achieves substantially higher rewards than \ours and all other baselines. 
Because $\mathcal{G}^\text{full}$ turns the game environment into a fully observable MDP and encodes accurate state information with no error accumulation, \oursgtf represents the performance upper-bound of all the $\mathcal{G}_t$-based baselines.
The scores achieved by \oursgtf reinforce our intuition that belief graphs improve text-based game agents.
At the same time, the performance gap between \ours and \oursgtf invites investigation into better ways to learn accurate graph representations of text. 

% Table~\ref{tab:test_score} also reports test performance for \oursgtp  and \oursgtf. 
% Consistent with \ours, we find \oursgtp also performs better when using text observations as additional input to the action scorer.
% Although the variant of \oursgtp with text observations outperforms all the text-based baselines, its overall performance is substantially poorer vis-\`a-vis its unsupervised counterparts. 
% This seeming surprising result can be explained by the low margin for error that sparse 0-1 adjacency tensors provide when for \oursgtp graph updater. 

% On the other hand, delivering to the expectations of an upper-bound, \oursgtf which relies totally on ground-truth graphs even during reward optimization exceedingly outperforms all the other baselines in Table~\ref{tab:test_score} with a clear reason of not accumulating any errors from the observation representations. While this huge difference between the scores achieved by \oursgtf reinforces our intuition of using graph representation learning for optimizing rewards for text-based games; it also demands much deeper investigation in the topic by the community. 

\begin{figure}[t!]
\centering
\begin{subfigure}{.5\textwidth}
  \centering
  \includegraphics[width=.95\linewidth]{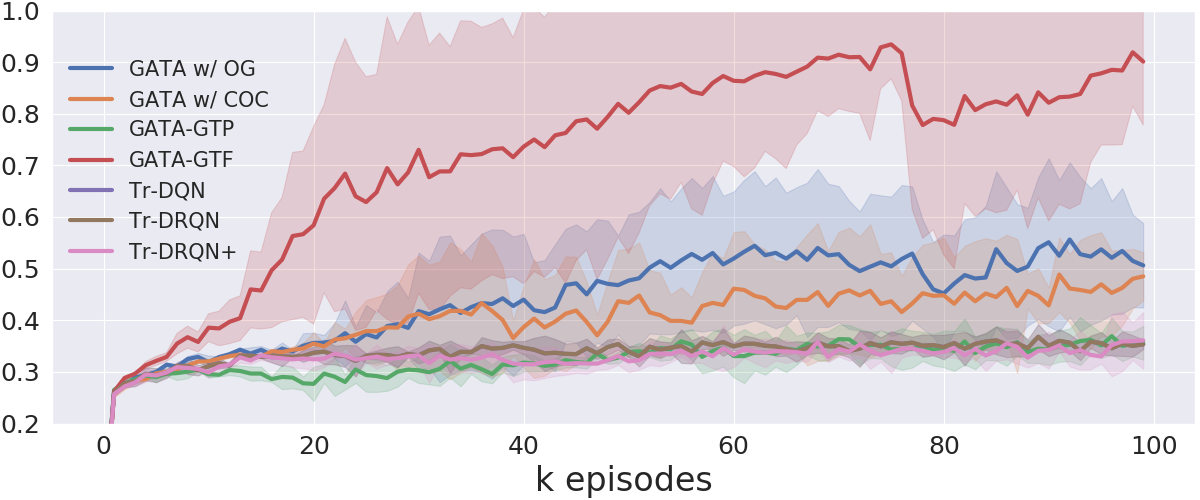}
\end{subfigure}%
\begin{subfigure}{.5\textwidth}
  \centering
  \includegraphics[height=3cm,trim={0 0 5cm 0},clip]{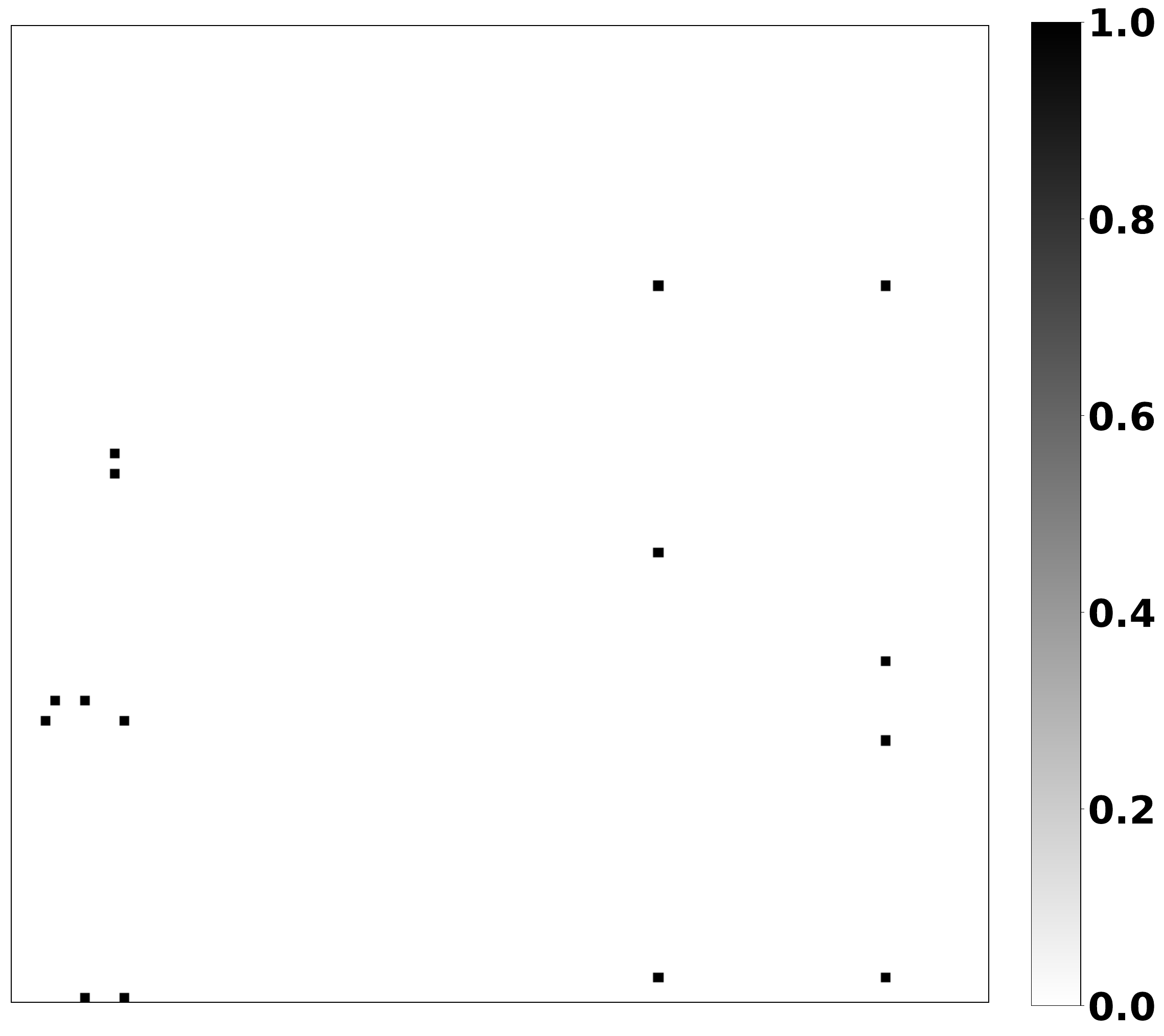}
  \includegraphics[height=3cm,trim={0 0 0 0},clip]{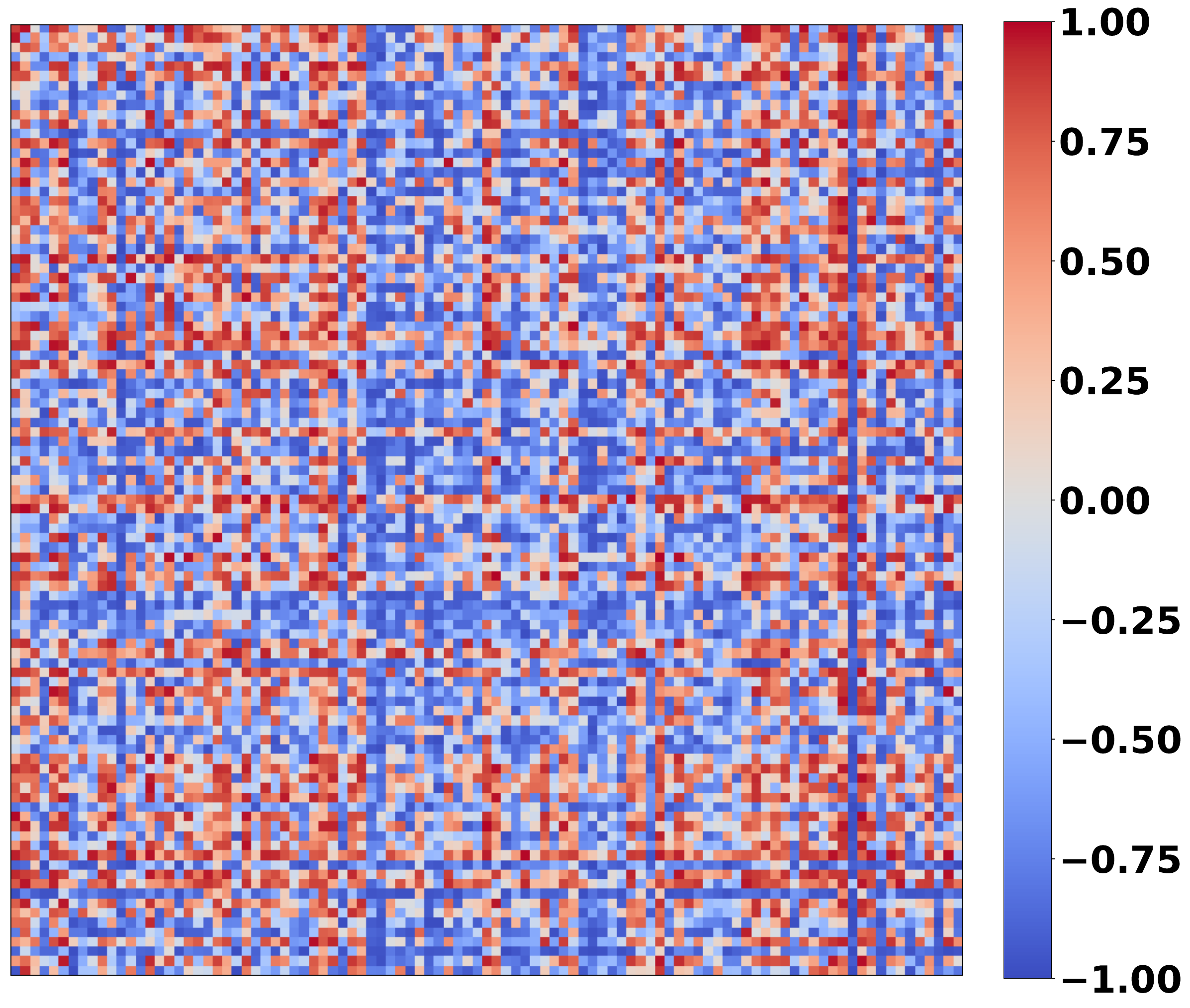}
\end{subfigure}
\caption{\textbf{Left:} Training curves on 20 level 2 games (averaged over 3 seeds).
% --normalized score along y-axis; 1000 episodes along x-axis.
%\textbf{Mid, Right:} A ground-truth graph and a belief graph $\mathcal{G}$ generated by the COC pre-training procedure.}
\textbf{Right:} Density comparison between a ground-truth graph (binary) and a belief graph $\mathcal{G}$ generated by the COC pre-training procedure. Both matrices are slices of adjacency tensors corresponding the \cmd{is} relation.}
\label{fig:training_curve_and_qualitative_analysis}
\end{figure}

% \subsection*{Q3: Ablations and Additional Results}
\subsection*{Additional Results}
We also show the agents' training curves and examples of the belief graphs $\mathcal{G}$ generated by \ours.
Figure~\ref{fig:training_curve_and_qualitative_analysis} (\textbf{Left}) shows an example of all agents' training curves.
We observe consistent trends with the testing results of Table~\ref{tab:test_score} --- \ours outperforms the text-based baselines and \oursgtp, but a significant gap exists between \ours and \oursgtf (which uses ground-truth graphs as input to the action scorer).
Figure~\ref{fig:training_curve_and_qualitative_analysis} (\textbf{Right}) highlights the sparsity of a ground-truth graph compared to that of a belief graph $\mathcal{G}$. % generated by \ours.
Since generation of $\mathcal{G}$ is unsupervised by any ground-truth graphs, we do not expect $\mathcal{G}$ to be interpretable nor sparse.
Further, since the self-supervised models learn belief graphs directly from text, some of the learned features may correspond to the underlying grammar or other features useful for the self-supervised tasks, 
% as opposed to their ground-truth counterparts. 
rather than only being indicative of relationships between objects.
%From the comparison in Figure~\ref{fig:training_curve_and_qualitative_analysis}, we can clearly see that $\mathcal{G}$ is much denser than the ground-truth graph.
% However, in Appendix~\ref{appendix:highres_visualization} we show the effectiveness of the encoded $\mathcal{G}$ using a relation prediction probing task.
However, we show $\mathcal{G}$ encodes useful information for a relation prediction probing task in Appendix~\ref{appendix:highres_visualization}.
% Nevertheless, we show in Appendix~\ref{appendix:highres_visualization} that $\mathcal{G}$ can advance a relation prediction probing task.

% Figure~\ref{fig:training_curve_and_qualitative_analysis} (\textbf{Mid} and \textbf{Right}) depict an example of belief graphs generated by \ours, when pre-trained with the COC and OG approaches, respectively.
% Because both the self-supervised models learn the belief graphs directly from text, some of the learned features may correspond to the underlying grammar or other features used in the self-supervised tasks, rather than indicative of relationships between objects.
% For visualization, we found that removing the mean graph across all games and steps (which we treat as a baseline) neatly exposes a sparser view of the belief graphs.

Given space limitations, we only report a representative selection of our results in this section. Appendix~\ref{appendix:more_results} provides an exhaustive set of results including training curves, training scores, and test scores for all experimental settings introduced in this work.
We also provide a detailed qualitative analysis including hi-res visualizations of the belief graphs.
We encourage readers to refer to it.%the extended experimental results.
% We strongly encourage readers to consult the extended experimental results.

\section{Related Work}
\label{section:related}

\noindent\textbf{Dynamic graph extraction:}\quad
Numerous recent works have focused on constructing graphs to encode structured representations of raw data, for various tasks.
\citet{Kipf2020Contrastive} propose contrastive methods to learn latent structured world models (C-SWMs) as state representations for vision-based environments. Their work, however, does not focus on learning policies to play games or to generalize across varying environments.
\citet{das18dynamickg} leverage a machine reading comprehension mechanism to query for entities and states in short text passages and use a dynamic graph structure to track changing entity states.
\citet{fan19kg} propose to encode graph representations by linearizing the graph as an input sequence in NLP tasks.
\citet{johnson2016learning} construct graphs from text data using gated graph transformer neural networks. 
\citet{Yang2018GLoMoUL} learn transferable latent relational graphs from raw data in a self-supervised manner. 
Compared to the existing literature, our work aims to infer multi-relational KGs dynamically from partial text observations of the state and subsequently use these graphs to inform general policies.
Concurrently, \citet{srinivas2020curl} propose to learn state representations with contrastive learning methods to facilitate RL training.
However, they focus on vision-based environments and they do not investigate generalization.

More generally, we want to note that compared to traditional knowledge base construction (KBC) works, our approach is more related to the direction of neural relational inference \citep{kipf2018neural}. 
In particular, we seek to generate task-specific graphs, which tend to be dynamic, contextual and relatively small, whereas traditional KBC focus on generating large, static graphs.

\noindent\textbf{Playing Text-based Games:}\quad
Recent years have seen a host of work on playing text-based games.
Various deep learning agents have been explored~\citep{narasimhan15lstmdqn, he15drrn, hausknecht19jericho,zahavy18nottolearn,jain19algo,ammanabrolu2020graph,Zelinka2018, yin19comprehensible}.
\citet{fulda17rock} use pre-trained embeddings to reduce the action space.
\citet{zahavy18nottolearn}, \citet{seurin19sorry}, and \citet{jain19algo} explicitly condition an agent's decisions on game feedback.
Most of this literature trains and tests on a single game without considering generalization.
\citet{urbanek19learn} use memory networks and ranking systems to tackle adventure-themed dialog tasks.
\citet{yuan2018counting} propose a count-based memory to explore and generalize on simple unseen text-based games.
\citet{madotto2020exploration} use GoExplore \cite{Ecoffet2019GoExploreAN} with imitation learning to generalize.
\citet{adolphs19ledeepchef} and \citet{yin19learn} also investigate the multi-game setting.
These methods rely either on reward shaping by heuristics, imitation learning, or rule-based features as inputs.
We aim to minimize hand-crafting, so our action selector is optimized only using raw rewards from games while other components of our model are pre-trained on related data.
Recently, \citet{ammanabrolu19graph,ammanabrolu2020graph, yin19learn} leverage graph structure by using rule-based, untrained mechanisms to construct KGs to play text-based games.

\section{Conclusion}
\label{section:conclusion}

In this work, we investigate how an RL agent can play and generalize within a distribution of text-based games using graph-structured representations inferred from text.
We introduce \ours, a novel neural agent that infers and updates latent belief graphs as it plays text-based games.
We use a combination of RL and self-supervised learning to teach the agent to encode essential dynamics of the environment in its belief graphs. 
We show that \ours achieves good test performance, outperforming a set of strong baselines including agents pre-trained with ground-truth graphs.
This evinces the effectiveness of generating graph-structured representations for text-based games.

\section{Broader Impact}
\label{section:ethicla_discussion}

%We do not foresee immediate societal consequences from our work, which focuses on text-based games.
Our work's immediate aim---improved performance on text-based games---might have limited consequences for society; however, taking a broader view of our work and where we'd like to take it forces us to consider several social and ethical concerns.
We use text-based games as a proxy to model and study the interaction of machines with the human world, through language.
Any system that interacts with the human world impacts it.
As mentioned previously, an example of language-mediated, human-machine interaction is online customer service systems.
\begin{itemize}
    \item In these systems, especially in products related to critical needs like healthcare, providing inaccurate information could result in serious harm to users. Likewise, failing to communicate clearly, sensibly, or convincingly might also cause harm. It could waste users' precious time and diminish their trust. 
    \item The responses generated by such systems must be inclusive and free of bias. They must not cause harm by the act of communication itself, nor by making decisions that disenfranchise certain user groups. Unfortunately, many data-driven, free-form language generation systems currently exhibit bias and/or produce problematic outputs.
    \item Users' privacy is also a concern in this setting. Mechanisms must be put in place to protect it. Agents that interact with humans almost invariably train on human data; their function requires that they solicit, store, and act upon sensitive user information (especially in the healthcare scenario envisioned above).
    Therefore, privacy protections must be implemented throughout the agent development cycle, including data collection, training, and deployment.
    \item Tasks that require human interaction through language are currently performed by people. As a result, advances in language-based agents may eventually displace or disrupt human jobs. This is a clear negative impact.
\end{itemize}

Even more broadly, any systems that generate convincing natural language could be used to spread misinformation.

Our work is immediately aimed at improving the performance of RL agents in text-based games, in which agents must understand and act in the world through language. Our hope is that this work, by introducing graph-structured representations, endows language-based agents with greater accuracy and clarity, and the ability to make better decisions. Similarly, we expect that graph-structured representations could be used to constrain agent decisions and outputs, for improved safety. Finally, we believe that structured representations can improve neural agents' interpretability to researchers and users. This is an important future direction that can contribute to accountability and transparency in AI. As we have outlined, however, this and future work must be undertaken with awareness of its hazards.

\section{Acknowledgements}
We thank Alessandro Sordoni and Devon Hjelm for the helpful discussions about the probing task.
We also thank 
David Krueger, 
Devendra Singh Sachan, 
Harm van Seijen, 
Harshita Sahijwani, 
Jacob Miller, 
Koustuv Sinha, 
Loren Lugosch, 
Meng Qu, 
Travis LaCroix, 
and the anonymous ICML 2020 and NeurIPS 2020 reviewers and ACs for their insightful comments on an earlier draft of this work. The work was funded in part by an academic grant from Microsoft Research, an NSERC Discovery Grant RGPIN-2019-05123, an IVADO Fundamental Research Project Grant PRF-2019-3583139727, as well as Canada CIFAR Chairs in AI, held by Prof. Hamilton, Prof. Poupart and Prof. Tang. 

\bibliographystyle{apalike}
\bibliography{biblio}

\clearpage
\appendix

\textbf{\large{Contents in Appendices:}}
\begin{itemize}
    \item In Appendix~\ref{appendix:model}, we describe each of the components in \ours in detail.
    \item In Appendix~\ref{appendix:pretrain_continuous}, we provide detailed information on how we pre-train \ours's graph updater with the two proposed methods (i.e., OG and COC).
    \item In Appendix~\ref{appendix:gatagtp}, we provide detailed information on \oursgtp, the discrete version of \ours. Since the action scorer module is the same as in \ours, this appendix elaborates on how a discrete graph updater works and how to pre-train the discrete graph updater.
    \item In Appendix~\ref{appendix:more_results}, we provide additional results and discussions. This includes training curves, training scores, testing scores, and high-res examples of the belief graphs learned by \ours. We provide a set of probing experiments to show that the belief graphs learned by \ours can capture useful information for relation classification tasks. We also provide qualitative analysis on \ours's OG task, which also suggests the belief graphs contain useful information for reconstructing the text observation $O_t$.
    \item In Appendix~\ref{appendix:implementation_details}, we provide implementation details for all our experiments.
    \item In Appendix~\ref{appendix:ftwp}, we show examples of graphs in TextWorld games.
\end{itemize}

\section{Details of \ours}
\label{appendix:model}

\subsection*{Notations}
In this section, we use $O_t$ to denote text observation at game step $t$, $C_t$ to denote a list of action candidate provided by a game, and $\mathcal{G}_{t}$ to denote a belief graph that represents \ours's belief to the state.

We use $L$ to refer to a linear transformation and $L^{f}$ means it is followed by a non-linear activation function $f$. 
Brackets $[\cdot;\cdot]$ denote vector concatenation.
Overall structure of \ours is shown in Figure~\ref{fig:agent_diagram}.

\subsection{Graph Encoder}
\label{appd:graph_encoder}

As briefly mentioned in Section~\ref{subsection:action_selection}, \ours utilizes a graph encoder which is based on R-GCN \citep{schlichtkrull2018rgcn}.

To better leverage information from relation labels, when computing each node's representation, we also condition it on a relation representation $E$:
\begin{equation}
\tilde{h}_i = \sigma\left(\sum_{r\in \mathcal{R}}\sum_{j\in \mathcal{N}^r_i} W^{l}_r [h^{l}_j; E_r] + W^{l}_0 [h^{l}_i; E_r]\right),
\end{equation}
in which, $l$ denotes the $l$-th layer of the R-GCN, $\mathcal{N}^r_i$ denotes the set of neighbor indices of node $i$ under relation $r \in \mathcal{R}$, $\mathcal{R}$ indicates the set of different relations, $W^{l}_r$ and $W^{l}_0$ are trainable parameters.
Since we use continuous graphs, $\mathcal{N}^r_i$ includes all nodes (including node $i$ itself).
To stabilize the model and preventing from the potential explosion introduced by stacking R-GCNs with continuous graphs, we use Sigmoid function as $\sigma$ (in contrast with the commonly used ReLU function).

As the initial input $h^{0}$ to the graph encoder, we concatenate a node embedding vector and the averaged word embeddings of node names.
Similarly, for each relation $r$, $E_r$ is the concatenation of a relation embedding vector and the averaged word embeddings of $r$'s label.
Both node embedding and relation embedding vectors are randomly initialized and trainable.

To further help our graph encoder to learn with multiple layers of R-GCN, we add highway connections \citep{srivastava15highway} between layers:
\begin{equation}
\begin{aligned}
g &= L^{\sigm}(\tilde{h}_i), \\
h^{l+1}_i &= g \odot \tilde{h}_i + (1 - g) \odot h^{l}_i, \\
\end{aligned}
\end{equation}
where $\odot$ indicates element-wise multiplication.

We use a 6-layer graph encoder, with a hidden size $H$ of 64 in each layer.
The node embedding size is 100, relation embedding size is 32.
The number of bases we use is 3.

\subsection{Text Encoder}
\label{appd:text_encoder}

We use a transformer-based text encoder, which consists of a word embedding layer and a transformer block \citep{vaswani17transformer}.
Specifically, word embeddings are initialized by the 300-dimensional fastText \cite{mikolov18fasttext} word vectors trained on Common Crawl (600B tokens) and kept fixed during training in all settings.

The transformer block consists of a stack of 5 convolutional layers, a self-attention layer, and a 2-layer MLP with a ReLU non-linear activation function in between. 
In the block, each convolutional layer has 64 filters, each kernel's size is 5.
In the self-attention layer, we use a block hidden size $H$ of 64, as well as a single head attention mechanism.
Layernorm \citep{ba16layernorm} is applied after each component inside the block. 
Following standard transformer training, we add positional encodings into each block's input.

We use the same text encoder to process text observation $O_t$ and the action candidate list $C_t$. The resulting representations are $h_{O_t} \in \mathbb{R}^{L_{O_t} \times H}$ and $h_{C_t} \in \mathbb{R}^{N_{C_t} \times L_{C_t} \times H}$, where $L_{O_t}$ is the number of tokens in $O_t$, $N_{C_t}$ denotes the number of action candidates provided, $L_{C_t}$ denotes the maximum number of tokens in $C_t$, and $H = 64$ is the hidden size.

\subsection{Representation Aggregator}
\label{appd:rep_aggregator}

The representation aggregator aims to combine the text observation representations and graph representations together.
Therefore this module is activated only when both the text observation $O_t$ and the graph input $\mathcal{G}_t$ are provided.
In cases where either of them is absent, for instance, when training the agent with only $\mathcal{G}^{\text{belief}}$ as input, the aggregator will be deactivated and the graph representation will be directly fed into the scorer.

For simplicity, we omit the subscript $t$ denoting game step in this subsection.
At any game step, the graph encoder processes graph input $\mathcal{G}$, and generates the graph representation $h_{\mathcal{G}} \in \mathbb{R}^{N_{\mathcal{G}} \times H}$.
The text encoder processes text observation $O$ to generate text representation $h_O \in \mathbb{R}^{L_{O} \times H}$.
$N_{\mathcal{G}}$ denotes the number of nodes in the graph $\mathcal{G}$, $L_{O}$ denotes the number of tokens in $O$. 

We adopt a standard representation aggregation method from question answering literature \citep{yu18qanet} to combine the two representations using attention mechanism.

Specifically, the aggregator first uses an MLP to convert both $h_{\mathcal{G}}$ and $h_O$ into the same space, the resulting tensors are denoted as $h_{\mathcal{G}}' \in \mathbb{R}^{N_{\mathcal{G}} \times H}$ and $h_O' \in \mathbb{R}^{L_O \times H}$. Then, a trilinear similarity function \citep{seo2016bidaf} is used to compute the similarities between each token in $h_O'$ with each node in $h_{\mathcal{G}}'$.
The similarity between $i$th token in $h_O'$ and $j$th node in $h_{\mathcal{G}}'$ is thus computed by:
\begin{equation}
\text{Sim}(i, j) = W(h_{O_i}', h_{\mathcal{G}_j}', h_{O_i}' \odot h_{\mathcal{G}_j}'),
\end{equation}
where $W$ is trainable parameters in the trilinear function. 
By applying the above computation for each pair of $h_O'$ and $h_{\mathcal{G}}'$, a similarity matrix $S \in \mathbb{R}^{L_O \times N_{\mathcal{G}}}$ is resulted.

Softmax of the similarity matrix $S$ along both dimensions (number of nodes $N_{\mathcal{G}}$ and number of tokens $L_O$) are computed, producing $S_{\mathcal{G}}$ and $S_O$. 
The information contained in the two representations are then aggregated by:
\begin{equation}
\begin{aligned}
h_{O\mathcal{G}} &= [h_O'; P; h_O'\odot P; h_O' \odot Q], \\
P &= S_{\mathcal{G}} h_{\mathcal{G}}'^{\top}, \\
Q &= S_{\mathcal{G}} S_{O}^{\top} h_O'^{\top}, \\
\end{aligned}
\end{equation}
where $h_{O\mathcal{G}} \in \mathbb{R}^{L_O \times 4H}$ is the aggregated observation representation, each token in text is represented by the weighted sum of graph representations.
Similarly, the aggregated graph representation $h_{\mathcal{G}O} \in \mathbb{R}^{N_{\mathcal{G}} \times 4H}$ can also be obtained, where each node in the graph is represented by the weighted sum of text representations. Finally, a linear transformation projects the two aggregated representations to a space with size $H$ of 64:
\begin{equation}
\begin{aligned}
h_{\mathcal{G}O} &= L(h_{\mathcal{G}O}), \\
h_{O\mathcal{G}} &= L(h_{O\mathcal{G}}). \\
\end{aligned}
\end{equation}

\subsection{Scorer}
\label{appd:scorer}

The scorer consists of a self-attention layer, a masked mean pooling layer, and a two-layer MLP.
As shown in Figure~\ref{fig:agent_diagram} and described above, the input to the scorer is the action candidate representation $h_{C_t}$, and one of the following game state representation:
\begin{center}
$
  s_t = 
      \begin{cases}
          h_{\mathcal{G}_t}                             & \text{if only graph input is available,}\\
          h_{O_t}                                       & \text{if only text observation is available, this degrades \ours to a \transdqn,}\\
          h_{\mathcal{G}O_t}$, $h_{O\mathcal{G}_t}      & \text{if both are available.}
      \end{cases}
$\\
\end{center}

First, a self-attention is applied to the game state representation $s_t$, producing $\hat{s_t}$.
If $s_t$ includes graph representations, this self-attention mechanism will reinforce the connection between each node and its related nodes.
Similarly, if $s_t$ includes text representation, the self-attention mechanism strengthens the connection between each token and other related tokens.
Further, masked mean pooling is applied to the self-attended state representation $\hat{s_t}$ and the action candidate representation $h_{C_t}$, this results in a state representation vector and a list of action candidate representation vectors.
We then concatenate the resulting vectors and feed them into a 2-layer MLP with a ReLU non-linear activation function in between.
The second MLP layer has an output dimension of 1, after squeezing the last dimension, the resulting vector is of size $N_{C_t}$, which is the number of action candidates provided at game step $t$.
We use this vector as the score of each action candidate.

\subsection{The $\mathrm{f_\Delta}$ Function}
\label{appd:f_delta}

As mentioned in Eqn.~\ref{eqn:continuous_graph_updater}, $\mathrm{f_\Delta}$ is an aggregator that combines information in $\mathcal{G}_{t-1}$, $A_{t-1}$, and $O_t$ to generate the graph difference $\Delta g_t$.

In specific, $\mathrm{f_\Delta}$ uses the same architecture as the representation aggregator described in Appendix~\ref{appd:rep_aggregator}.
Denoting the aggregator as a function $\mathrm{Aggr}$:
\begin{equation}
h_{PQ}, h_{QP} = \mathrm{Aggr}(h_{P}, h_{Q}),
\end{equation}
$\mathrm{f_\Delta}$ takes text observation representations $h_{O_t} \in \mathbb{R}^{L_{O_t} \times H}$, belief graph representations $h_{\mathcal{G}_{t-1}} \in \mathbb{R}^{N_{\mathcal{G}} \times H}$, and action representations $h_{A_{t-1}} \in \mathbb{R}^{L_{A_{t-1}} \times H}$ as input. 
$L_{O_t}$ and $L_{A_{t-1}}$ are the number of tokens in $O_t$ and $A_{t-1}$, respectively; $N_{\mathcal{G}}$ is the number of nodes in the graph; $H$ is hidden size of the input representations.

We first aggregate $h_{O_t}$ with $h_{\mathcal{G}_{t-1}}$, then similarly $h_{A_{t-1}}$ with $h_{\mathcal{G}_{t-1}}$:
\begin{equation}
\begin{aligned}
h_{O\mathcal{G}}, h_{\mathcal{G}O} &= \mathrm{Aggr}(h_{O_t}, h_{\mathcal{G}_{t-1}}), \\
h_{A\mathcal{G}}, h_{\mathcal{G}A} &= \mathrm{Aggr}(h_{A_{t-1}}, h_{\mathcal{G}_{t-1}}). \\
\end{aligned}
\end{equation}

The output of $\mathrm{f_\Delta}$ is:
\begin{equation}
\Delta g_t = [\bar{h_{O\mathcal{G}}}; \bar{ h_{\mathcal{G}O}}; \bar{h_{A\mathcal{G}}}; \bar{h_{\mathcal{G}A}}], 
\end{equation}
where $\bar{X}$ is the masked mean of $X$ on the first dimension.
The resulting concatenated vector $\Delta g_t$ has the size of $\mathbb{R}^{4H}$.

\subsection{The $\mathrm{f_d}$ Function}
\label{appd:f_d}

$\mathrm{f_d}$ is a decoder that maps a hidden graph representation $h_t \in \mathbb{R}^{H}$ (generated by the RNN) into a continuous adjacency tensor 
$\mathcal{G} \in [-1, 1]^{2\mathcal{R} \times \mathcal{N} \times \mathcal{N}}$.

Specifically, $\mathrm{f_d}$ consists of a 2-layer MLP:
\begin{equation}
\begin{aligned}
h_1 &= L^{\text{ReLU}}_1(h_t),\\
h_2 &= L^{\text{tanh}}_2(h_1).\\
\end{aligned}
\end{equation}
In which, $h_1 \in \mathbb{R}^{H}$, $h_2 \in [-1, 1]^{\mathcal{R} \times \mathcal{N} \times \mathcal{N}}$.
To better facilitate the message passing process of R-GCNs used in \ours's graph encoder, we explicitly use the transposed $h_2$ to represent the inversed relations in the belief graph. Thus, we have $\mathcal{G}$ defined as:
\begin{equation}
\mathcal{\mathcal{G}} = [h_2; h^{T}_2].
\end{equation}
The transpose is performed on the last two dimensions (both of size $\mathcal{N}$),
the concatenation is performed on the dimension of relations.

The tanh activation function on top of the second layer of the MLP restricts the range of our belief graph $\mathcal{G}$ within $[-1, 1]$. 
Empirically we find it helpful to keep the input of the multi-layer graph neural networks (the R-GCN graph encoder) in this range.

\section{Details of Pre-training Graph Updater for \ours}
\label{appendix:pretrain_continuous}

As briefly described in Section~\ref{subsection:graph_update}, we design two self-supervised tasks to pre-train the graph updater module of \ours.
As training data, we gather a collection of transitions from the \ftwp dataset.
Here, we denote a transition as a 3-tuple $(O_{t-1}, A_{t-1}, O_t)$.
Specifically, given text observation $O_{t-1}$, an action $A_{t-1}$ is issued; this leads to a new game state and $O_t$ is returned from the game engine.
Since the graph updater is recurrent (we use an RNN as its graph operation function), the set of transitions are stored in the order they are collected.

\subsection{Observation Generation (OG)}
\label{appendix:detail_og}
\begin{figure}[h!]
    \centering
    \includegraphics[width=0.8\textwidth]{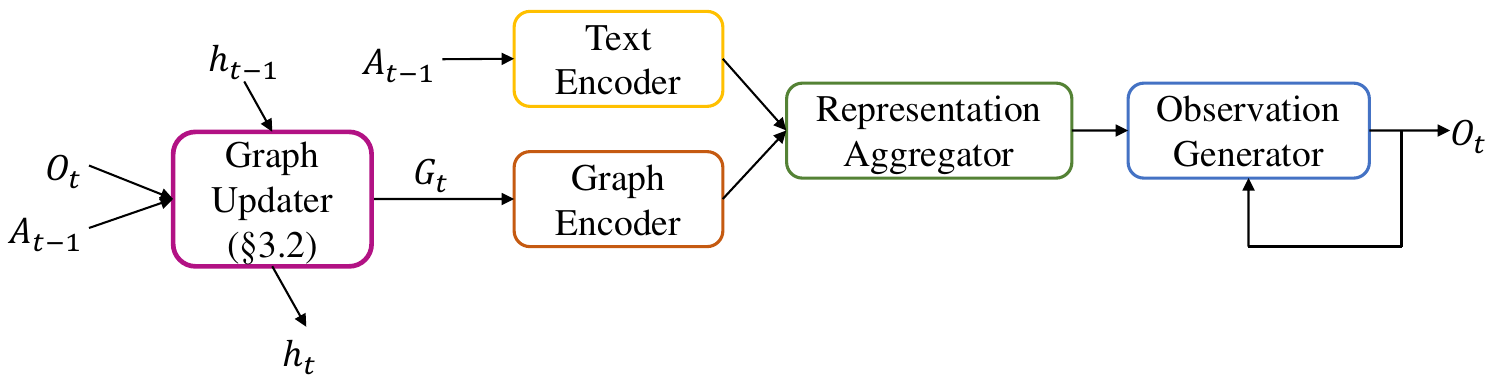}
    \caption{Observation generation model.}
    \label{fig:og_model}
\end{figure}

As shown in Figure~\ref{fig:og_model}, given a transition $(O_{t-1}, A_{t-1}, O_t)$, we use the belief graph $\mathcal{G}_t$ and $A_{t-1}$ to reconstruct $O_t$.
$\mathcal{G}_t$ is generated by the graph updater, conditioned on the recurrent information $h_{t-1}$ carried over from previous data point in the transition sequence.

\subsubsection{Observation Generator Layer}
\label{appd:observation_generator}

The observation generator is a transformer-based decoder. 
It consists of a word embedding layer, a transformer block, and a projection layer.

Similar to the text encoder, the embedding layer is frozen after initializing with the pre-trained fastText \citep{mikolov18fasttext} word embeddings.
Inside the transformer block, there is one self attention layer, two attention layers and a 3-layer MLP with ReLU non-linear activation functions in between.
Taking word embedding vectors and the two aggregated representations produced by the representation aggregator as input, the self-attention layer first generates a contextual encoding vectors for the words.
These vectors are then fed into the two attention layers to compute attention with graph representations and text observation representations respectively.
The two resulting vectors are thus concatenated, and they are fed into the 3-layer MLP.
The block hidden size of this transformer is $H=64$.

Finally, the output of the transformer block is fed into the projection layer, which is a linear transformation with output size same as the vocabulary size.
The resulting logits are then normalized by a softmax to generate a probability distribution over all words in vocabulary.

Following common practice, we also use a mask to prevent the decoder transformer to access ``future'' information during training.

\subsection{Contrastive Observation Classification (COC)}
\label{appendix:detail_coc}

The contrastive observation classification task shares the same goal of ensuring the generated belief graph $\mathcal{G}_t$ encodes the necessary information describing the environment state at step $t$.
However, instead of generating $O_t$ from $\mathcal{G}_t$, it requires a model to differentiate the real $O_t$ from some $\tilde{O_t}$ that are randomly sampled from other data points.
In this task, the belief graph does not need to encode the syntactical information as in the observation generation task, rather, a model can use its full capacity to learn the semantic information of the current environmental state.

\begin{figure}[h!]
    \centering
    \includegraphics[width=0.8\textwidth]{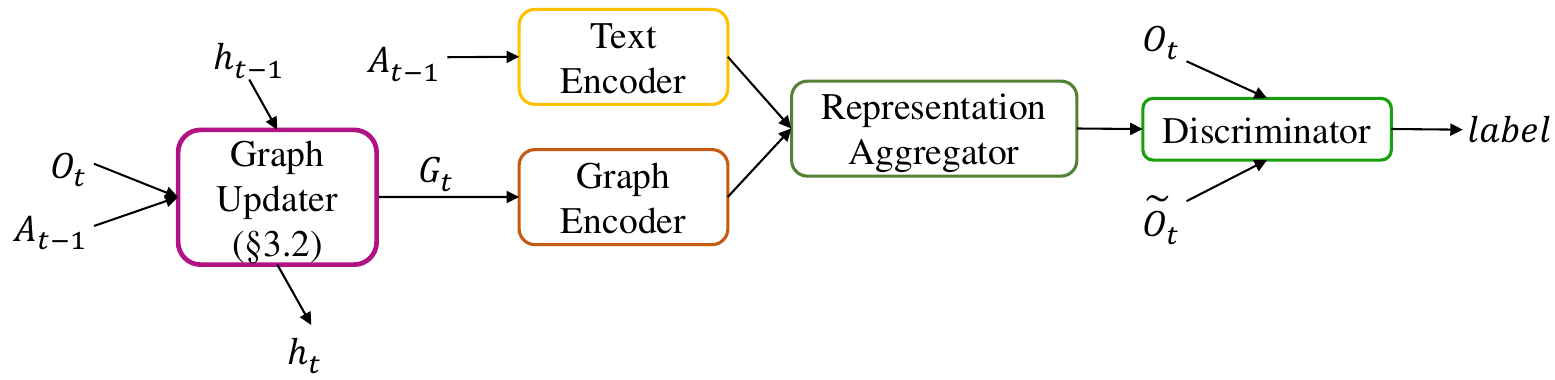}
    \caption{Contrastive observation classification model.}
    \label{fig:coc_model}
\end{figure}

We illustrate our contrastive observation classification model in Figure~\ref{fig:coc_model}.
This model shares most components with the previously introduced observation generation model, except replacing the observation generator module by a discriminator. 

\subsection{Reusing Graph Encoder in Action Scorer}
\label{appendix_reuse_continuous}

Both of the graph updater and action selector modules rely heavily on the graph encoder layer. It is natural to reuse the graph updater's graph encoder during the RL training of action selector.
Specifically, we use the pre-trained graph encoder (and all its dependencies such as node embeddings and relation embeddings) from either the above model to initialize the graph encoder in action selector. 
In such settings, we fine-tune the graph encoders during RL training.
In Appendix~\ref{appendix:more_results}, we compare \ours's performance between reusing the graph encoders with randomly initialize them.

\section{\oursgtp and Discrete Belief Graph}
\label{appendix:gatagtp}

\begin{figure}[t!]
    \centering
    \includegraphics[width=0.95\textwidth]{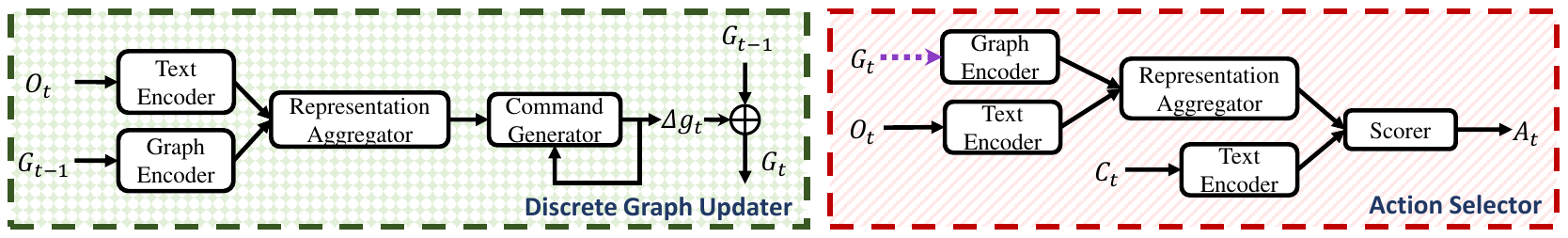}
    \caption{\oursgtp in detail. The coloring scheme is same as in Figure~\ref{fig:kg}. The \textcolor{green1}{discrete graph updater} first generates $\Delta g_t$ using $\mathcal{G}_{t-1}$ and $O_t$. Afterwards the \textcolor{red1}{action selector} uses $O_t$ and the updated graph $\mathcal{G}_t$ to select $A_t$ from the list of action candidates $C_t$. Purple dotted line indicates a detached connection (i.e., no back-propagation through such connection).}
    \label{fig:gatagt_diagram}
\end{figure}

As mentioned in Section~\ref{subsection:gata_with_gt}, since the TextWorld API provides ground-truth (discrete) KGs that describe game states at each step, we provide an agent that utilizes this information, as a strong baseline to \ours.
To accommodate the discrete nature of KGs provided by TextWorld, we propose \oursgtp, which has the same action scorer with \ours, but equipped with a discrete graph updater.
We show the overview structure of \oursgtp in Figure~\ref{fig:gatagt_diagram}.

\subsection{Discrete Graph Updater}
\label{appendix:discrete_graph_updater}

In the discrete graph setting, we follow \citep{zelinka2019building}, updating $\mathcal{G}_t$ with a set of discrete update operations that act on $\mathcal{G}_{t-1}$. 
In particular, we model the (discrete) $\Delta g_t$ as a set of update operations, wherein each update operation is a sequence of tokens. 
We define the following two elementary operations so that any graph update can be achieved in $k \geq 0$ such operations:
\begin{itemize}[leftmargin=*]
  \item \cmd{add(node1, node2, relation)}: add a directed edge, named \cmd{relation}, between \cmd{node1} and \cmd{node2}.
  \item \cmd{delete(node1, node2, relation)}: delete a directed edge, named \cmd{relation}, between \cmd{node1} and \cmd{node2}. If the edge does not exist, ignore this command.
\end{itemize}
Given a new observation string $O_t$ and $\mathcal{G}_{t-1}$, the agent generates $k\geq 0$ such operations to merge the newly observed information into its belief graph.
\begin{table}[h!]
    \small
    \centering
    \caption{Update operations matching the transition in Figure~\ref{fig:kg}.}
    \label{tab:example_cmds}
    \vspace{0.8em}
    \begin{tabular}{l}
        \toprule
         \cmd{<s> \textcolor{red}{add} player shed at <|> \textcolor{red}{add} shed backyard west\_of <|> \textcolor{red}{add} wooden door shed } \\
         \cmd{east\_of <|> \textcolor{red}{add} toolbox shed in <|> \textcolor{red}{add} toolbox closed is <|> \textcolor{red}{add} workbench }\\
         \cmd{shed in <|> \textcolor{red}{delete} player backyard at </s>}\\
        \bottomrule
    \end{tabular}
\end{table}

We formulate the update generation task as a sequence-to-sequence (Seq2Seq) problem and use a transformer-based model \citep{vaswani17transformer} to generate token sequences for the operations. 
We adopt the decoding strategy from \cite{meng19order}, where given an observation sequence $O_t$ and a belief graph $\mathcal{G}_{t-1}$, the agent generates a sequence of tokens that contains multiple graph update operations as subsequences, separated by a delimiter token \cmd{<|>}.

Since Seq2Seq set generation models are known to learn better with a consistent output ordering~\citep{meng19order}, we sort the ground-truth operations (e.g., always \cmd{add} before \cmd{delete}) for training. 
For the transition shown in Figure~\ref{fig:kg}, the generated sequence is shown in Table~\ref{tab:example_cmds}.

\subsection{Pre-training Discrete Graph Updater}
\label{appendix:train_discrete_graph_updater}

As described above, we frame the discrete graph updating behavior as a language generation task.
We denote this task as command generation (CG). 
Similar to the continuous version of graph updater in \ours, we pre-train the discrete graph updater using transitions collected from the \ftwp dataset.
It is worth mentioning that despite requiring ground-truth KGs in \ftwp dataset, \oursgtp does not require any ground-truth graph in the RL game to train and evaluate the action scorer.

For training discrete graph updater, we use the $\mathcal{G}^\text{seen}$ type of graphs provided by the TextWorld API. 
Specifically, at game step $t$, $\mathcal{G}^\text{seen}_t$ is a discrete partial KG that contains information the agent has observed from the beginning until step $t$. 
It is only possible to train an agent to generate belief about the world it has seen and experienced.

\begin{figure}[h!]
    \centering
    \includegraphics[width=0.7\textwidth]{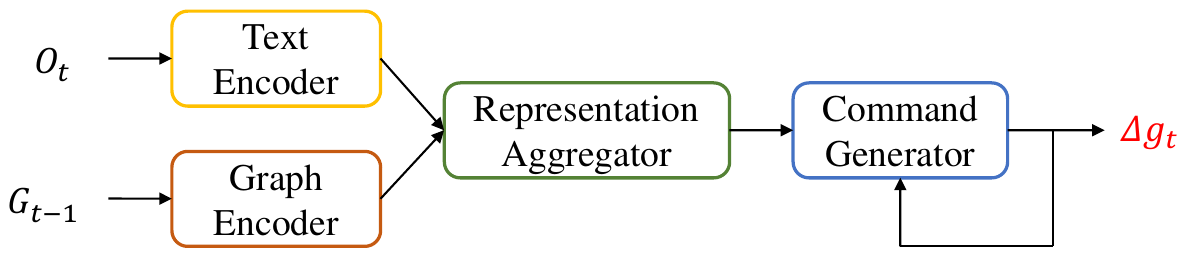}
    \caption{Command Generation Model.}
    \label{fig:cg_model}
\end{figure}

In the collection \ftwp transitions, every data point contains two consecutive graphs, we convert the difference between the graphs to ground-truth update operations (i.e., \cmd{add} and \cmd{delete} commands). 
We use standard teacher forcing technique to train the transformer-based Seq2Seq model.
Specifically, conditioned on the output of representation aggregator, the command generator is required to predict the $k^{\text{th}}$ token of the target sequence given all the ground-truth tokens up to time step $k-1$.
The command generator module is transformer-based decoder, similar to the observation generator described in Appendix~\ref{appd:observation_generator}.
Negative log-likelihood is used as loss function for optimization.
An illustration of the command generation model is shown in Figure ~\ref{fig:cg_model}.

During the RL training of action selector, the graph updater is detached without any back-propagation performed.
It generates token-by-token started by a begin-of-sequence token, until it generates an end-of-sequence token, or hitting the maximum sequence length limit.
The resulting tokens are consequently used to update the discrete belief graph.

\subsection{Pre-training a Discrete Graph Encoder for Action Scorer}
\label{appendix:pretrain_discrete_graph_encoder}

In the discrete graph setting, we take advantage of the accessibility of the ground-truth graphs.
Therefore we also consider various pre-training approaches to improve the performance of the graph encoder in the action selection module.
Similar to the training of graph updater, we use transitions collected from the \ftwp dataset as training data.

In particular, here we define a transition as a 6-tuple $(\mathcal{G}_{t-1}, O_{t-1}, C_{t-1}, A_{t-1}, \mathcal{G}_{t}, O_t$. 
Specifically, given $\mathcal{G}_{t-1}$ and $O_{t-1}$, an action $A_{t-1}$ is selected from the candidate list $C_{t-1}$; this leads to a new game state $\mathcal{S}_{t}$, thus $\mathcal{G}_{t}$ and $O_{t}$ are returned.
Note that $\mathcal{G}_{t}$ in transitions can either be $\mathcal{G}^\text{full}_t$ that describes the full environment state or $\mathcal{G}^\text{seen}_t$ that describes the part of state that the agent has experienced.

In this section, we start with providing details of the pre-training tasks and their corresponding models, and then show these models' performance for each of the tasks.

\subsubsection{Action Prediction (AP)}
\label{appendix:pretrain_ap}
Given a transition $(\mathcal{G}_{t-1}, O_{t-1}, C_{t-1}, A_{t-1}, \mathcal{G}_{t}, O_t, r_{t-1})$, we use $A_{t-1}$ as positive example and use all other action candidates in $C_{t-1}$ as negative examples. 
A model is required to identify $A_{t-1}$ amongst all action candidates given two consecutive graphs $\mathcal{G}_{t-1}$ and $\mathcal{G}_t$.

\begin{figure}[h!]
    \centering
    \includegraphics[width=0.6\textwidth]{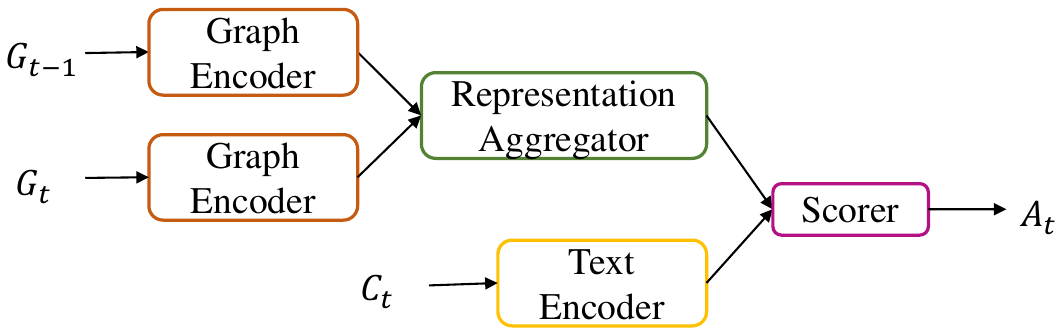}
    \caption{Action Prediction Model.}
    \label{fig:ap_model}
\end{figure}

We use a model with similar structure and components as the action selector of \ours.
As illustrated in Figure~\ref{fig:ap_model}, the graph encoder first converts the two input graphs $\mathcal{G}_{t-1}$ and $\mathcal{G}_t$ into hidden representations, the representation aggregator combines them using attention mechanism.
The list of action candidates (which includes $A_{t-1}$ and all negative examples) are fed into the text encoder to generate action candidate representations.
The scorer thus takes these representations and the aggregated graph representations as input, and it outputs a ranking over all action candidates.

In order to achieve good performance in this setting, the bi-directional attention between $\mathcal{G}_{t-1}$ and $\mathcal{G}_t$ in the representation aggregator needs to effectively determine the difference between the two sparse graphs. 
To achieve that, the graph encoder has to extract useful information since often the difference between $\mathcal{G}_{t-1}$ and $\mathcal{G}_t$ is minute (e.g., before and after taking an apple from the table, the only change is the location of the apple).

\subsubsection{State Prediction (SP)}
\label{appendix:pretrain_sp}

\begin{figure}[h!]
    \centering
    \includegraphics[width=0.6\textwidth]{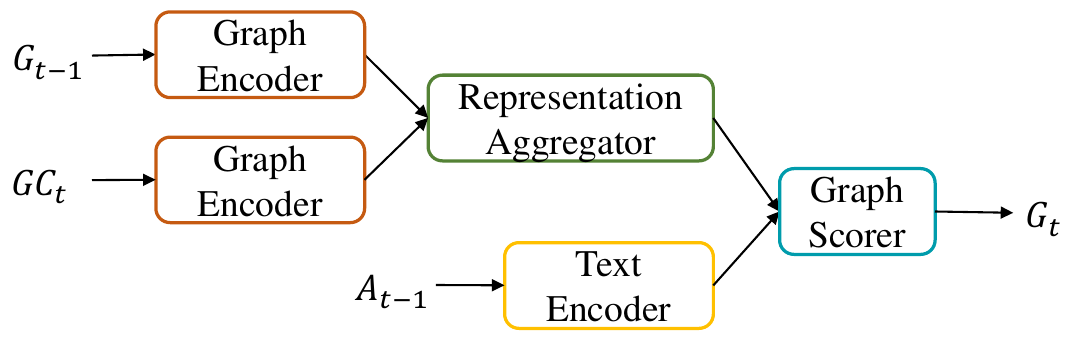}
    \caption{State Prediction Model.}
    \label{fig:sp_model}
\end{figure}

Given a transition $(\mathcal{G}_{t-1}, O_{t-1}, C_{t-1}, A_{t-1}, \mathcal{G}_{t}, O_t, r_{t-1})$, we use $\mathcal{G}_{t}$ as positive example and gather a set of game states by issuing all other actions in $C_{t-1}$ except $A_{t-1}$. We use the set of graphs representing the resulting game states as negative samples.
In this task, a model is required to identify $\mathcal{G}_{t}$ amongst all graph candidates $GC_t$ given the previous graph $\mathcal{G}_{t-1}$ and the action taken $A_{t-1}$.

As shown in Figure~\ref{fig:sp_model}, a similar model is used to train both the SP and AP tasks.
% The only difference is that the representation aggregator needs to compute multiple times to generate a combined graph representation for $\mathcal{G}_{t-1}$ with each graph in $GC_t$. 

\subsubsection{Deep Graph Infomax (DGI)}
\label{appendix:pretrain_dgi}

\begin{figure}[h!]
    \centering
    \includegraphics[width=0.6\textwidth]{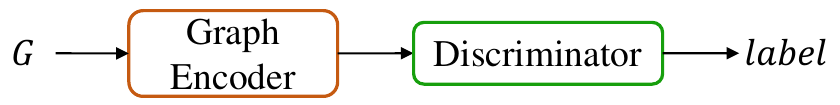}
    \caption{Deep Graph Infomax Model.}
    \label{fig:dgi_model}
\end{figure}

This is inspired by Velickovic et al., \citep{velickovic2018deep}.
Given a transition $(\mathcal{G}_{t-1}, O_{t-1}, C_{t-1}, A_{t-1}, \mathcal{G}_{t}, O_t, r_{t-1})$, we map the graph $\mathcal{G}_{t}$ into its node embedding space. The node embedding vectors of $\mathcal{G}_{t}$ is denoted as $H$.
We randomly shuffle some of the node embedding vectors to construct a ``corrupted'' version of the node representations, denoted as $\tilde{H}$.

Given node representations $H = \{\overrightarrow{h_1}, \overrightarrow{h_2}, ..., \overrightarrow{h_N}\}$ and corrupted representations of these nodes $\tilde{H} = \{\tilde{\overrightarrow{h_1}}, \tilde{\overrightarrow{h_2}}, ..., \tilde{\overrightarrow{h_N}}\}$, where $N$ is the number of vertices in the graph, 
a model is required to discriminate between the original and corrupted representations of nodes.
As shown in Figure~\ref{fig:dgi_model}, the model is composed of a graph encoder and a discriminator.
Specifically, following \citep{velickovic2018deep}, we utilize a noise-contrastive objective with a binary cross-entropy (BCE) loss between the samples from the joint (positive examples) and the product of marginals (negative examples). 
To enable the discriminator to discriminate between $\mathcal{G}_{t}$ and the negative samples, the graph encoder must learn useful graph representations at both global and local level.

\subsubsection{Performance on Graph Encoder Pre-training Tasks}
\label{appendix:pretrain_scores}

We provide test performance of all the models described above for graph representation learning.
We fine-tune the models on validation set and report their performance on test set.

Additionally, as mentioned in Section~\ref{subsection:action_selection} and Appendix~\ref{appendix:model}, we adapt the original R-GCN to condition the graph representation on additional information contained by the relation labels.
We show an ablation study for this in Table~\ref{tab:pretrain_scores}, where R-GCN denotes the original R-GCN~\citep{schlichtkrull2018rgcn} and R-GCN w/ R-Emb denotes our version that considers relation labels.

Note, as mentioned in previous sections, the dataset to train, valid and test these four pre-training tasks are extracted from the \ftwp dataset.
There exist unseen nodes (ingredients in recipe) in the validation and test sets of \ftwp, it requires strong generalizability to get decent performance on these datasets.

From Table~\ref{tab:pretrain_scores}, we show the relation label representation significantly boosts the generalization performance on these datasets.
Compared to AP and SP, where relation label information has significant effect, both models perform near perfectly on the DGI task.
This suggests the corruption function we consider in this work is somewhat simple, we leave this for future exploration.

\begin{table}[h!]
    \small
    \centering
    \caption{Test performance of models on all pre-training tasks.}
    \label{tab:pretrain_scores}
    \vspace{0.8em}
    \begin{tabular}{c|c|c|c}
        \toprule
        Task & Graph Type  & R-GCN & R-GCN w/ R-Emb \\
        \midrule
        \multicolumn{4}{c}{Accuracy}\\
        \midrule
        \multirow{2}{*}{AP}   & full    & 0.472 & \textbf{0.891} \\
        & seen    & 0.631 & \textbf{0.873} \\
        \midrule
        \multirow{2}{*}{SP}    & full    & 0.419 & \textbf{0.926} \\
        & seen    & 0.612 & \textbf{0.971} \\
        \midrule
        \multirow{2}{*}{DGI}   & full    & 0.999 & \textbf{1.000} \\
        & seen    & \textbf{1.000} & \textbf{1.000} \\
        % \midrule
        % \multicolumn{4}{c}{$\text{F}_1$ Score}\\
        % \midrule
        % CG   & seen    & 0.910 & \textbf{0.968} \\
        \bottomrule
    \end{tabular}
\end{table}

\section{Additional Results and Discussions}
\label{appendix:more_results}

\subsection{Training Curves}

\begin{sidewaysfigure*}
    \centering
    \includegraphics[width=1.0\textwidth]{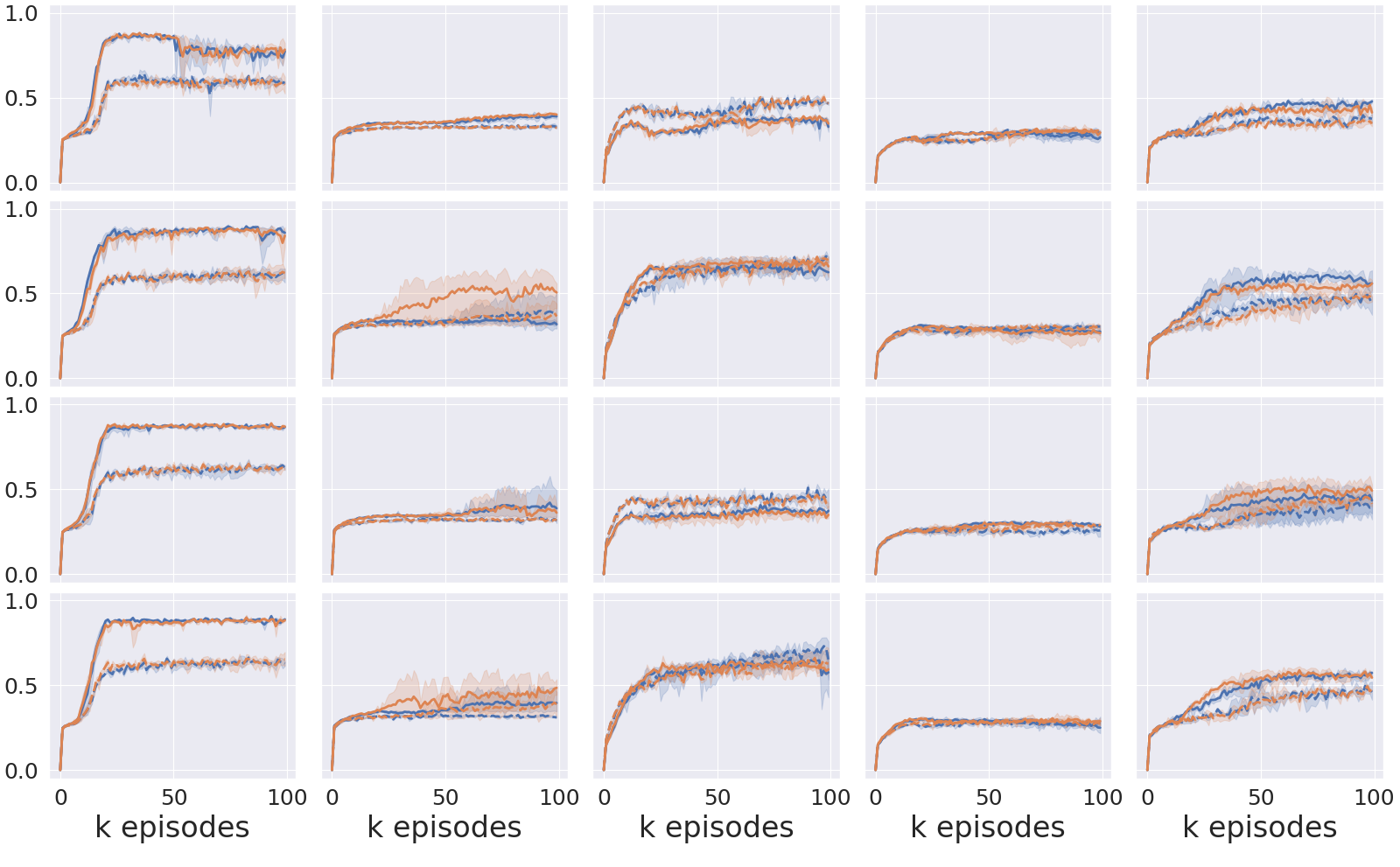}
    \caption{\ours's training curves (averaged over 3 seeds, band represents standard deviation). 
    Columns are difficulty levels 1/2/3/4/5.
    The upper two rows are \ours using belief graphs generated by the graph updater pre-trained with observation generation task;
    The lower two rows are \ours using belief graphs generated by the graph updater pre-trained with contrastive observation classification task.
    In the 4 rows, the presence of text observation are False/True/False/True.
    In the figure, \textcolor{blue1}{blue} lines indicate the graph encoder in action selector is \textcolor{red}{randomly} initialized; \textcolor{orange}{orange} lines indicate the graph encoder in action selector is initialized by the pre-trained \textcolor{blue1}{observation generation} and \textcolor{orange}{contrastive observation classification} tasks.
    Solid lines indicate 20 training games, dashed lines indicate 100 training games.}
    \label{fig:train_full_gata}
\end{sidewaysfigure*}

\begin{sidewaysfigure*}
    \centering
    \includegraphics[width=1.0\textwidth]{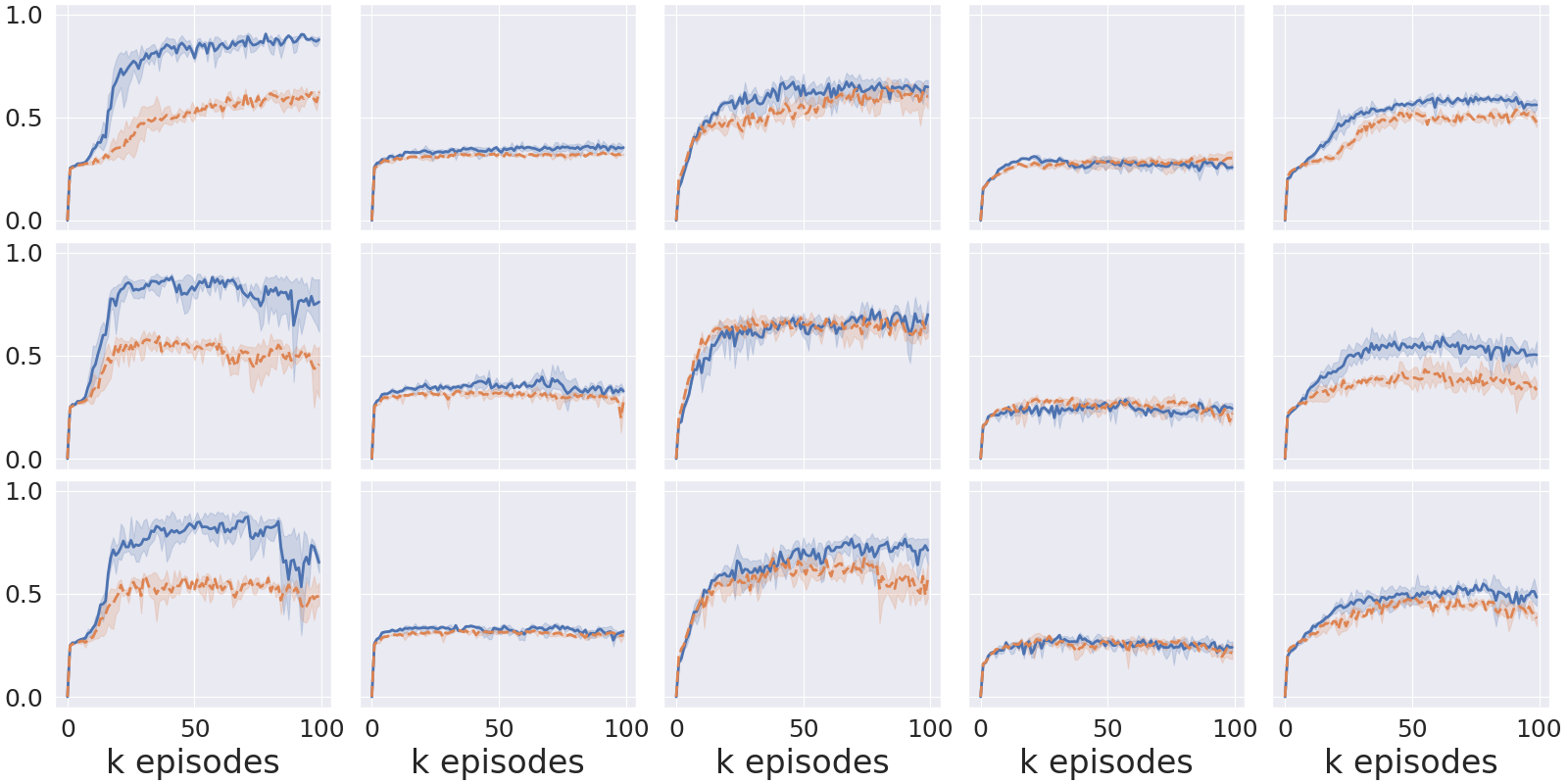}
    \caption{The text-based baseline agents' training curves (averaged over 3 seeds, band represents standard deviation). 
    Columns are difficulty levels 1/2/3/4/5, rows are \transdqn, \transdrqn and \transdrqnp, respectively.
    All of the three agents take text observation $O_t$ as input.
    In the figure, \textcolor{blue1}{blue} solid lines indicate the training set with 20 games; \textcolor{orange}{orange} dashed lines indicate the training set with 100 games.}
    \label{fig:train_full_baseline}
\end{sidewaysfigure*}

\begin{sidewaysfigure*}
    \centering
    \includegraphics[width=1.0\textwidth]{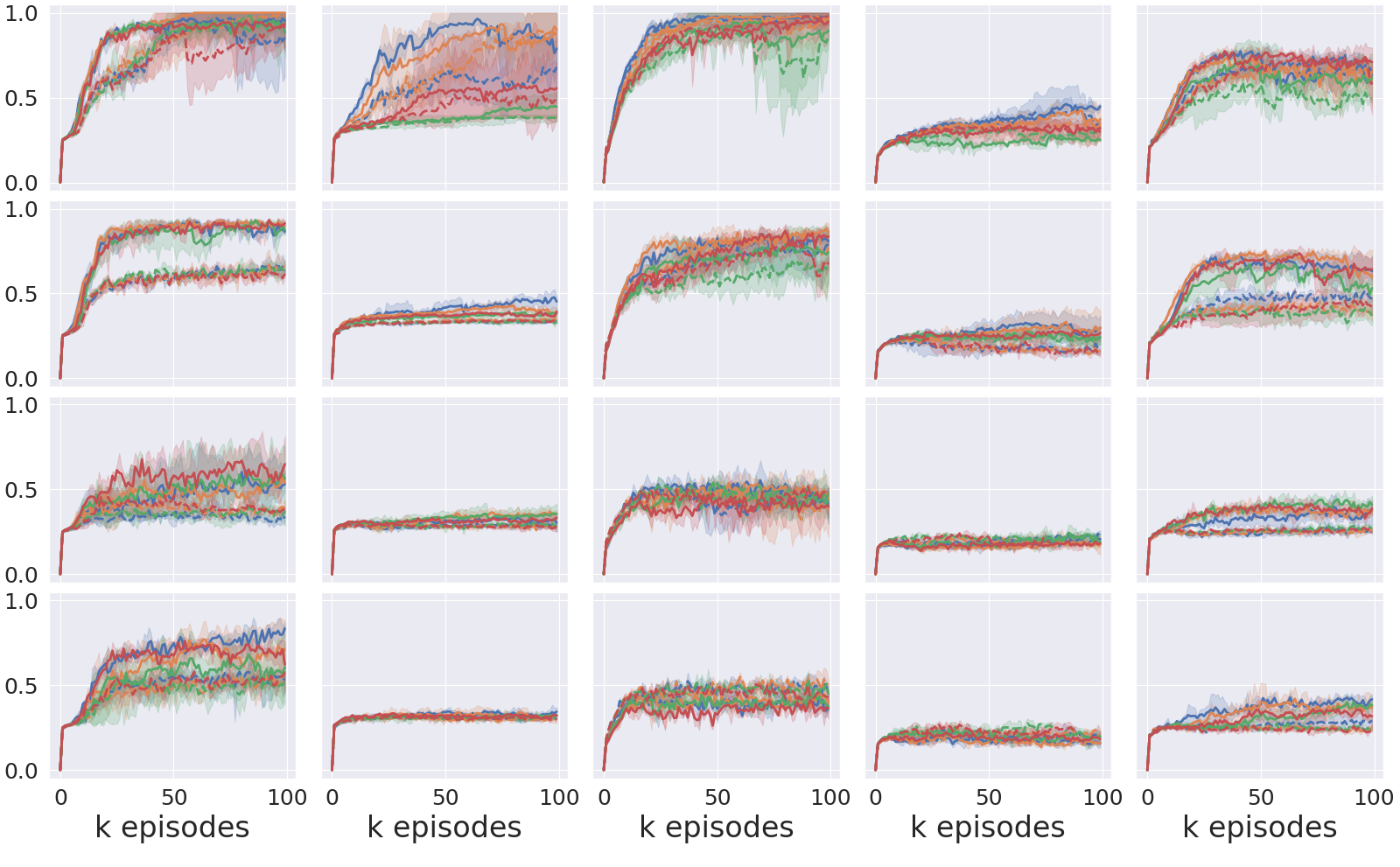}
    \caption{\oursgtp and \oursgtf's training curves (averaged over 3 seeds, band represents standard deviation). 
    Columns are difficulty levels 1/2/3/4/5.
    The upper two rows are \oursgtf when text observation is absent and present as input;
    the lower two rows are \oursgtp when text observation is absent and present as input.
    In the figure, \textcolor{blue1}{blue}/\textcolor{orange}{orange}/\textcolor{green1}{green} indicate the agent's graph encoder is initialized with \textcolor{blue1}{AP}/\textcolor{orange}{SP}/\textcolor{green1}{DGI} pre-training tasks.
    \textcolor{red}{Red} lines indicate the graph encoder is \textcolor{red}{randomly} initialized.
    Solid lines indicate 20 training games, dashed lines indicate 100 training games.}
    \label{fig:train_full_gatagt}
\end{sidewaysfigure*}

We report the training curves of all our mentioned experiment settings.
Figure~\ref{fig:train_full_gata} shows the \ours's training curves.
Figure~\ref{fig:train_full_baseline} shows the training curves of the three text-based baseline (\transdqn, \transdrqn, \transdrqnp).
Figure~\ref{fig:train_full_gatagt} shows the training curve of \oursgtf (no graph updater, the action scorer takes ground-truth graphs as input) and \oursgtp (graph updater is trained using ground-truth graphs \textit{from the \ftwp dataset}, the trained graph updater maintains a discrete belief graph throughout the RL training).

\clearpage

\clearpage

\clearpage

\subsection{Training Scores}
In Table~\ref{tab:train_score_full} we provide all agents' max training scores, each score is averaged over 3 random seeds. All scores are normalized.
Note as described in Section~\ref{subsection:action_selection}, we use ground-truth KGs to train the action selector, $\mathcal{G}^{\text{belief}}$ is only used during evaluation.

\begin{table}[h!]
    \scriptsize
    \centering
    \caption{Agents' \textbf{Max} performance on \textbf{Training} games, averaged over 3 random seeds. In this table, \sspade, \sdiamond represent $O_t$ and $\mathcal{G}^\text{full}_t$, respectively.
    \sclub represents discrete belief graph generated by \oursgtp (trained with ground-truth graphs of \ftwp).
    \sstar and \sinfinite indicate continuous belief graph generated by \ours, pre-trained with observation generation (OG) task and contrastive observation classification (COC) task, respectively. Light blue shadings represent numbers that are greater than or equal to \transdqn; light yellow shading represent number that are greater than or equal to all of \transdqn, \transdrqn and \transdrqnp.}
    \label{tab:train_score_full}
    \vspace{0.8em}
    \begin{tabular}{c|c|cccccc|cccccc|c}
        \toprule
        \multicolumn{2}{c|}{} & \multicolumn{6}{c|}{20 Training Games} & \multicolumn{6}{c|}{100 Training Games}  & Avg.\\
        \midrule
        \multicolumn{2}{c|}{Difficulty Level} & 1 & 2 & 3 & 4 & 5 & $\%\uparrow$ & 1 & 2 & 3 & 4 & 5 & $\%\uparrow$ & $\%\uparrow$ \\
        \midrule   
        Input & Agent & \multicolumn{13}{c}{Text-based Baselines} \\
        \midrule   
        \sspade & \transdqn         & 90.8 & 36.9 & 69.3 & 31.5 & 61.2 & ----- & 63.4 & 33.2 & 66.1 & 31.9 & 55.2 & ----- & ----- \\
        \midrule   
        \sspade & \transdrqn        & 88.8 & 41.7 & 76.6 & 29.6 & 60.7 & +2.9 & 60.8 & 33.7 & 71.7 & 30.6 & 44.9 & -3.4 & -0.2 \\
        \midrule   
        \sspade & \transdrqnp      & 89.1 & 35.6 & 78.0 & 30.9 & 58.1 & +0.0 & 61.1 & 32.8 & 70.0 & 30.0 & 50.3 & -2.8 & -1.4  \\
        \midrule   
        \multicolumn{1}{c}{} &\multicolumn{1}{c|}{Pre-training} & \multicolumn{13}{c}{\ours} \\
        \midrule 
        \sstar  & N/A           & 87.9 & \cellcolor{lb}40.4 & 40.1 & 30.8 & 50.1 & -11.2  & \cellcolor{ly}65.1 & \cellcolor{ly}34.6 & 51.2 & \cellcolor{ly}32.0 & 41.8  & -7.9  & -9.6  \\
        \sstar  & OG            & 88.8 & \cellcolor{lb}40.8 & 40.1 & \cellcolor{ly}32.1 & 48.2 & -10.6  & \cellcolor{ly}63.8 & \cellcolor{ly}33.9 & 51.9 & \cellcolor{ly}32.6 & 39.3  & -9.1  & -9.8  \\
        \midrule 
        \sstar \sspade & N/A        & 90.2 & 35.4 & 69.0 & \cellcolor{ly}32.0 & \cellcolor{ly}62.8 & -0.2  & \cellcolor{ly}63.9 & \cellcolor{ly}41.2 & \cellcolor{ly}72.2 & \cellcolor{ly}32.2 & 50.8  & \cellcolor{ly}+5.4 & \cellcolor{ly}+2.6  \\
        \sstar \sspade & OG         & 90.0 & \cellcolor{ly}57.1 & \cellcolor{lb}70.6 & \cellcolor{ly}31.7 & 57.9 & \cellcolor{ly}+10.2  & \cellcolor{ly}64.2 & \cellcolor{ly}38.9 & \cellcolor{ly}72.5 & \cellcolor{ly}32.4 & 50.1  & \cellcolor{ly}+4.1 & \cellcolor{ly}+7.1  \\
        \midrule 
        \sinfinite  & N/A           & 89.0 & \cellcolor{ly}43.1 & 41.2 & \cellcolor{ly}31.8 & 48.7 & -9.0  & \cellcolor{ly}65.8 & \cellcolor{lb}33.4 & 51.5 & 29.1 & 44.0 & -9.4 & -9.2  \\
        \sinfinite  & COC            & 89.6 & \cellcolor{lb}41.0 & 39.2 & \cellcolor{ly}31.9 & 54.4 & -8.7  & \cellcolor{ly}65.8 & \cellcolor{lb}33.4 & 48.6 & 31.2 & 47.2 & -7.8 & -8.2  \\
        \midrule 
        \sinfinite \sspade & N/A        & \cellcolor{ly}90.9 & \cellcolor{lb}41.4 & 66.1 & 31.3 & 58.8 & \cellcolor{lb}+0.6  & \cellcolor{ly}67.1 & 33.1 & \cellcolor{ly}73.6 & 29.9 & 51.2 & \cellcolor{ly}+0.7 & \cellcolor{ly}+0.6  \\
        \sinfinite \sspade & COC         & 90.2 & \cellcolor{ly}50.8 & 66.4 & \cellcolor{ly}31.9 & 59.3 & \cellcolor{ly}+6.2  & \cellcolor{ly}67.4 & \cellcolor{ly}41.5 & \cellcolor{lb}66.6 & 31.4 & 50.8 & \cellcolor{ly}+4.5 & \cellcolor{ly}+5.4  \\
        
        \midrule 
        \multicolumn{1}{c}{} &\multicolumn{1}{c|}{} & \multicolumn{13}{c}{\oursgtp} \\
        \midrule 

        \sclub & N/A               & 73.3 & 34.5 & 50.5 & 21.7 & 43.5 & -22.6 & 49.3 & 31.1 & 54.3 & 25.1 & 28.8  & -23.1 & -22.8 \\
        \sclub & AP                & 68.4 & 34.8 & 61.3 & 23.8 & 43.1 & -19.2 & 40.9 & 31.3 & 55.4 & 24.8 & 28.8  & -25.7 & -22.4 \\
        \sclub & SP                & 62.7 & \cellcolor{lb}38.1 & 57.5 & 23.5 & 44.1 & -19.6 & 50.2 & 30.8 & 55.0 & 23.7 & 28.0  & -24.0 & -21.8 \\
        \sclub & DGI               & 64.9 & \cellcolor{lb}37.0 & 55.6 & 25.7 & 47.4 & -17.8 & 43.4 & 31.8 & 58.3 & 25.3 & 30.2  & -22.7 & -20.3 \\
        % \sdag & CG                & 77.1 & 35.8 & 61.1 & 19.6 & 42.5 & -19.6 & 42.1 & 31.0 & 53.2 & 23.8 & 27.4  & -27.3 & -23.5 \\
        \midrule 
        
        \sclub \sspade & N/A           & 77.5 & 33.9 & 45.6 & 26.3 & 40.2 & -21.6 & 59.5 & 32.3 & 55.4 & 29.1 & 27.6  & -16.8 & -19.2 \\
        \sclub \sspade & AP            & 87.5 & 35.8 & 50.4 & 22.3 & 45.4 & -17.8 & 61.3 & 32.2 & 56.3 & 25.3 & 33.0  & -16.4 & -17.1 \\
        \sclub \sspade & SP            & 80.0 & 35.5 & 50.2 & 23.5 & 44.0 & -19.4 & 57.3 & 32.1 & 58.3 & 27.1 & 29.2  & -17.4 & -18.4 \\
        \sclub \sspade & DGI           & 70.3 & 33.9 & 51.4 & 26.3 & 42.1 & -20.9 & 57.7 & 32.7 & 55.6 & 28.8 & 29.8  & -16.4 & -18.6 \\
        % \sdag \sspade & CG            & 72.5 & 34.9 & 52.8 & 24.9 & 45.2 & -19.3 & 61.3 & 32.4 & 56.0 & 25.3 & 27.6  & -18.5 & -18.9 \\
        
        \midrule 
        \multicolumn{1}{c}{} &\multicolumn{1}{c|}{} & \multicolumn{13}{c}{\oursgtf} \\
        \midrule 

        \sdiamond & N/A               & \cellcolor{ly}98.6 & \cellcolor{ly}58.4 & \cellcolor{ly}95.6 & \cellcolor{ly}36.1 & \cellcolor{ly}80.9 & \cellcolor{ly}+30.3 & \cellcolor{ly}96.0 & \cellcolor{ly}53.4 & \cellcolor{ly}97.9 & \cellcolor{ly}36.0 & \cellcolor{ly}76.4 & \cellcolor{ly}+42.3 & \cellcolor{ly}+36.3  \\
        \sdiamond & AP                & \cellcolor{ly}98.7 & \cellcolor{ly}97.5 & \cellcolor{ly}98.3 & \cellcolor{ly}48.1 & \cellcolor{ly}79.3 & \cellcolor{ly}+59.4 & \cellcolor{ly}97.1 & \cellcolor{ly}74.7 & \cellcolor{ly}98.3 & \cellcolor{ly}44.5 & \cellcolor{ly}75.9 & \cellcolor{ly}+60.8 & \cellcolor{ly}+60.1 \\
        \sdiamond & SP                & \cellcolor{ly}100.0 & \cellcolor{ly}96.9 & \cellcolor{ly}98.3 & \cellcolor{ly}44.9 & \cellcolor{ly}76.6 & \cellcolor{ly}+56.5 & \cellcolor{ly}98.6 & \cellcolor{ly}90.5 & \cellcolor{ly}99.0 & \cellcolor{ly}38.9 & \cellcolor{ly}73.4 & \cellcolor{ly}+66.6  & \cellcolor{ly}+61.5 \\
        \sdiamond & DGI               & \cellcolor{ly}96.9 & \cellcolor{ly}45.4 & \cellcolor{ly}95.3 & 28.7 & \cellcolor{ly}72.6 & \cellcolor{ly}+15.4 & \cellcolor{ly}98.2 & \cellcolor{ly}39.1 & \cellcolor{ly}90.1 & \cellcolor{ly}33.0 & \cellcolor{ly}62.4 &  \cellcolor{ly}+25.1 &  \cellcolor{ly}+20.2 \\
        \midrule 

        \sdiamond \sspade & N/A           & \cellcolor{ly}91.7 & \cellcolor{ly}55.9 & \cellcolor{ly}80.9 & \cellcolor{ly}33.6 & \cellcolor{ly}63.2 & \cellcolor{ly}+15.8 & \cellcolor{ly}73.5 & \cellcolor{ly}48.1 & \cellcolor{lb}67.7 & 31.8 & \cellcolor{ly}56.7 & \cellcolor{ly}+13.1 & \cellcolor{ly}+14.5  \\
        \sdiamond \sspade & AP            & 87.9 & \cellcolor{ly}62.4 & \cellcolor{ly}78.8 & \cellcolor{ly}32.4 & \cellcolor{ly}62.8 & \cellcolor{ly}+17.0 & \cellcolor{ly}76.8 & \cellcolor{ly}54.0 & \cellcolor{ly}73.7 & \cellcolor{ly}34.1 & \cellcolor{ly}55.6 & \cellcolor{ly}+20.6 & \cellcolor{ly}+18.8  \\
        \sdiamond \sspade & SP            & 90.7 & \cellcolor{ly}55.8 & \cellcolor{ly}83.8 & 30.7 & \cellcolor{ly}64.2 & \cellcolor{ly}+14.9 & 60.4 & \cellcolor{ly}40.1 & \cellcolor{lb}67.4 & 31.1 & 51.5 & \cellcolor{ly}+1.8 & \cellcolor{ly}+8.3  \\
        \sdiamond \sspade & DGI           & 88.1 & \cellcolor{lb}38.1 & \cellcolor{lb}73.0 & \cellcolor{ly}32.5 & \cellcolor{ly}62.5 & \cellcolor{lb}+2.2 & \cellcolor{ly}66.4 & \cellcolor{ly}35.5 & 59.1 & 30.4 & 49.6 & -2.8 & -0.3  \\
        \bottomrule
    \end{tabular}
\end{table}

\clearpage

\subsection{Test Results}

In Table~\ref{tab:test_score_full} we provide all our agent variants and the text-based baselines' test scores.
We report agents' test score corresponding to their best validation scores.

\begin{table}[h!]
    \scriptsize
    \centering
    \caption{Agents' performance on \textbf{test} games, model selected using best validation performance. \textbf{Boldface} and \underline{underline} represent the highest and second highest values in a setting (excluding \oursgtf which has access to the ground-truth graphs of the RL games). 
    In this tabel, \sspade, \sdiamond represent $O_t$ and $\mathcal{G}^\text{full}_t$, respectively.
    \sclub represents discrete belief graph generated by \oursgtp (pre-trained with ground-truth graphs of \ftwp).
    \sstar and \sinfinite indicate continuous belief graph generated by \ours, pre-trained with observation generation (OG) task and contrastive observation classification (COC) task, respectively. Light blue shadings represent numbers that are greater than or equal to \transdqn; light yellow shading represent number that are greater than or equal to all of \transdqn, \transdrqn and \transdrqnp.}
    \label{tab:test_score_full}
    \vspace{0.8em}
    \begin{tabular}{c|c|cccccc|cccccc|c}
        \toprule
        \multicolumn{2}{c|}{} & \multicolumn{6}{c|}{20 Training Games} & \multicolumn{6}{c|}{100 Training Games} &  Avg.\\
        \midrule
        \multicolumn{2}{c|}{Difficulty Level} & 1   & 2  & 3   &  4  &  5  & $\%\uparrow$ & 1   & 2  & 3   &  4  &  5  & $\%\uparrow$  & $\%\uparrow$\\
        \midrule   
        Input & Agent  & \multicolumn{13}{c}{Text-based Baselines}\\
        \midrule   
        \sspade & \transdqn         & 66.2 & 26.0 & 16.7 & 18.2 & \textbf{27.9} & -----  & 62.5 & 32.0 & 38.3 & 17.7 & 34.6 & ----- & -----\\
        \midrule   
        \sspade & \transdrqn        & 62.5 & 32.0 & 28.3 & 12.7 & 26.5 & +10.3 & 58.8 & 31.0 & 36.7 & 21.4 & 27.4 & -2.6 & +3.9 \\
        \midrule   
        \sspade & \transdrqnp      & 65.0 & 30.0 & 35.0 & 11.8 & 18.3 & +10.7 & 58.8 & 33.0 & 33.3 & 19.5 & 30.6 & -3.4 & +3.6  \\
        \midrule   
        \multicolumn{1}{c}{} &\multicolumn{1}{c|}{Pre-training} & \multicolumn{13}{c}{\ours}  \\
        \midrule   
        %%%%%%%%%%%%%%%%%%%%%%
        \sstar            & N/A           & \cellcolor{ly}70.0 & 20.0 & \cellcolor{lb}20.0 & \cellcolor{ly}18.6 & 26.3 & -0.2 & \cellcolor{ly}62.5 & \cellcolor{lb}32.0 & \cellcolor{ly}46.7 & \cellcolor{ly}\textbf{27.7} & \cellcolor{ly}35.4 & \cellcolor{ly}+16.1 & \cellcolor{ly}+8.0  \\
        \sstar            & OG            & \cellcolor{ly}66.2 & \cellcolor{lb}28.0 & \cellcolor{lb}21.7 & 15.9 & 24.3 & \cellcolor{lb}+2.4 & \cellcolor{ly}\underline{66.2} & \cellcolor{ly}34.0 & \cellcolor{ly}40.0 & \cellcolor{ly}21.4 & 34.0 & \cellcolor{ly}+7.2  & \cellcolor{ly}+4.8  \\
        \midrule   
        %%%%%%%%%%%%%%%%%%%%%%
        \sstar \sspade        & N/A           & \cellcolor{ly}66.2 & \cellcolor{ly}34.0 & \cellcolor{lb}30.0 & 12.7 & 24.3 & \cellcolor{ly}+13.5 & \cellcolor{ly}\underline{66.2} & \cellcolor{ly}\textbf{38.0} & 36.7 & \cellcolor{ly}\underline{27.3} & \cellcolor{ly}36.1 & \cellcolor{ly}+15.8 & \cellcolor{ly}+14.6  \\
        \sstar \sspade        & OG            & \cellcolor{ly}66.2 & \cellcolor{ly}\textbf{48.0} & \cellcolor{lb}26.7 & 15.5 & 26.3 & \cellcolor{ly}+24.8 & \cellcolor{ly}\underline{66.2} & \cellcolor{ly}\underline{36.0} & \cellcolor{ly}\textbf{58.3} & 14.1 & \cellcolor{ly}\textbf{45.0} & \cellcolor{ly}+16.1 & \cellcolor{ly}+20.4  \\
        \midrule   
        %%%%%%%%%%%%%%%%%%%%%%
        \sinfinite            & N/A           & \cellcolor{ly}\underline{73.8} & \cellcolor{ly}42.0 & \cellcolor{lb}26.7 & \cellcolor{ly}\textbf{20.9} & 24.5 & \cellcolor{ly}+27.1 & \cellcolor{ly}62.5 & 30.0 & \cellcolor{ly}\underline{51.7} & \cellcolor{ly}23.6 & \cellcolor{ly}36.0 & \cellcolor{ly}+13.2 & \cellcolor{ly}+20.2  \\
        \sinfinite            & COC            & \cellcolor{ly}66.2 & \cellcolor{lb}29.0 & \cellcolor{lb}30.0 & \cellcolor{ly}18.2 & \underline{27.7} & \cellcolor{ly}+18.1 & \cellcolor{ly}\underline{66.2} & \cellcolor{ly}34.0 & \cellcolor{ly}41.7 & \cellcolor{lb}19.1 & \cellcolor{ly}\underline{40.3} & \cellcolor{ly}+9.1 & \cellcolor{ly}+13.6  \\
        \midrule   
        %%%%%%%%%%%%%%%%%%%%%%
        \sinfinite \sspade        & N/A           & \cellcolor{ly}68.8 & \cellcolor{ly}33.0 & \cellcolor{ly}\underline{41.7} & 17.7 & 27.0 & \cellcolor{ly}+34.9 & \cellcolor{ly}62.5 & \cellcolor{ly}33.0 & \cellcolor{ly}46.7 & \cellcolor{ly}25.9 & 33.4 & \cellcolor{ly}+13.6 & \cellcolor{ly}+24.2  \\
        \sinfinite \sspade        & COC            & \cellcolor{ly}66.2 & \cellcolor{ly}\underline{44.0} & \cellcolor{lb}16.7 & \cellcolor{ly}\underline{20.0} & 21.7 & \cellcolor{ly}+11.4 & \cellcolor{ly}\textbf{70.0} & \cellcolor{ly}34.0 & \cellcolor{ly}45.0 & 12.3 & \cellcolor{ly}36.2 & \cellcolor{ly}+2.0 & \cellcolor{ly}+6.7  \\
        \midrule
        %%%%%%%%%%%%%%%%%%%%%%%%%%%%%%%%%%%%%%%%%%%%%%%%%%%%%%%%%%%%%%%%%%%%%%%%%%%%%%%%%%%%%%%%%
        \multicolumn{1}{c}{} &\multicolumn{1}{c|}{} & \multicolumn{13}{c}{\oursgtp}  \\
        \midrule   
        \sclub & N/A               & 56.2 & 23.0 & \cellcolor{ly}\underline{41.7} & 11.4  & 22.1 & \cellcolor{ly}+13.0 & 45.0 & \cellcolor{lb}32.0 & 30.0 & 10.5 & 17.4 & -28.0 & -7.5  \\
        \sclub & AP                & 50.0 & 20.0 & \cellcolor{lb}25.0 & 9.5  & 24.3 & -11.7 & 45.0 & 31.0 & \cellcolor{ly}50.0 & 15.9 & 24.4 & -8.0 &  -9.9 \\
        \sclub & SP                & 45.0 & 25.0 & \cellcolor{ly}38.3 & 11.8 & 22.6 & \cellcolor{lb}+7.9 & 42.5 & \cellcolor{lb}32.0 & \cellcolor{ly}50.0 & 11.4 & 22.5 & -14.4 &  -3.3 \\
        \sclub & DGI               & 56.2 & \cellcolor{lb}26.0 & \cellcolor{ly}40.0 & 17.3 & 17.7 & \cellcolor{ly}+16.6 & 37.5 & 31.0 & \cellcolor{ly}45.0 & 13.6 & 18.7 & -18.9 & -1.2  \\
        % \sclub & CG                & 56.2 & 19.0 & \cellcolor{lb}26.7 & 6.8   & 24.1 & -11.7 & 48.8 & 27.0 & 36.7 & 17.3 & 19.9 & -17.3 & -14.5  \\
        % \midrule   
        \midrule 
        %%%%%%%%%%%%%%%%%%%%%%
        \sclub \sspade & N/A           & \cellcolor{ly}\underline{73.8} & \cellcolor{lb}31.0 & \cellcolor{lb}28.3 & 8.2  & 22.5 & \cellcolor{lb}+5.2 & \cellcolor{ly}62.5 & 29.0 & \cellcolor{ly}38.3 & 13.2 & 19.8 & -15.5 & -5.2  \\
        \sclub \sspade & AP            & 62.5 & \cellcolor{ly}32.0 & \cellcolor{ly}\textbf{46.7} & 12.3 & 21.1 & \cellcolor{ly}+28.1 & 58.8 & 30.0 & \cellcolor{ly}40.0 & 10.9 & 29.2 & -12.4 & \cellcolor{ly}+7.9  \\
        \sclub \sspade & SP            & 65.0 & \cellcolor{ly}32.0 & \cellcolor{ly}\underline{41.7} & 12.3 & 23.5 & \cellcolor{ly}+24.6 & \cellcolor{ly}62.5 & \cellcolor{lb}32.0 & \cellcolor{ly}\underline{51.7} & \cellcolor{ly}21.8 & 23.5 & \cellcolor{ly}+5.2 & \cellcolor{ly}+14.9  \\
        \sclub \sspade & DGI           & \cellcolor{ly}\textbf{75.0} & \cellcolor{lb}27.0 & \cellcolor{lb}33.3   & 17.3   & 24.3 & \cellcolor{ly}+19.7 &  \cellcolor{ly}62.5 & 31.0 & \cellcolor{ly}46.7 & \cellcolor{lb}19.5 & 24.7 & \cellcolor{ly}+0.1 & \cellcolor{ly}+9.9  \\
        % \sclub \sspade & CG            & \cellcolor{ly}70.0 & \cellcolor{ly}34.0 & \cellcolor{ly}36.7 & 5.5  & 18.5 & \cellcolor{ly}+10.6 & \cellcolor{ly}62.5 & 28.0 & 36.7 & 11.8 & 22.5 & -17.0 & -3.2  \\
        \midrule   
        \multicolumn{1}{c}{} &\multicolumn{1}{c|}{} & \multicolumn{13}{c}{\oursgtf}  \\
        \midrule 
        %%%%%%%%%%%%%%%%%%%%%%%%%%%%%%%%%%%%%%%%%%%%%%%%%%%%%%%%%%%%%%%%%%%%%%%%%%%%%%%%%%%%%%%%%
        \sdiamond & N/A               & \cellcolor{ly}83.8 & \cellcolor{ly}53.0 & \cellcolor{lb}33.3 & \cellcolor{ly}23.6 & 24.8 & \cellcolor{ly}+49.7 & \cellcolor{ly}100.0 & \cellcolor{ly}90.0 & \cellcolor{ly}68.3 & \cellcolor{ly}37.3 & \cellcolor{ly}52.7 & \cellcolor{ly}+96.5 & \cellcolor{ly}+73.1 \\
        \sdiamond & AP                & \cellcolor{ly}85.0 & \cellcolor{ly}39.0 & \cellcolor{lb}26.7 & \cellcolor{ly}26.4 & 27.5 & \cellcolor{ly}+36.4 & \cellcolor{ly}92.5 & \cellcolor{ly}88.0 & \cellcolor{ly}63.3 & \cellcolor{ly}53.6 & \cellcolor{ly}51.6 & \cellcolor{ly}+108.0 & \cellcolor{ly}+72.2 \\
        \sdiamond & SP                & 48.7 & \cellcolor{ly}61.0 & \cellcolor{ly}46.7 & \cellcolor{ly}23.6 & \cellcolor{ly}28.9 & \cellcolor{ly}+64.2 & \cellcolor{ly}95.0 & \cellcolor{ly}95.0 & \cellcolor{ly}70.0 & \cellcolor{ly}37.3 & \cellcolor{ly}52.8 & \cellcolor{ly}+99.0 & \cellcolor{ly}+81.6 \\
        \sdiamond & DGI               & \cellcolor{ly}85.0 & \cellcolor{lb}27.0 & \cellcolor{lb}31.7 & 14.1 & 22.1 & \cellcolor{ly}+15.7 & \cellcolor{ly}100.0 & \cellcolor{ly}40.0 & \cellcolor{ly}70.0 & \cellcolor{ly}31.8 & \cellcolor{ly}50.6 & \cellcolor{ly}+58.7 & \cellcolor{ly}+37.2 \\
        \midrule 
        %%%%%%%%%%%%%%%%%%%%%%
        \sdiamond \sspade & N/A           & \cellcolor{ly}92.5 & \cellcolor{ly}39.0 & \cellcolor{lb}30.0 & 15.9 & 23.6 & \cellcolor{ly}+28.3 & \cellcolor{ly}96.3 & \cellcolor{ly}56.0 & \cellcolor{ly}55.0 & 14.5 & \cellcolor{ly}46.6 & \cellcolor{ly}+37.9 & \cellcolor{ly}+33.1 \\
        \sdiamond \sspade & AP            & \cellcolor{ly}73.8 & \cellcolor{ly}36.0 & \cellcolor{ly}46.7 & \cellcolor{ly}25.9 & 23.9 & \cellcolor{ly}+51.5 & \cellcolor{ly}85.0 & \cellcolor{ly}42.0 & \cellcolor{ly}68.3 & \cellcolor{ly}36.4 & \cellcolor{ly}47.5 & \cellcolor{ly}+57.7 & \cellcolor{ly}+54.6 \\
        \sdiamond \sspade & SP            & 62.5 & 24.0 & \cellcolor{ly}36.7 & 14.5 & \cellcolor{ly}28.6 & \cellcolor{ly}+17.7 & 60.0 & \cellcolor{ly}43.0 & \cellcolor{ly}46.7 & \cellcolor{ly}25.0 & \cellcolor{ly}47.9 & \cellcolor{ly}+26.4 & \cellcolor{ly}+21.1 \\
        \sdiamond \sspade & DGI           & \cellcolor{ly}81.2 & \cellcolor{lb}30.0 & \cellcolor{lb}25.0 & 16.8 & \cellcolor{ly}30.7 & \cellcolor{ly}+18.0 & \cellcolor{ly}73.8 & \cellcolor{ly}39.0 & \cellcolor{ly}48.3 & 15.0 & \cellcolor{ly}40.6 & \cellcolor{ly}+13.6 & \cellcolor{ly}+15.8 \\
        \midrule   
        \bottomrule

    \end{tabular}
\end{table}

\clearpage

\subsection{Other Remarks}
\label{appendix:other_remarks}

\paragraph{Pre-training graph encoder helps.}
In Table~\ref{tab:train_score_full} and Table~\ref{tab:test_score_full}, we also show \ours, \oursgtp and \oursgtf's training and test scores when their action scorer's graph encoder are initialized with pre-trained parameters as introduced in Section~\ref{subsection:graph_update} (for \ours) and Appendix~\ref{appendix:pretrain_discrete_graph_encoder} (for \oursgtp and \oursgtf).
We observe in most settings, pre-trained graph encoders produce better training and test results compared to their randomly initialized counterparts.
This is particularly obvious in \oursgtp and \oursgtf, where graphs are discrete. 
For instance, from Table~\ref{tab:test_score_full} we can see that only with text observation as additional input (\sclub\sspade), and when graph encoder are initialized with AP/SP/DGI, the \oursgtp agent can outperform the text-based baselines on test game sets.

\paragraph{Fine-tuning graph encoder helps.}
For all experiment settings where the graph encoder in action scorer is initialized with pre-trained parameters (OG/COC for \ours, AP/SP/DGI for \oursgtp), we also compare between freezing vs. fine-tuning the graph encoder in RL training. 
By freezing the graph encoders, we can effectively reduce the number of parameters to be optimized with RL signal.
However, we see consistent trends that fine-tuning the graph encoders can always provide better training and testing performance in both \ours and \oursgtp.

\paragraph{Text input helps more when graphs are imperfect.}
We observe clear trends that for \oursgtf, using text together with graph as input (to the action selector) does not provide obvious performance increase.
Instead, \oursgtf often shows better performance when text observation input is disabled.
This observation is coherent with the intuition of using text observations as additional input.
When the input graph to the action selector is imperfect (e.g., belief graph maintained by \ours or \oursgtp itself), the text observation provides more accurate information to help the agent to recover from errors.
On the other hand, \oursgtf uses the ground-truth full graph (which is even accurate than text) as input to the action selector, the text observation might confuse the agent by providing redundant information with more uncertainty.

\paragraph{Learning across difficulty levels.}
We have a special set of RL games --- level 5 --- which is a mixture of the other four difficulty levels. 
We use this set to evaluate an agent's generalizability on both dimensions of game configurations and difficulty levels.
From Table~\ref{tab:train_score_full}, we observe that almost all agents (including baseline agents) benefit from a larger training set, i.e., achieve better test results when train on 100 level 5 games than 20 of them.
Results show \ours has a more significant performance boost from larger training set. 
We notice that all \oursgtp variants perform worse than text-based baselines on level 5 games, whereas \ours outperforms text-based baselines when training on 100 games.
This may suggest the continuous belief graphs can better help \ours to adapt to games across difficulty levels, whereas its discrete counterpart may struggle more.
For example, both games in level 1 and 2 have only single location, while level 3 and 4 games have multiple locations.
\oursgtp might thus get confused since sometimes the direction relations (e.g., \cmd{west\_of}) are unused.
In contrast, \ours, equipped with continuous graphs, may learn such scenario easier.

\subsection{Probing Task and Belief Graph Visualization}
\label{appendix:highres_visualization}

In this section, we investigate whether generated belief graphs contain any useful information about the game dynamics. We first design a probing task to check if $\mathcal{G}$ encodes the existing relations between two nodes. Next, we visualize a few slices of the adjacency tensor associated to $\mathcal{G}$.

\subsubsection{Probing Task}
We frame the probing task as a multi-label classification of the relations between a pair of nodes.
Concretely, given two nodes $i$, $j$, and the vector $\mathcal{G}_{i, j} \in [-1, 1]^\mathcal{R}$ (in which $\mathcal{R}$ denotes the number of relations) extracted from the belief graph $\mathcal{G}$ corresponding to the nodes $i$ and $j$, the task is to learn a function $f$ such that it minimizes the following binary cross-entropy loss:
\begin{equation}
    \mathcal{L}_{\text{BCE}}\left(f(\mathcal{G}_{i, j}, h_i, h_j), Y_{i,j}\right),
\end{equation}
where $h_i$, $h_j$ are the embeddings for nodes $i$ and $j$, $Y_{i,j} \in \{0, 1\}^\mathcal{R}$ is a binary vector representing the presence of each relation between the nodes (there are $R$ different relations). 
Following \citet{Alain2017UnderstandingIL}, we use a linear function as $f$, since we assume the useful information should be easily accessible from $\mathcal{G}$.

We collect a dataset for this probing task by following the walkthroughs of 120 games.
At every game step, we collect a tuple $(\mathcal{G}, \mathcal{G}^\text{seen})$ (see Appendix~\ref{appendix:train_discrete_graph_updater} for the definition of $\mathcal{G}^\text{seen}$).
We used tuples from 100 games as training data and the remaining for evaluation.

From each tuple in the dataset, we extract several node pairs $(i, j)$ and their corresponding $Y_{i,j}$ from $\mathcal{G}^\text{seen}$ (positive examples, denoted as ``\textcolor{red1}{$+$}''). 
To make sure a model can only achieve good performance on this probing task by using the belief graph $\mathcal{G}$, without overfitting by memorising node-relation pairs (e.g., the unique relation between player and kitchen is \cmd{at}), we augment the dataset by adding plausible node pairs (i.e., $Y_{i,j} = \vec{0}^\mathcal{R}$) but that have no relation according to the current $\mathcal{G}$ (negative examples, denoted as ``\textcolor{green1}{$-$}'').
For instance, if at a certain game step the player is in the bedroom, the relation between the player and kitchen should be empty ($\vec{0}^\mathcal{R}$). We expect $\mathcal{G}$ to have captured that information.

We use two metrics to evaluate the performance on this probing task:\\
\textbf{$\bullet$\quad Exact match~} represents the percentage of predictions that have all their labels classified correctly, i.e., when $f(\mathcal{G}_{i, j},h_i,h_j) = Y^n_{i,j}$.\\
\textbf{$\bullet$\quad $\text{F}_1$ score~} which is the harmonic mean between precision and recall. We report the macro-averaging of $\text{F}_1$ over all the predictions.

To better understand the probe's behaviors on each settings, we also report their training and test performance on the positive samples (\textcolor{red1}{$+$}) and negative samples (\textcolor{green1}{$-$}) separately.

\begin{table}[]
    \small
    \centering
    \caption{Probing task results showing that belief graphs obtained from OG and COC do contain information about the game dynamics, i.e. node relationships.}
    \label{tab:probing_task}
    \vspace{0.8em}
    \begin{tabular}{l|ccc|ccc|ccc|ccc}
    \toprule
     & \multicolumn{6} {c|} {Exact Match} & \multicolumn{6} {c} {$\text{F}_1$ score} \\
    %\midrule
    \hline
     & \multicolumn{3}{c|}{Train} & \multicolumn{3}{c|}{Test} & \multicolumn{3}{c|}{Train} & \multicolumn{3}{c}{Test} \\
    Model & \pc\textcolor{red1}{$+$} & \nc\textcolor{green1}{$-$} & \oc Avg & \pc\textcolor{red1}{$+$} & \nc\textcolor{green1}{$-$} & \oc Avg & \pc\textcolor{red1}{$+$} & \nc\textcolor{green1}{$-$} & \oc Avg & \pc\textcolor{red1}{$+$} & \nc\textcolor{green1}{$-$} & \oc Avg \\
    %\midrule
    \hline
    Random &            \pc0.00 & \nc0.99 & \oc0.49 & \pc0.00 & \nc0.99 & \oc0.49 & \pc0.00 & \nc0.99 & \oc0.49 & \pc0.00 & \nc0.99 & \oc0.49 \\
    Ground-truth &      \pc0.98 & \nc0.96 & \oc0.97 & \pc0.97 & \nc0.96 & \oc0.97 & \pc0.98 & \nc0.96 & \oc0.97 & \pc0.98 & \nc0.96 & \oc0.97 \\
    %\midrule
    \hline
    \transdrqn &        \pc0.61 & \nc0.84 & \oc0.73 & \pc0.61 & \nc0.83 & \oc0.72 & \pc0.61 & \nc0.84 & \oc0.73 & \pc0.61 & \nc0.83 & \oc0.72  \\
    \ours (OG) &                \pc0.69 & \nc0.86 & \oc\textbf{0.78} & \pc0.70 & \nc0.86 & \oc\textbf{0.78} & \pc0.71 & \nc0.85 & \oc\textbf{0.78} & \pc0.72 & \nc0.86 & \oc\textbf{0.79} \\
    \ours (COC) &               \pc0.65 & \nc0.86 & \oc0.75 & \pc0.65 & \nc0.84 & \oc0.75 & \pc0.67 & \nc0.85 & \oc0.76 & \pc0.67 & \nc0.84 & \oc0.75 \\
    \bottomrule
    \end{tabular}
\end{table}

From Table~\ref{tab:probing_task}, we observe that belief graphs $\mathcal{G}$ generated by models pre-trained with either OG or COC do contain useful information about the relations between a pair of nodes. 
We first compare against a random baseline where each $\mathcal{G}$ is randomly sampled from $\mathcal{N}(0,1)$ and kept fixed throughout the probing task. 
We observe the linear probe fails to perform well on the training set (and as a result also fails to generalize on test set). 
Interestingly, with random belief graphs provided, the probe somehow overfits on negative samples and always outputs zeros all the time.
In both training and testing phases, it produces zero performance on positive examples.
This baseline suggests the validity of our probing task design --- there is no way to correctly predict the relations without having the information encoded in the belief graph $\mathcal{G}$.

Next, we report the performance of using ground-truth graphs ($\mathcal{G}^\text{seen}$) as input to $f$. 
We observe the linear model can perform decently on training data, and can generalize from training to testing data --- on both sets, the linear probe achieves near-perfect performance.
This also verifies the probing task by showing that given the ground-truth knowledge, the linear probe is able to solve the task easily.

Given the two extreme cases as upper bound and lower bound, we investigate the belief graphs $\mathcal{G}$ generated by \ours, pre-trained with either of the two self-supervised methods, OG and COC (proposed in Section~\ref{subsection:graph_update}).
From Table~\ref{tab:probing_task}, we can see $\mathcal{G}$ generated by both OG and COC methods help similarly in the relation prediction task, both provide more than 75\% of testing exact match scores.

In Section~\ref{section:exp}, we show that \ours outperforms a set of baseline systems, including \transdrqn (as described in Section~\ref{subsection:baselines}), an agent with recurrent components.
To further investigate if the belief graphs generated by \ours can better facilitate the linear probe in this relation prediction task, we provide an additional setting.
We modify the \transdrqn agent by replacing its action scorer by a text generator (the same decoder used in OG training), and train this model with the same data and objective as OG.
After pre-training, we obtain a set of probing data by collecting the recurrent hidden states produced by this agent given the same probing game walkthroughs.
Since these recurrent hidden states are computed from the same amount of information as \ours's belief graphs, they could theoretically contain the same information as $\mathcal{G}$.
However, from Table~\ref{tab:probing_task}, we see that the scores of \transdrqn are consistently lower than \ours's score.
This is coherent with our findings in the RL experiments (Section~\ref{section:exp}), except the gap between \ours and \transdrqn is less significant in the relation prediction task setting.

While being able to perform the classification correctly in a large portion of examples, we observe a clear performance gap comparing \ours's belief graphs with ground-truth graphs.
The cause of the performance gap can be twofold. 
First, compared to ground-truth graphs that accurately represent game states without information loss, $\mathcal{G}$ (iteratively generated by a neural network across game steps) can inevitably suffer from information loss. 
Second, the information encoded in $\mathcal{G}$ might not be easily extracted by a linear probe (compared to ground-truth).
Both aspects suggest potential future directions to improve the belief graph generation module.

We optimize all probing models for 10 epochs with Adam optimizer, using the default hyperparameters and a learning rate of $0.0001$. 
Note in all the probing models, only parameters of the linear layer $f$ are trainable, everything else (including node embeddings) are kept fixed.

% \begin{table}[]
%     \centering
%     \caption{Probing task results showing that belief graphs obtained from OG and COC do contain information about the game dynamics, i.e. node relationships.}
%     \label{tab:probing_task}
%     \vspace{0.8em}
%     \begin{tabular}{lcccc}
%     \toprule
%      & \multicolumn{2} {c} {Exact Match} & \multicolumn{2} {c} {$\text{F}_1$ score} \\
%     \midrule
%     Model & Train & Test & Train & Test \\
%     \midrule
%     Random & 0.427 & 0.089 & 0.504 & 0.159 \\
%     Ground-truth & 0.783 & 0.715 & 0.797 & \textbf{0.734} \\
%     \midrule
%     OG & 0.542 & 0.371 & 0.593 & 0.459 \\
%     COC & 0.542 & 0.207 & 0.593 & 0.254 \\
%     \midrule
%     OG w/ Graph Encoder & 0.740 & \textbf{0.718} & 0.753 & \textbf{0.734} \\
%     COC w/ Graph Encoder & 0.745 & 0.706 & 0.755 & 0.715 \\
%     \bottomrule
%     \end{tabular}
% \end{table}

\subsubsection{Belief Graph Visualization}
\begin{figure}
    \centering
    \includegraphics[width=\textwidth]{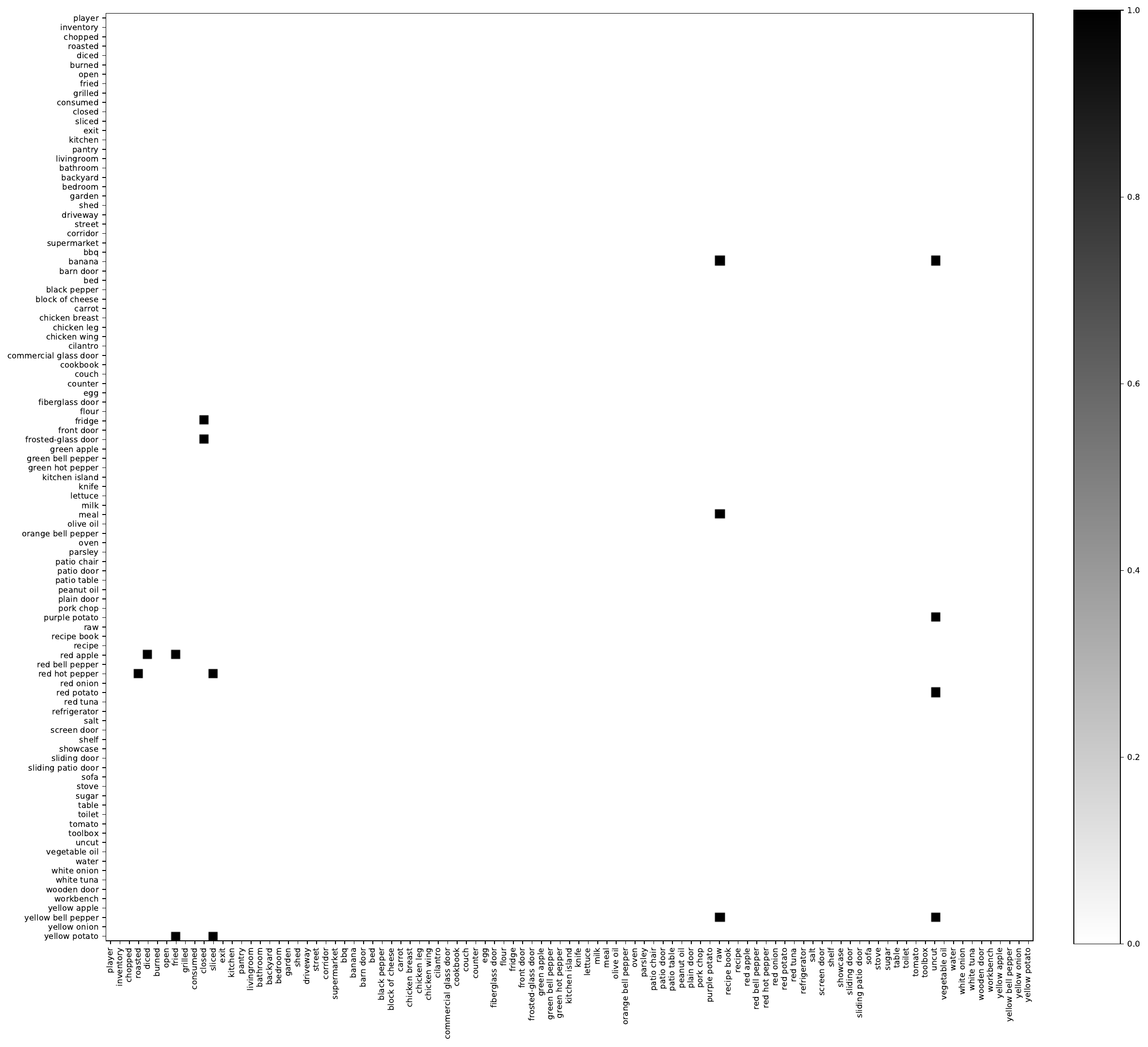}
    \caption{Slice of a ground-truth adjacency tensor representing the \cmd{is} relation.}
    \label{fig:heatmap_gt}
\end{figure}

\begin{figure}
    \centering
    \includegraphics[width=0.9\textwidth]{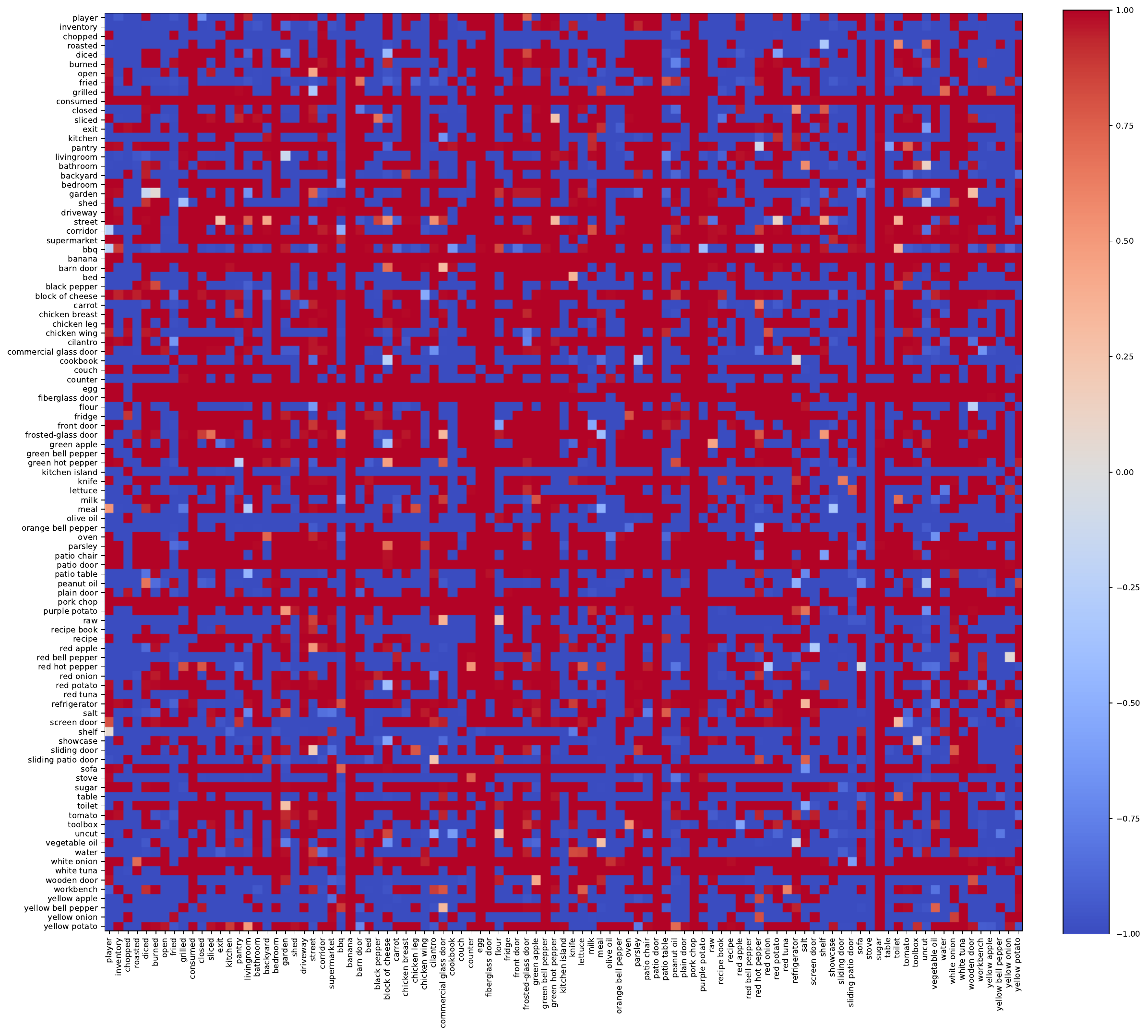}
    \includegraphics[width=0.9\textwidth]{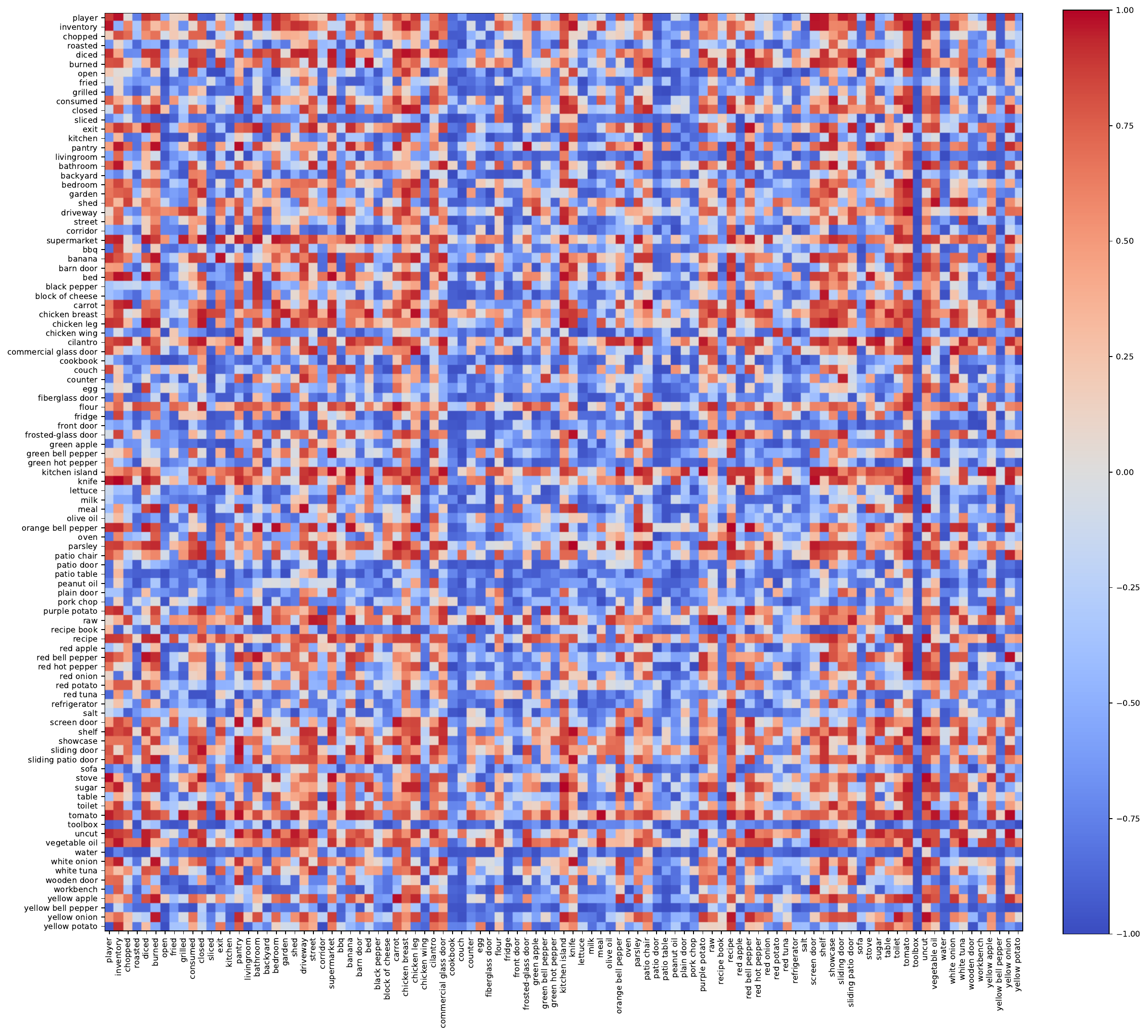}
    \caption{Adjacency tensor's slices for $\mathcal{G}$ generated by \ours, pre-trained with OG task  \textbf{(top)} and COC task \textbf{(bottom)}.}
    \label{fig:heatmap_og_coc_original}
\end{figure}

In Figure~\ref{fig:heatmap_gt}, we show a slice of the ground-truth adjacency tensor representing the \cmd{is} relation. To give context, that tensor has been extracted at the end of a game with a recipe requesting a fried diced red apple, a roasted sliced red hot pepper, and a fried sliced yellow potato.
Correspondingly, for the same game and same time step, Figure~\ref{fig:heatmap_og_coc_original} shows the same adjacency tensor's slice for the belief graphs $\mathcal{G}$ generated by \ours pre-trained on observation generation (OG) and contrastive observation classification (COC) tasks.

\begin{figure}
    \centering
    \includegraphics[width=0.9\textwidth]{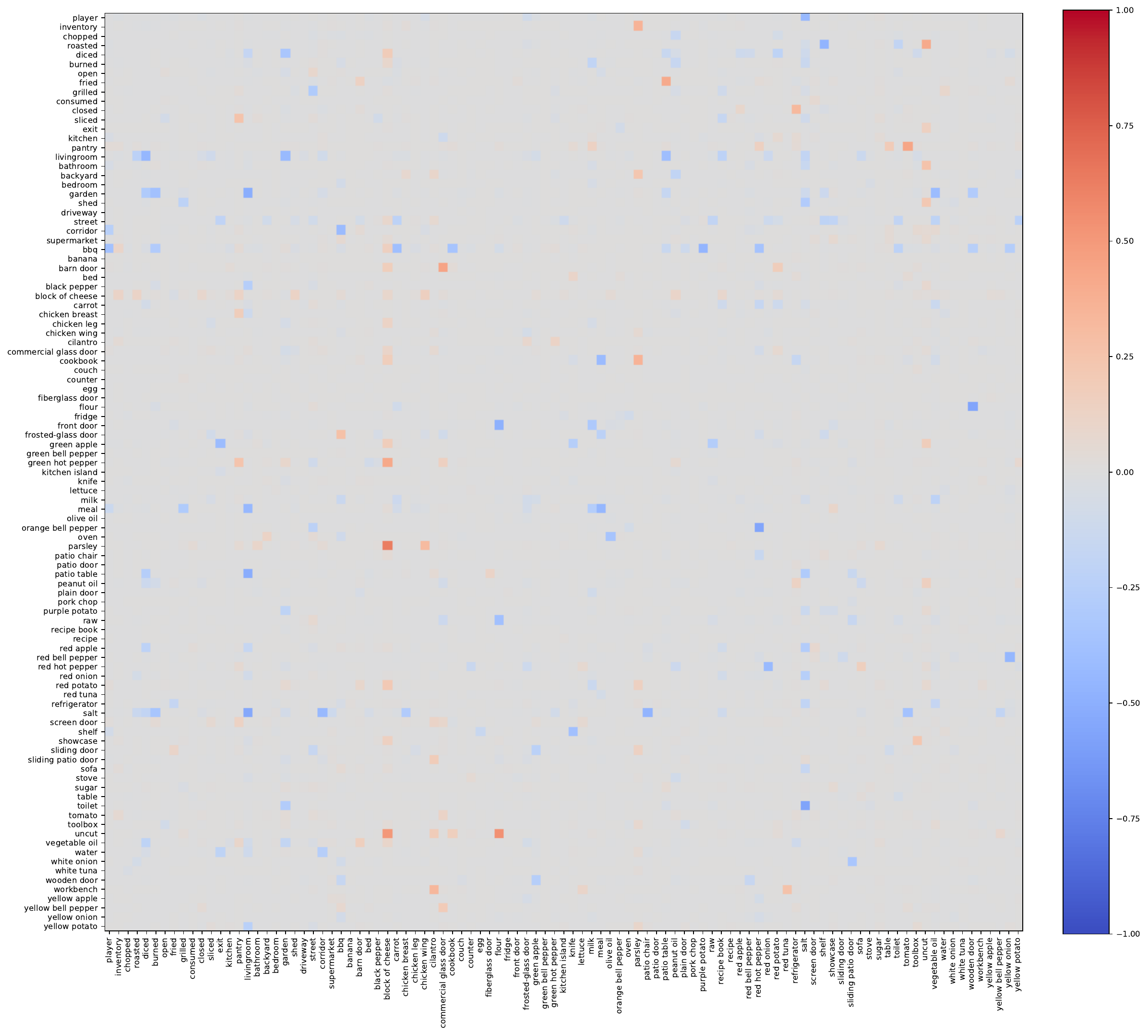}
    \includegraphics[width=0.9\textwidth]{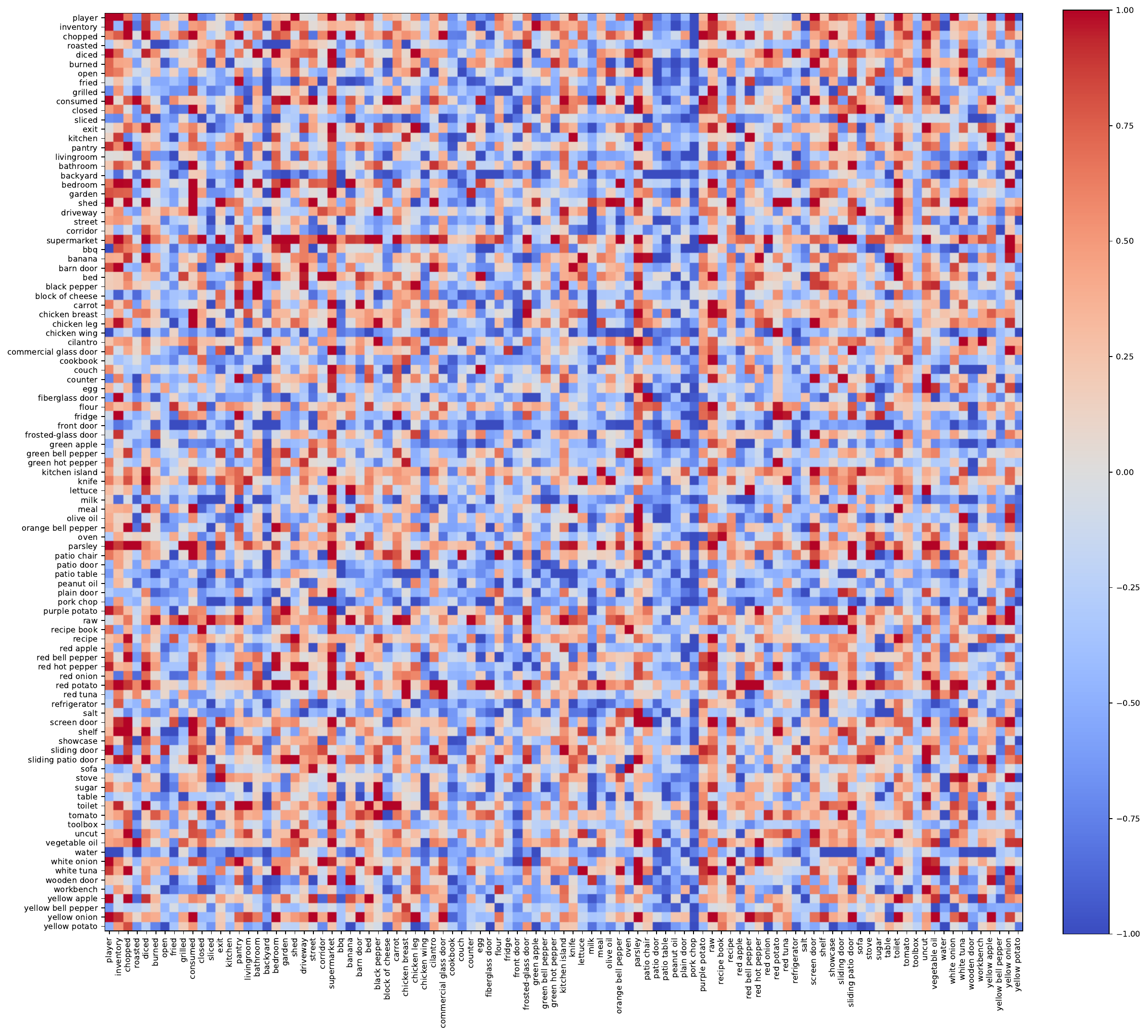}
    \caption{Adjacency tensor's slices after subtracting the mean for $\mathcal{G}$ generated by \ours, pre-trained with OG task  \textbf{(top)} and COC task \textbf{(bottom)}.}
    \label{fig:heatmap_og_coc}
\end{figure}

For visualization, we found that subtracting the mean adjacency tensor, computed across all games and steps, helps by removing information about the marginal distribution of the observations (e.g., underlying grammar or other common features needed for the self-supervised tasks). Those ``cleaner'' graphs are shown in Figure~\ref{fig:heatmap_og_coc}.
One must keep in mind that there is no training signal to force the belief graphs to align with any ground-truth graphs since the belief graph generators are trained with pure self-supervised methods.

% \begin{figure}
%     \centering
%     \includegraphics[width=0.8\textwidth]{figures/heatmap_infomax.pdf}
%     \caption{An example slice of $\mathcal{G}$ generated by \ours, pre-trained with COC task.}
%     \label{fig:heatmap_coc_original}
% \end{figure}
\clearpage

% \begin{figure}
%     \centering
%     \includegraphics[width=\textwidth]{figures/heatmap_infomax_centered.pdf}
%     \caption{An example slice of $\mathcal{G}$ generated by \ours, pre-trained with COC task. Baseline subtracted.}
%     \label{fig:heatmap_coc}
% \end{figure}
\clearpage

\subsection{\ours's Performance on Graph Updater Pre-training Tasks}

In this subsection, we report the test performance of the two graph updater pre-training tasks for \ours.

In the observation generation (OG) task, our model achieves a test loss of 22.28 (when using ground-truth history tokens as input to the decoder), and a test $F_1$ score of 0.68 (when using the previously generated token as input to the decoder).

In the contrastive observation classification (COC) task, our model achieves a test loss of 0.06, and a test accuracy of 0.97.

In Table~\ref{tab:cherry_pick_og}, we show a selected test example from the observation generation task. In the example, the prediction has fair amount of overlap (semantically, not necessarily word by word) with the ground-truth. 
This suggests the belief graph $\mathcal{G}^{\text{belief}}_t$ generated by \ours's updater model can to some extent capture and encode the state information of the environment --- since the model is able to reconstruct $O_t$ using $\mathcal{G}^{\text{belief}}_t$.

\begin{table}[t!]
    \centering
    \scriptsize
    \caption{Selected example of test set of the observation generation task. On this specific data point, The model gets an $F_1$ score of 0.64. We use same colors to shade the information overlap between ground-truth observation with the model prediction. }
    \label{tab:cherry_pick_og}
    \vspace{0.8em}
    \begin{tabular}{l}
        \toprule
        \multicolumn{1}{c}{Ground-Truth} \\
        \midrule
        \colorbox{lred}{-= kitchen = - of every kitchen you could have sauntered into , you had to saunter into a normal one . } \colorbox{lgreen}{you can make out a fridge .} \\
        the fridge contains a raw chicken wing , \colorbox{lgreen}{a raw chicken leg} and a block of cheese . \colorbox{lblue}{you can make out an oven .} \colorbox{lorange}{you hear a noise} \\
        \colorbox{lorange}{behind you and spin around , but you ca n't see anything other than a table .} wow , is n't textworld just the best ? you see a cookbook\\
        on the table . hmmm ...  what else , what else ? \colorbox{lpurple}{you make out a counter} . now why would someone leave that there ? the counter is vast . \\
        on the counter you make  out a raw purple potato and a knife . \colorbox{lcyan}{you can make out a stove . the stove is conventional . unfortunately ,} \\
        \colorbox{lcyan}{there is n't a thing on it .} hopefully this does n't make you too upset . \colorbox{lmagenta}{there is an open frosted - glass door leading east .} there is an \\
        exit to the north . do n't worry , there is no door . you need an exit without a door ? you should try going west . there is a white onion \\
        on the floor .\\
        \midrule
        \multicolumn{1}{c}{Prediction} \\
        \midrule
         \colorbox{lred}{-= kitchen = - you 're now in a kitchen . } you begin looking for stuff . \colorbox{lgreen}{you can make out a fridge . the fridge contains a raw chicken} \\
        \colorbox{lgreen}{leg .} \colorbox{lblue}{you can make out a} closed \colorbox{lblue}{oven} . \colorbox{lorange}{you can make out a table} . the table is massive . on the table you see a knife . \colorbox{lpurple}{you see a } \\
        \colorbox{lpurple}{counter .} the counter is vast . on the counter you can make out a cookbook . \colorbox{lcyan}{you can make out a stove . the stove is conventional .} \\
        \colorbox{lcyan}{but the thing has n't got anything on it .} you can make out a stove . the stove is conventional . but the thing has n't got anything on it .\\
        \colorbox{lmagenta}{there is an open frosted - glass door leading east .} there is an exit to the east .\\
        \bottomrule
    \end{tabular}
\end{table}

\subsection{Future Directions}
\label{appendix:future_work}

%\paragraph{Sparsity and Interpretability}
As mentioned in Appendix~\ref{appendix:highres_visualization}, the belief graphs generated by \ours lack of interpretability because the training is not supervised by any ground-truth graph.
Technically, they are recurrent hidden states that encode the game state, we only (weakly) ground these real-valued graphs by providing node and relation vocabularies (word embeddings) for the message passing in R-GCN.

Therefore, there can be two potential directions deriving from the current approach.
First, it would be interesting to investigate regularization methods and auxiliary tasks that can make the belief graph sparser (without relying on ground-truth graphs to train).
A sparser belief graph may increase \ours's interpretability, however, it does not guarantee to produce better performance on playing text-based games (which is what we care more about).

Second, it would also be interesting to see how \ours can be adapted to environments where the node and relation names are unknown.
This will presumably make the learned belief graphs even far away from interpretable, but at the same time it will further relax \ours from the need of requiring any prior knowledge about the environments.
We believe this is an essential property for an agent that is generalizabile to out-of-distribution environments.
For instance, without the need of a pre-defined node and relation vocabularies, we can expand \ours to the setting where training on the cooking games, and testing on games from another genre, or even text-based games designed for humans \citep{hausknecht19jericho}.

\section{Implementation Details}
\label{appendix:implementation_details}

In Appendix~\ref{appendix:model}, we have introduced hyperparameters regarding model structures.
In this section, we provide hyperparameters we used in training and optimizing.

In all experiments, we use \emph{Rectified Adam} \citep{liu2019variance} as the step rule for optimization. The learning rate is set to 0.001. 
We clip gradient norm of 5.

\subsection{Graph Updater}
\label{appendix:implementation_details_graph_updater}

To pre-train the recurrent graph updater in \ours, we utilize the backpropagation through time (BPTT) algorithm. 
Specifically, we unfold the recurrent graph updater and update every 5 game steps.
We freeze the parameters in graph updater after its own training process.
Although it can theoretically be trained on-the-fly together with the action selector (with reward signal), we find the standalone training strategy more effective and efficient.

\subsection{Action Selector}
\label{appendix:implementation_details_action_selector}

The overall training procedure of \ours's action selector is shown in Algorithm~\ref{alg:dqn}.

\begin{algorithm}[t]
  \caption{Training Strategy for \ours Action Selector}
  \label{alg:dqn}
\small
\begin{algorithmic}[1]
  \State {\bfseries Input:} games $\mathcal{X}$, replay buffer $B$,
  update frequency $F$, patience $P$, tolerance $\tau$, evaluation frequency $E$.
  \State Initialize counters $k\gets1, p\gets0$, best validation score $V\gets0$, transition cache $C$, policy $\pi$, checkpoint $\Pi$.
  \For{$e\gets1$ {\bfseries to} NB\_EPISODES}
  \State Sample a game $x \in \mathcal{X}$, reset $C$.
  \For{$i\gets1$ {\bfseries to} NB\_STEPS}
  \State play game, push transition into $C$, $k \gets k+1$
  \State \algorithmicif{ $k \% F = 0$ } \algorithmicthen{ sample batch from $B$, $\text{Update}(\pi)$ }
  \State \algorithmicif{ done } \algorithmicthen{ break }
  \EndFor
    % selective replay buffer pushing
  \If{$\text{average score in } C > \tau \cdot \text{average score in } B$}
  \State \algorithmicforall{ item in $C$ } \algorithmicdo{ push item into $B$ }
  \EndIf
  % auto-reload
  \State \algorithmicif{ $e\%E \neq 0$ } \algorithmicthen{ continue }
  \State $v\gets\text{Evaluate}(\pi)$
  \State \algorithmicif{ $v >= V$ } \algorithmicthen{ $\Pi \gets \pi$, $p\gets0$, continue }
  \State \algorithmicif{ $p > P$ } \algorithmicthen{ $\pi \gets \Pi$, $p\gets0$ }
  \State \algorithmicelse{ $p \gets p + 1$ }
  \EndFor
\end{algorithmic}
\end{algorithm}

We report two strategies that we empirically find effective in DQN training.
First, we discard the underachieving trajectories without pushing them into the replay buffer (lines 10--12). 
Specifically, we only push a new trajectory that has an average reward greater than $\tau\in\mathbb{R}^+_0$ times the average reward for all transitions in the replay buffer. 
We use $\tau=0.1$, since it keeps around some weaker but acceptable trajectories and does not limit exploration too severely.
Second, we keep track of the best performing policy $\Pi$ on the validation games. 
During training, when \ours stops improving on validation games, we load $\Pi$ back to the training policy $\pi$ and resume training.
After training, we report the performance of $\Pi$ on test games.
Note these two strategies are not designed specifically for \ours; rather, we find them effective in DQN training in general.

We use a prioritized replay buffer with memory size of 500,000, and a priority fraction of 0.6. 
We use $\epsilon$-greedy, where the value of $\epsilon$ anneals from 1.0 to 0.1 within 20,000 episodes.
We start updating parameters after 100 episodes of playing.
We update our network after every 50 game steps (update frequency $F$ in Algorithm~\ref{alg:dqn}) \footnote{50 is the total steps performed within a batch. For instance, when batch size is 1, we update per 50 steps; whereas when batch size is 10, we update per 5 steps. Note the batch size here refers to the parallelization of the environment, rather than the batch size for backpropagation.}.
During update, we use a mini-batch of size 64. 
We use a discount $\gamma = 0.9$.
We update target network after every 500 episodes.
For multi-step learning, we sample the multi-step return $n \sim \text{Uniform}[1, 3]$.
We refer readers to \citet{hessel18rainbow} for more information about different components of DQN training.

In our implementation of the \transdrqn and \transdrqnp baselines, following \citet{yuan2018counting}, we sample a sequence of transitions of length 8, use the first 4 transitions to estimate reasonable recurrent states and use the last for to update.
For counting bonus, we use a $\gamma_c = 0.5$, the bonus is scaled by an coefficient $\lambda_c = 0.1$.

For all experiment settings, we train agents for 100,000 episodes (NB\_EPISODES in Algorithm~\ref{alg:dqn}).
For each game, we set maximum step of 50 (NB\_STEPS in Algorithm~\ref{alg:dqn}). 
When an agent has used up all its moves, the game is forced to terminate.
We evaluate them after every 1,000 episodes (evaluation frequency $E$ in Algorithm~\ref{alg:dqn}).
Patience $P$ and tolerance $\tau$ in Algorithm~\ref{alg:dqn} are 3 and 0.1, respectively.
The agents are implemented using PyTorch \citep{paszke17automatic}.

\subsection{Wall Clock Time}
\label{appendix:wall_clock_time}

We report our experiments' wall clock time. 
We run all the experiments on single NVIDIA P40/P100 GPUs.

\begin{table}[h!]
    \scriptsize
    \centering
    \caption{Wall clock time for all experiments.}
    \label{tab:wall_clock_time}
    \vspace{0.8em}
    \begin{tabular}{c|c|c}
        \toprule
        Setting/Component & Batch Size & Approximate Time \\
        \midrule   
        \multicolumn{3}{c}{\ours} \\
        \midrule 
        Graph Updater - OG (Section~\ref{subsection:graph_update})
        & 48 & 2 days \\
        \midrule 
        Graph Updater - COC (Section~\ref{subsection:graph_update})
        & 64 & 2 days \\
        \midrule 
        Action Scorer (Section~\ref{subsection:action_selection})
        & 64 (backprop) & 2 days \\
        \midrule   
        \multicolumn{3}{c}{\oursgtp} \\
        \midrule 
        Discrete Graph Updater (Appendix~\ref{appendix:train_discrete_graph_updater})
        & 128 & 2 day \\
        \midrule 
        Action Scorer (same as (Section~\ref{subsection:action_selection}))
        & 64 (backprop) & 2 days \\
        \midrule   
        \multicolumn{3}{c}{\oursgtf} \\
        \midrule 
        Action Scorer (same as (Section~\ref{subsection:action_selection}))
        & 64 (backprop) & 2 days \\
        \midrule   
        \multicolumn{3}{c}{Discrete Graph Encoder Pre-training} \\
        \midrule 
        Action Prediction w/ full graph, for \oursgtf (Appendix~\ref{appendix:pretrain_discrete_graph_encoder})
        & 256 & 2 days \\
        \midrule 
        Action Prediction w/ seen graph, for \oursgtp (Appendix~\ref{appendix:pretrain_discrete_graph_encoder})
        & 256 & 2 days \\
        \midrule 
        State Prediction w/ full graph, for \oursgtf (Appendix~\ref{appendix:pretrain_discrete_graph_encoder})
        & 48 & 5 days \\
        \midrule 
        State Prediction w/ seen graph, for \oursgtp (Appendix~\ref{appendix:pretrain_discrete_graph_encoder})
        & 48 & 5 days \\
        \midrule 
        Deep Graph Infomax w/ full graph, for \oursgtf (Appendix~\ref{appendix:pretrain_discrete_graph_encoder})
        & 256 & 1 day \\
        \midrule 
        Deep Graph Infomax w/ seen graph, for \oursgtp (Appendix~\ref{appendix:pretrain_discrete_graph_encoder})
        & 256 & 1 day \\
        \midrule   
        \multicolumn{3}{c}{Text-based Baselines} \\
        \midrule 
        \transdqn (Section~\ref{subsection:baselines})
        & 64 (backprop) & 2 days \\
        \midrule 
        \transdrqn (Section~\ref{subsection:baselines})
        & 64 (backprop) & 2 days \\
        \midrule 
        \transdrqnp (Section~\ref{subsection:baselines})
        & 64 (backprop) & 2 days \\
        \bottomrule
    \end{tabular}
\end{table}

\clearpage
\section{Details of the \ftwp dataset}
\label{appendix:ftwp}

Previously, \citet{trischler19ftwp} presented the \emph{First TextWorld Problems (FTWP)} dataset, which consists of TextWorld games that follow a cooking theme across a wide range of difficulty levels. 
Although this dataset is analogous to what we use in this work, it has only 10 games per difficulty level.
This is insufficient for reliable experiments on generalization, so we generate new game sets for our work.
As mentioned in Section~\ref{subsection:graph_update} and Appendix~\ref{appendix:pretrain_continuous}, we use a set of transitions collected from the \ftwp dataset.
To ensure the fairness of using this dataset, we make sure there is no overlap between the \ftwp and the games we use to train and evaluate our action selector.

\subsection*{Extracting Ground-truth Graphs from \ftwp Dataset}

Under the hood, TextWorld relies on predicate logic to handle the game dynamics. Therefore, the underlying game state consists of a set of predicates, and logic rules (i.e. actions) can be applied to update them. TextWorld's API allows us to obtain such underlying state $S_t$ at a given game step $t$ for any games generated by the framework. We leverage $S_t$ to extract both $\mathcal{G}^\text{full}_t$ and $\mathcal{G}^\text{seen}_t$. 

In which, $\mathcal{G}^\text{full}_t$ is a discrete KG that contains the full information of the current state at game step $t$; $\mathcal{G}^\text{seen}_t$ is a discrete partial KG that contains information the agent has observed from the beginning until step $t$. 

\begin{figure*}[htpb]
    \centering
    \begin{subfigure}[b]{0.3\textwidth}
        \centering
        \includegraphics[width=\textwidth]{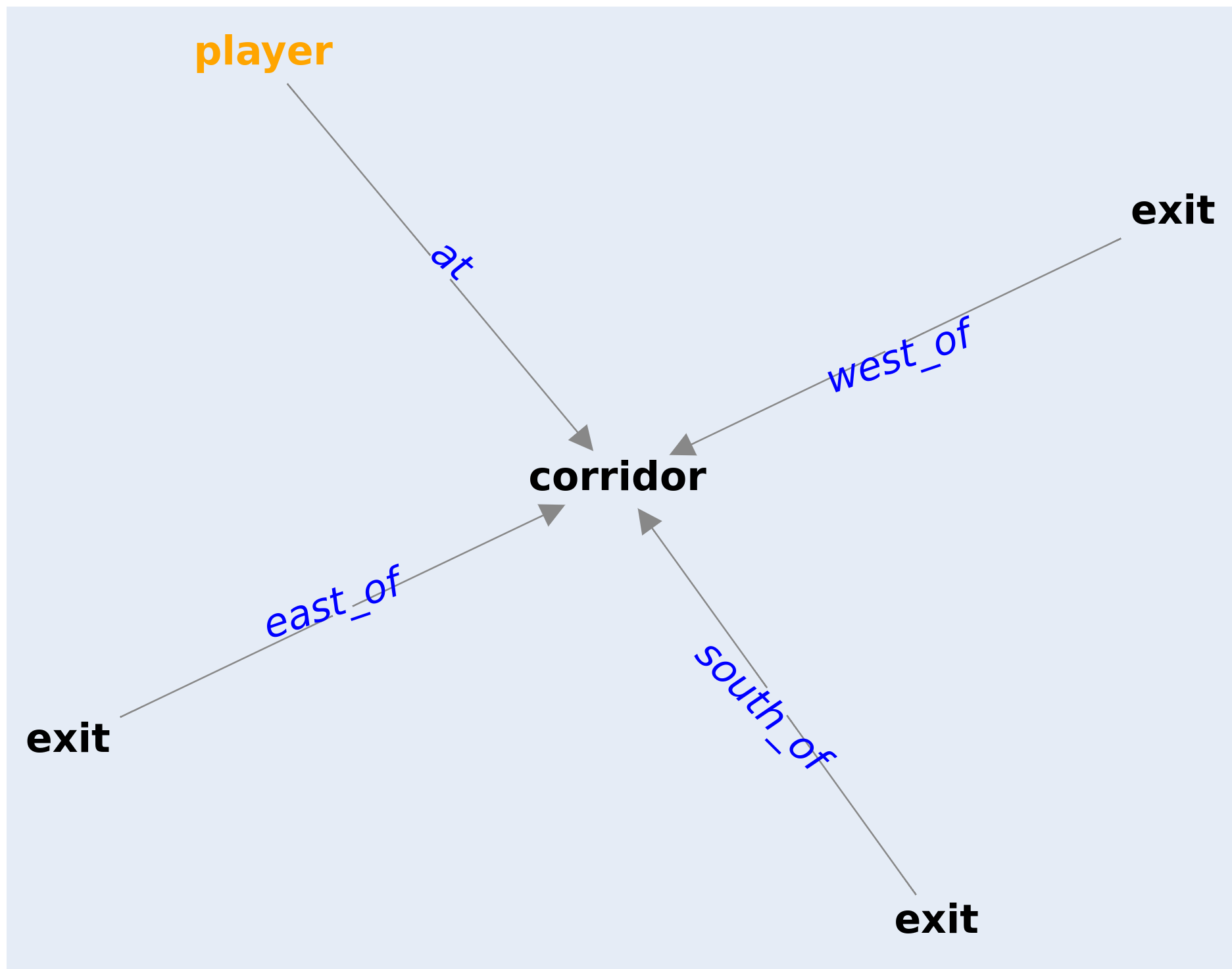}
        \caption{$\mathcal{G}^\text{seen}_0$ after starting the game.}
    \end{subfigure}
    \begin{subfigure}[b]{0.3\textwidth}
        \centering
        \includegraphics[width=\textwidth]{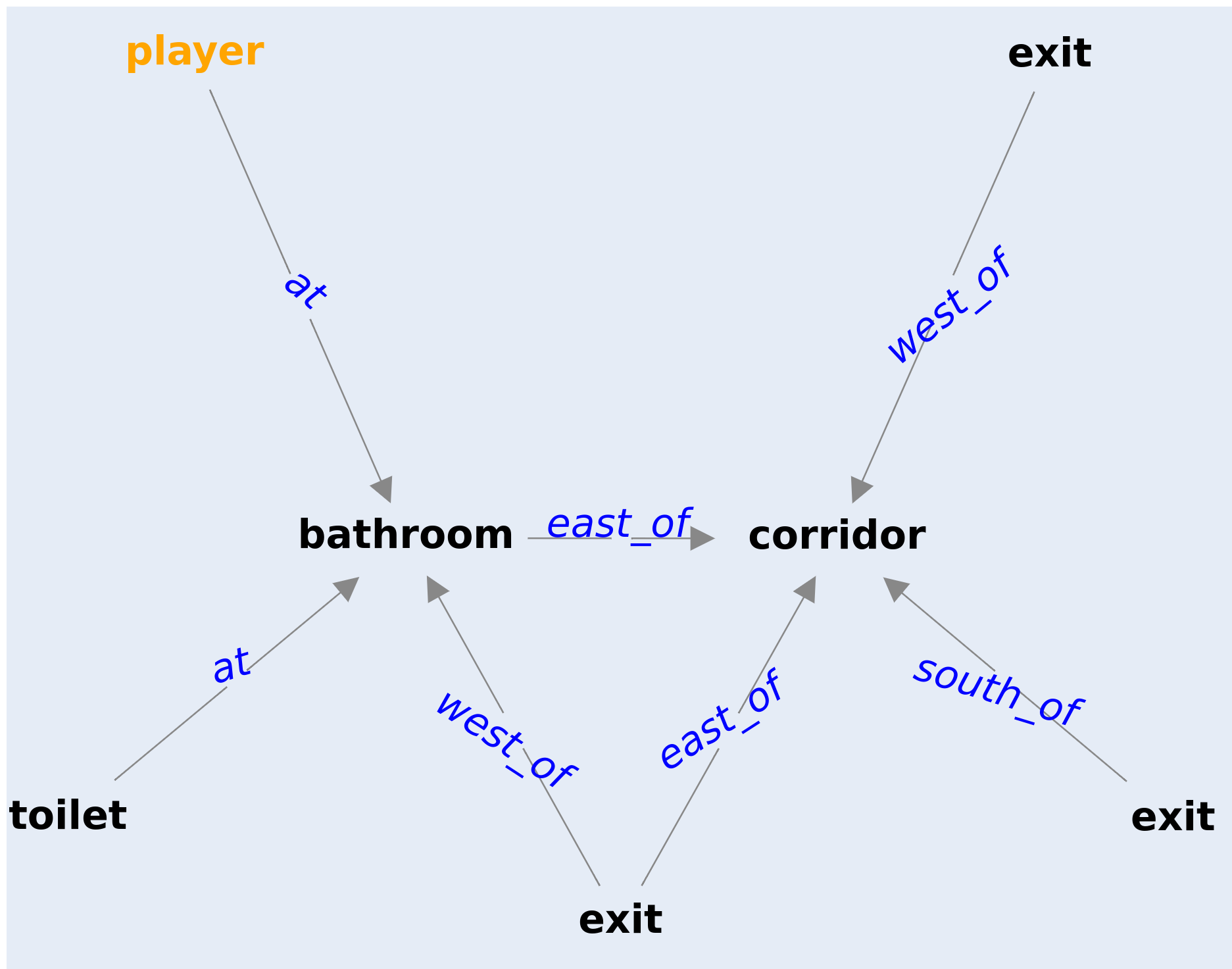}
        \caption{$\mathcal{G}^\text{seen}_1$ after \cmd{go east}.}
    \end{subfigure}
    \begin{subfigure}[b]{0.3\textwidth}
        \centering
        \includegraphics[width=\textwidth]{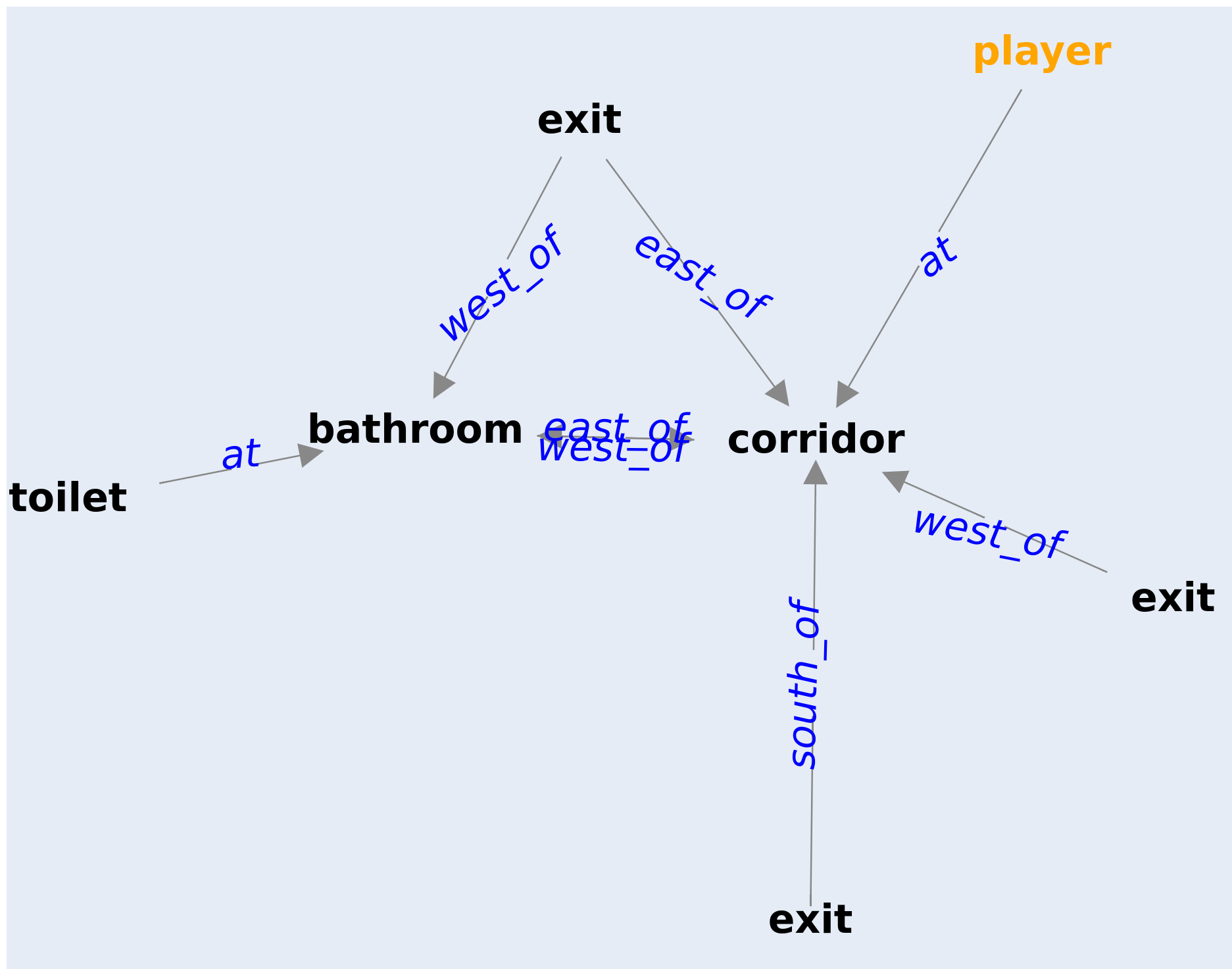}
        \caption{$\mathcal{G}^\text{seen}_2$ after \cmd{go west}.}
    \end{subfigure}
    \begin{subfigure}[b]{0.91\textwidth}
        \centering
        \includegraphics[width=\textwidth]{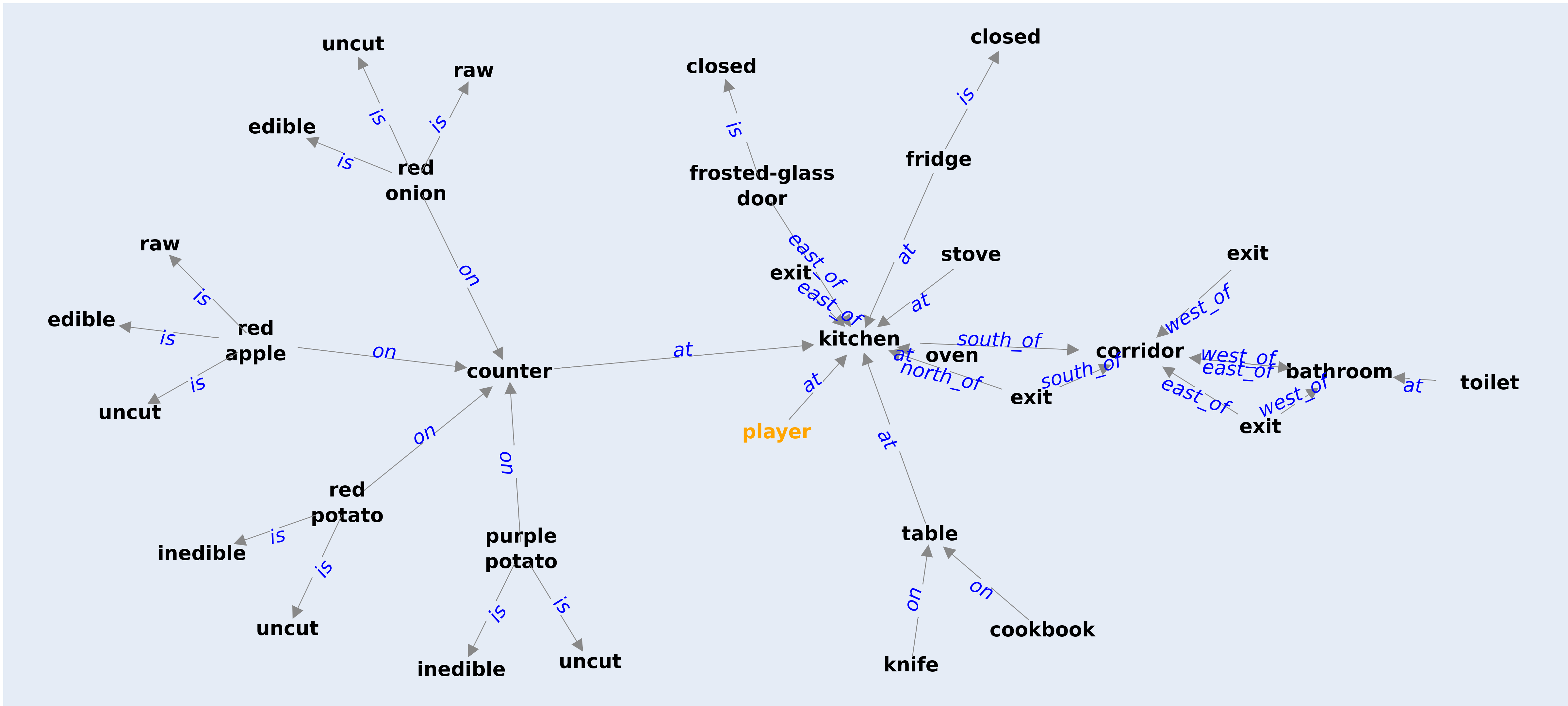}
        \caption{$\mathcal{G}^\text{seen}_3$ after \cmd{go south} that leads to the kitchen which contains many objects.}
    \end{subfigure}
    \caption{A sequence of $\mathcal{G}^\text{seen}$ extracted after issuing three consecutive actions in a \ftwp game.}
    \label{fig:kg_example_g_seen}
\end{figure*}

\begin{figure*}
    \centering
    \includegraphics[width=0.9\textwidth]{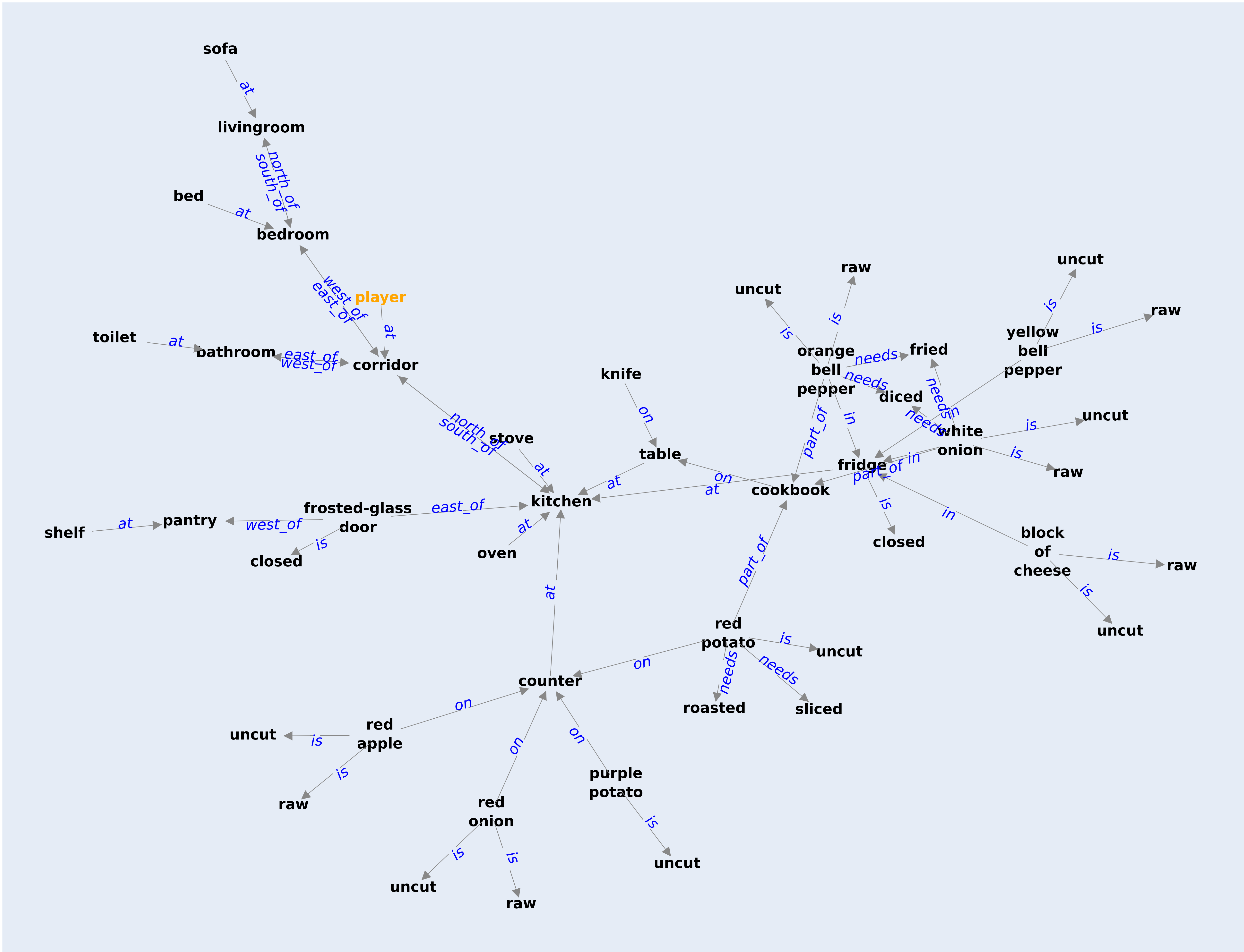}
    \caption{$\mathcal{G}^\text{full}$ at the start of a \ftwp game.}
    \label{fig:kg_example_g_full}
\end{figure*}

Figure~\ref{fig:kg_example_g_seen} shows an example of consecutive $\mathcal{G}^\text{seen}_t$ as the agent explores the environment of a \ftwp game. Figure~\ref{fig:kg_example_g_full} shows the $\mathcal{G}^\text{full}$ extracted from the same game.

\end{document}